\RequirePackage{fix-cm}
\documentclass[smallcondensed]{svjour3_copie}       
\smartqed  
\usepackage{graphicx}
\usepackage{algorithm}
\usepackage{algpseudocode}
\usepackage{lineno}
\usepackage{multirow}
\usepackage{color}
\usepackage{scrextend}
\usepackage{natbib}
\usepackage[overload]{empheq}
\usepackage{filecontents}
\usepackage[utf8]{inputenc}
\usepackage{lscape}
\usepackage{mathtools}
\usepackage{latexsym}
\usepackage{amssymb,amsmath}
\usepackage{subfig}
\usepackage{pgf}
\usepackage{tikz}
\usepackage{textgreek}
\usepackage{upgreek}
\usetikzlibrary{arrows,shapes,snakes,automata,backgrounds,petri}
\newtheorem{defi}{Definition}

\newcommand\numberthis{\addtocounter{equation}{1}\tag{\theequation}}

\newcommand*{\LargerCdot}{\raisebox{-0.25ex}{\scalebox{1.2}{$\cdot$}}}
\DeclareMathAlphabet\mathbfcal{OMS}{cmsy}{b}{n}
\journalname{Machine Learning}

\begin{document}

\title{Deep Recurrent Gaussian Process with Variational Sparse Spectrum Approximation
}
\subtitle{}


\author{Roman Föll \and
       Bernard Haasdonk \and
			 Markus Hanselmann \and
			 Holger Ulmer 
}


\institute{Roman Föll$^{1,2}$\\
					 \email{foell@mathematik.uni-stuttgart.de}, Tel.: +49-711-3423-2991\\
					 \\[-0.1cm]
           Bernard Haasdonk$^1$\\
					 \email{haasdonk@mathematik.uni-stuttgart.de}, Tel.: +49-711-685-65542 \\
					 \\[-0.1cm]
					 Markus Hanselmann$^2$\\
					 \email{markus.hanselmann@etas.com}, Tel.: +49-711-3423-2951\\
					 \\[-0.1cm]
					 Holger Ulmer$^2$\\
					 \email{holger.ulmer@etas.com}, Tel.: +49-711-3423-2359\\
					 \at $^1$ Institute of Applied Analysis and Numerical Simulation, \\
					 University of Stuttgart, Stuttgart, 70569, Germany\\
					 \at $^2$ ETAS GmbH, Stuttgart, 70469, Germany\\
}

\date{}

\maketitle

\begin{abstract}
Modeling sequential data has become more and more important in practice. Some applications are autonomous driving, virtual sensors and weather forecasting. To model such systems so called recurrent models are used. In this article we introduce two new Deep Recurrent Gaussian Process (DRGP) models based on the Sparse Spectrum Gaussian Process (SSGP) and the improved variational version called Variational Sparse Spectrum Gaussian Process (VSSGP). We follow the recurrent structure given by an existing DRGP based on a specific sparse Nyström approximation. Therefore, we also variationally integrate out the input-space and hence can propagate uncertainty through the layers. We can show that for the resulting lower bound an optimal variational distribution exists. Training is realized through optimizing the variational lower bound. Using Distributed Variational Inference (DVI), we can reduce the computational complexity. We improve over current state of the art methods in prediction accuracy for experimental data-sets used for their evaluation and introduce a new data-set for engine control, named Emission. Furthermore, our method can easily be adapted for unsupervised learning, e.g. the latent variable model and its deep version.
\end{abstract}

\section{Introduction}

Modeling sequential data for simulation tasks in the context of machine learning is hard for several reasons. Its internal structure poses the problem of modeling short term behavior and long term behavior together for different types of data variables, where data variables itself might differ in the information gain in the chosen time frequency. Recurrent models \citep{hochreiter1997long,nelles2013nonlinear,pascanu2013construct} have proven to perform well on these tasks by memorizing the sequential behavior of the output data additionally with the sequential behavior of the input data in a high-dimensional input-space.\\
The general form of a \textit{recurrent model} is given by
\begin{align*}
\mathbf{h}_i & = \upzeta(\mathbf{h}_{i-1},\dots,\mathbf{h}_{i-H_{\mathbf{h}}},\mathbf{x}_{i-1},\dots,\mathbf{x}_{i-H_{\mathbf{x}}}) + \boldsymbol\epsilon_i^{\mathbf{h}},\numberthis\label{PHI}\\
\mathbf{y}_i & = \uppsi(\mathbf{h}_i,\dots,\mathbf{h}_{i-H_{\mathbf{h}}})+\boldsymbol\epsilon_i^{\mathbf{y}},\numberthis\label{PSI}
\end{align*}
where $\mathbf{x}_i$ is an external input, $\mathbf{y}_i$ is an output observation, $\mathbf{h}_i$ is a latent hidden representation or state (details on dimensions and ranges will be specified in upcoming sections) at time $i=2,\dots,N$, where $N\in\mathbb{N}$ is the number of data samples, $H_{\mathbf{x}}$, $H_{\mathbf{h}}\in\mathbb N$ are the chosen time horizons, $\upzeta,\uppsi$ are non-linear functions modeling \textit{transition} and \textit{observation} and $\boldsymbol\epsilon_i^{\mathbf{h}}, \boldsymbol\epsilon_i^{\mathbf{y}}$ are transition, observation noise, which are adjusted for the specific problem. In this article we will deal with deep learning in a recurrent fashion for modeling sequential data in a Bayesian non-parametric approach by using \textit{Gaussian Processes} (GP). To make a connection to the general recurrent model, the deep structure arises by defining $\upzeta$ in \eqref{PHI} in a deep manner~\citep[see][Section~3]{pascanu2013construct}.\\
\textit{Bayesian Regression} (BR) for modeling in this context has several advantages. It solves the problem of model selection by maximizing the likelihood and offers the possibility to solve the problem of propagating uncertainty across sequential time, which has been adressed too short in the past literature. \textit{Gaussian Process Regression} (GPR) is often used in this context. GPs~\citep{rasmussen2006gaussian} have become a state of the art tool for modeling distributions over non-linear functions. The complexity of model selection by maximizing the likelihood, which comes with a cost of $\mathcal O(N^3)$ per iteration, where $N\in\mathbb N$ is again the number of data-samples now used in training, can be reduced by using sparse approximations. This comes with a cost of $\mathcal O(NM^2)$, where $M\in\mathbb N$ is the sparsity parameter typically satisfying $M\ll N$. The two approximations mainly used and also investigated in this paper are the sparse Nyström approximation of~\citep{titsias2009variational,titsias2010bayesian} and the approximation of the spectral representation by Bochner's theorem, in particular the \textit{Sparse Spectrum} (SS) approximation introduced in~\citep{quia2010sparse} and the improved \textit{Variational Sparse Spectrum} (VSS) approximation by~\citep{gal2015improving}.\\
In control and dynamical system identification previous work on Bayesian recurrent approaches for modeling sequential data usually make use of \textit{linear or non-linear auto-regressive with exogenous inputs models} ((N)ARX) or \textsl{state-space models} (SSM) \citep[for both see][]{nelles2013nonlinear}. The general recurrent model in \eqref{PHI}, \eqref{PSI} represents both cases. Auto-regressive means that one wants to explain or predict a current observation $\mathbf{y}_i$ of a time series depending on the past values $\mathbf{y}_{i-k}$ (observed or predicted), for $k=1,\dots, H_{\mathbf{y}}$, for discrete time $i\in\mathbb N$ and the chosen time horizon $H_{\mathbf{y}}\in\mathbb N$. Exogenous inputs are current or past values $\mathbf{x}_{i-k}$, for $k=0,\dots, H_{\mathbf{x}}$, its time horizon again denoted by $H_{\mathbf{x}}\in\mathbb N$, of a driving (exogenous) time series $\mathbf{x}_i$, which influences $\mathbf{y}_i$, the time series of interest. Standard (N)ARX models lack on performing auto-regression directly with the possibly noisy outputs, which is not taken into account. The SSM approach uses unobservable latent variables, the state, and constructs auto-regressive dynamics in the unobserved latent space. It is a more general alternative to (N)ARX models.\\
A \textit{SSM with GPs} (GP-SSM) for transition and observation functions is used by~\citep{wang2005gaussian}, where the uncertainty in the latent states is not accounted for, which can lead to overconfidence.~\citep{turner2010state} solved this problem, but they have complicated approximate training and inference stages and the model is hard to scale, as mentioned by~\citep{al2016learning}.~\citep{frigola2014variational} used a GP for transition, while the observation is parametric.~\citep{svensson2015computationally} used an approximation of the spectral representation by Bochner's theorem in a particular form and with a reduced rank structure for the transition function. They realize inference in a fully Bayesian approach over the amplitudes and the noise parameters. This GP-SSM approach differs from ours in the sense that we use a different form of a spectral representation and a deep structure for the transition function, where the overall training is done through a variational approach.\\
Two state of the art approaches for DRGPs have been introduced by~\citep{mattos2015recurrent} based on the sparse Nyström approximation introduced by~\citep{titsias2009variational,titsias2010bayesian} and~\citep{al2016learning} based on deep kernels via a \textit{Long-short term memory} (LSTM) network, a special type of \textsl{Recurrent Neural Network} (RNN). Short introductions of them will follow in Section \ref{sec:DeepRecurrentGaussianProcesses}.\\
Following~\citep{damianou2013deep}, a \textit{Deep Gaussian Process} (DGP) in the BR case is a model, which assumes
\begin{align*}
\mathbf{y}_i &= \mathrm{f}^{(\text{L+1})}(\mathrm{f}^{(L)}(\mathrm{f}^{(L-1)}(\dots(\mathrm{f}^{(1)}({\mathbf{x}}_i) + \boldsymbol\epsilon_i^{{\mathbf{h}^{(1)}}})\dots) + \boldsymbol\epsilon_i^{\mathbf{h}^{(L-1)}}) + \boldsymbol\epsilon_i^{\mathbf{h}^{(L)}}) + \boldsymbol\epsilon_i^{\mathbf{y}}, \numberthis\label{DEEPGP}
\end{align*}
where the index $i=1,\dots,N$ is \textit{not} necessarily the time and where we define $\mathbf{h}_i^{(1)} \stackrel{\mathrm{def}}= \mathrm{f}^{(1)}(\mathbf{x}_i) + \boldsymbol\epsilon_i^{\mathbf{h}^{(1)}}$, $\mathbf{h}_i^{(l+1)} \stackrel{\mathrm{def}}= \mathrm{f}^{(l)}(\mathbf{h}_i^{(l)}) + \boldsymbol\epsilon_i^{\mathbf{h}^{(l)}}$, for $l=2\dots,L-1$, where $L\in\mathbb N$ is the number of hidden layers, $\mathbf{x}_i$ is the input data, $\mathbf{y}_i$ are the output observations and $\mathbf{h}_i^{(l)}$ represent the hidden states. The noise $\boldsymbol\epsilon_i^{\mathbf{h}^{(l)}}$, $\boldsymbol\epsilon_i^{\mathbf{y}}$ is now assumed to be Gaussian and the functions $\mathrm{f}^{(l)}$ are modeled with GPs for $l=1,\dots,L+1$.\\
One of the problems that is often encountered when dealing with DGPs is that the training is hard, because the resulting likelihood integral is analytically intractable. A solution to this is the \textit{Variational Inference} (VI) framework. Here, one specifies a variational distribution, whose structure is easy to evaluate, and tries to bring it as close as possible to the true posterior distribution of the Bayesian approach. This procedure is still challenging, as integrals of covariance functions have to be solved.~\citep{damianou2013deep} introduced these kind of DGPs based on the sparse approximation following~\citep{titsias2009variational,titsias2010bayesian}.~\citep{cutajar2016practical} introduced them for the so called \textit{Random Fourier Features} (RFF) approach, where the variational weights for each layer are optimized along with the hyperparameters. This approach does not variationally integrate out the latent inputs to carry through the uncertainty and no existence of a optimal distribution for the variational weights is proven, to eliminate these and to reduce the amount of parameters to optimize in training. Additionally, no recurrent structure for the whole setting is assumed to model sequential data.\\
One of many advantages of a deep structure as previously introduced is that it divides the one hard modeling problem in several smaller ones, which are easier to model. This is done by successively creating features, which are maintained by passing through the hierarchy from the input layer to the output layer. Therefore, one is able to solve the challenging situation of long-short term behavior and the different data types with their different information gain in the chosen time frequency, when adapting the deep framework for modeling sequential data. Clearly, for \eqref{DEEPGP} we have then to introduce an input structure like in \eqref{PHI} or \eqref{PSI}. The DGP prior, which is richer than a standard GP prior, is able to model the stationary and non-stationary processes better than standard GPs \citep[see][Section~6.1.,6.2]{damianou2015deep}. Through its compositional form the DGP model is non-stationary even when the underlying covariance functions of every GP layer are stationary. In our case, the resulting sparse covariance functions for the processes in between are non-stationary as well, even when the covariance functions used for approximation are stationary. This is an observation similar to~\citep{gal2015improving}. The DGP is no GP anymore. For a deeper understanding and a more complete overview of the recurrent modeling case of sequential data in a Bayesian and non-Bayesian approach we refer to the introductions of~\citep{turner2010state,frigola2014variational,mattos2015recurrent,svensson2015computationally,al2016learning}.\\	
The contribution of this article is the extension of the sparse GP based on the SS approximation introduced in~\citep{quia2010sparse} and the improved VSS approximation by~\citep{gal2015improving} to a DRGP, following the same deep recurrent structure as introduced in~\citep{mattos2015recurrent}. The DRGP of~\citep{mattos2015recurrent} is limited to a quite small class of deterministic covariance functions, because the covariance functions expectation has to be analytically tractable. These are for example the squared exponential or the linear covariance function. Using the SS approximation instead, and thus sampling from the spectral representation by Bochner's theorem, we can derive for every stationary covariance function a valid approximation, since the distribution of the spectral points defines which covariance function is approximated and the basis functions expectation is always tractable. For the VSS approximation we additionally have to take care of the tractability of some \textit{Kullback-Leibler} (KL) term for training. One can interpret these approximations as random finite rank covariance functions. We show that this approach improves over two state of the art approaches in prediction accuracy on the experimental data-sets used in~\citep{mattos2015recurrent,svensson2015computationally,al2016learning} in a simulation setting. For scalability, DVI~\citep{gal2014distributed} can lower the complexity from $\mathcal O(NM^2Q_{\text{max}}L)$ down to $\mathcal O(M^3)$, where $Q_{\text{max}}$ is the maximum over all input dimensions used in our defined deep structure for $\upzeta$ and $\uppsi$ in the Equations in \eqref{INPUT} of Section \ref{sec:DeepRecurrentGaussianProcesses}.\\
In the next Section \ref{sec:GaussianProcessesandGaussianProcessRegression} we will briefly introduce GP and GPR, while we assume a basic understanding of these models, followed by a short introduction of the two state of the art models. We then introduce and formalize our DRGPs in detail and present our results in Section \ref{sec:experiments} on several data-sets and compare them to other sparse GPs and the full GP using NARX structure, the GP-SSM of~\citep{svensson2015computationally} and the two competing state of the art DRGPs. Finally, we discuss our methods and give an overview on future work.

\section{Gaussian Processes and Gaussian Process Regression}
\label{sec:GaussianProcessesandGaussianProcessRegression}
Loosely speaking, a GP can be seen as a Gaussian distribution over functions. We will first introduce a general definition of GPs and will then introduce the specific case for data-based modeling. In the following we use the notation $\mathbf{X}$, $\mathbf{x}$, $\mathrm{x}, \mathrm{f}$ (upright) for our set of data points, a single data point, sample, function and $\boldsymbol{X}$, $\boldsymbol{x}$, $x$, $f$ (italic) for the corresponding set of random vectors, a single random vector, random variable, GP.\\ 
\begin{defi}[Gaussian Process]
Let $(\Omega,\mathcal A, P)$ be a probability space, $\left(\Omega',\mathcal A'\right)$ a measure space, $\mathcal{X}$ some infinite index set and \\ $f:\left(\Omega,\mathcal A, P\right)\times {\mathcal{X}}\to \left(\Omega',\mathcal A'\right), (\omega,\mathbf{x})\mapsto f_{\mathbf{x}}(\omega)$ a stochastic process, where\\ $f_{\mathbf{x}}:\omega \mapsto f_{\mathbf{x}}(\omega)$ is a random variable for a fixed $\mathbf{x}$.\\
A stochastic process $f$ is a GP if and only if for any $N\in\mathbb N$, $\mathbf{x}_1,\dots,\mathbf{x}_N\in \mathcal X$ and $c_1,\dots,c_N\in\mathbb R$, the random variable $c_1 f_{\mathbf{x}_1}+\dots+c_N f_{\mathbf{x}_N}$ is Gaussian. This holds in particular if the $f_{\mathbf{x}}$ are independent Gaussian random variables~\citep[Chapter~11]{kallenberg2006foundations}.\\
\end{defi}\vspace{-0.5cm}
In other words, a stochastic process $f$ is a GP if and only if any finite collection of random variables $f_{\mathbf{X}} \stackrel{\mathrm{def}}=\left[f_{\mathbf{x}_1},\dots,f_{\mathbf{x}_N}\right]^T$ forms a Gaussian random vector~\citep{rasmussen2006gaussian}. A GP is a conditional probabilistic model, where the random vector $f_{\mathbf{X}}$ can be written as $f_{\mathbf{X}}|\boldsymbol{X}$ and where the distribution for $\boldsymbol{X}$ is not specified explicitly. This seems to be tautologous, but we want to denote the random vector at position $\mathbf{X}$, not the whole GP $f$.\\
Let $\mathcal B$ be the Borel $\sigma$-Algebra and set $\mathcal A'=\mathcal B$. For data based modeling we will use the real GP $f:\left(\Omega,\mathcal A, P\right)\times\mathbb R^Q\to \left(\mathbb{R},\mathcal B\right),(\omega,\mathbf{x})\mapsto f_{\mathbf{x}}(\omega)$, where $Q\in\mathbb N$ is the input-space dimensionality. For simplicity we will skip $\omega$ from now on. We further write $c\sim\mathcal{N}(\mathrm{m}_{c},\upsigma_{c}^2)$ for a random variable $c$ belonging to a specific distribution, here a Gaussian distribution with mean $\mathrm{m}_{c}$, variance $\upsigma_{c}^2$.\\
A GP is completely defined by its mean function $m: \mathbb R^Q\to\mathbb{R},\mathbf{x}\mapsto m(\mathbf{x})$ and covariance function $k: \mathbb R^Q\times\mathbb R^Q\to\mathbb{R},(\mathbf{x},\mathbf{x'})\mapsto k(\mathbf{x},\mathbf{x'})$~\citep[see][Lemma 11.1]{kallenberg2006foundations}, where
\begin{align*}
m(\mathbf{x}) &\stackrel{\mathrm{def}}= E\left[f(\mathbf{x})\right],\\
k(\mathbf{x},\mathbf{x'}) &\stackrel{\mathrm{def}}= \text{cov}(f(\mathbf{x}),f(\mathbf{x'}))=E\left[(f(\mathbf{x})-m(\mathbf{x}))(f(\mathbf{x'})-m(\mathbf{x'}))\right],
\end{align*}
and the GP will be written as $f\sim\mathcal{GP}(m,k)$.\\
As in the following we will work with probability densities, we introduce the abbreviations $p_{c}$, which means that a random variable $c$ has a probability density $p_{c}$. We define our set of input-data as matrix $\mathbf{X}\stackrel{\mathrm{def}}=[\mathbf{x}_1,\dots,\mathbf{x}_N]^T\in\mathbb R^{N\times Q}$, the set of output observations as vector $\mathbf{y}\stackrel{\mathrm{def}}=[\mathrm{y}_1,\dots,\mathrm{y}_N]^T\in\mathbb{R}^{N}$, $K_{NN} \stackrel{\mathrm{def}}=(k(\mathbf{x}_i,\mathbf{x}_j))_{i,j=1}^N$, $K_{\ast\ast} \stackrel{\mathrm{def}}= k(\mathbf{x}^{\ast},\mathbf{x}^{\ast})\in\mathbb R$ for a test point $\mathbf{x}^{\ast}\in\mathbb R^Q$, $K_{N\ast} \stackrel{\mathrm{def}}= (k(\mathbf{x}_i,\mathbf{x}^{\ast}))_{i=1}^N\in\mathbb R^{N\times 1}$, $K_{\ast N} \stackrel{\mathrm{def}}=K_{N\ast}^T$.\\
A GPR model is a non-parametric and Bayesian conditional probabilistic model where we assume
\begin{align*}
\mathrm{y}_i = \mathrm{f}(\mathbf{x}_i) + \epsilon_i^{\mathrm{y}}, \quad\quad \epsilon_{i}^{\mathrm{y}}\sim\mathcal{N}(0,\upsigma_{\text{noise}}^2),
\end{align*}
for $i = 1,\dots,N$, where a function $\mathrm{f}:\mathbb R^Q\to\mathbb R$ maps an input-point $\mathbf{x}_i$ to the observation $\mathrm{y}_i$, which is corrupted by Gaussian noise $\epsilon_{i}^{\mathrm{y}}$. Our aim is to model any set of function values $\mathbf{f}=[\mathrm{f}(\mathbf{x}_1),\dots,\mathrm{f}(\mathbf{x}_N)]^T\in\mathbb{R}^N$ as samples from a random vector $f_{\mathbf{X}}=\left[f_{\mathbf{x}_1},\dots,f_{\mathbf{x}_N}\right]^T$. Therefore, we assume 
\begin{align*}
f_\mathbf{X}|\boldsymbol{X}\sim \mathcal{N}(\mathbf{0},K_{NN}),\quad \text{the prior},
\end{align*}
meaning that any set of function values $\mathbf{f}$ given $\mathbf{X}$ are jointly Gaussian distributed with mean $\mathbf{0}\in\mathbb R^N$  and a covariance matrix $K_{NN}\in\mathbb R^{N\times N}$.\\ Equivalently, the likelihood is $p_{\boldsymbol{y}|f_\mathbf{X},\boldsymbol{X}}(\mathbf{y}|\mathbf{f},\mathbf{X})$ with $\boldsymbol{y}|f_\mathbf{X},\boldsymbol{X}\sim \mathcal{N}(\mathbf{f},\upsigma_{\text{noise}}^2I_N)$, where $I_N\in\mathbb R^{N\times N}$ denotes the identity matrix. When we define the prior, the chosen covariance function $k$ might have some parameters, called hyperparameters. For brevity we skipped the dependence of $f_\mathbf{X}|\boldsymbol{X},\boldsymbol\theta$ on the hyperparameters\linebreak $\textnormal{\fontfamily{phv}\selectfont\straighttheta}\in\mathbb R^P,P\in\mathbb N$ of the covariance function.\\
Furthermore we can derive the \textit{marginal likelihood} (ML) as the integral over the likelihood times the prior
\begin{align*}
p_{\boldsymbol{y}|\boldsymbol{X}}(\mathbf{y}|\mathbf{X})= \int p_{\boldsymbol{y}|f_\mathbf{X},\boldsymbol{X}}(\mathbf{y}|\mathbf{f},\mathbf{X})p_{f_\mathbf{X}|\boldsymbol{X}}(\mathbf{f}|\mathbf{X}) d\mathbf{f}, 
\end{align*}
where $\boldsymbol{y}|\boldsymbol{X}\sim\mathcal{N}(\mathbf{0},K_{NN} + \upsigma_{\text{noise}}^2 I_N)$.\\
The predictive distribution $p_{f_\mathbf{x^{\ast}}|\boldsymbol{x^{\ast}},\boldsymbol{X},\boldsymbol{y}}$ for a test point $\mathbf{x}^{\ast}\in\mathbb R^{Q}$ can be derived through the joint probability model
\begin{align*}
\begin{bmatrix}f_\mathbf{x^{\ast}}\\\boldsymbol{y}\end{bmatrix}\Big|\begin{bmatrix}\boldsymbol{x}^{\ast}\\\boldsymbol{X}\end{bmatrix}\sim \mathcal{N}\left(\mathbf{0},\begin{bmatrix}K_{\ast\ast} & K_{N\ast}\\
K_{\ast N} & K_{NN} +\upsigma_{\text{noise}}^2I_N \end{bmatrix}\right),
\end{align*}
and conditioning through $p_{f_\mathbf{x^{\ast}}|\boldsymbol{x^{\ast}},\boldsymbol{X},\boldsymbol{y}} = \frac{p_{f_\mathbf{x^{\ast}},\boldsymbol{y}|\boldsymbol{x^{\ast}},\boldsymbol{X}}({\LargerCdot},\mathbf{y}|\mathbf{x}^{\ast},\mathbf{X})}{p_{\boldsymbol{y}|\boldsymbol{X}}(\mathbf{y}|\mathbf{X})}$ as
\begin{align*}
&f_\mathbf{x^{\ast}}|\boldsymbol{x^{\ast}},\boldsymbol{X},\boldsymbol{y}\\&\sim \mathcal{N}(K_{\ast N}(K_{NN} + \upsigma_{\text{noise}}^2 I_N)^{-1}\mathbf{y},K_{\ast\ast}- K_{\ast N}(K_{NN} + \upsigma_{\text{noise}}^2 I_N)^{-1}K_{N\ast}).
\end{align*}
The GPR model is trained by maximizing the log ML of the data given the model with respect to $\textnormal{\fontfamily{phv}\selectfont\straighttheta}$ and $\upsigma_{\text{noise}}$. It has the form
\begin{align*}
&\log(p_{\boldsymbol{y}|\boldsymbol{X}}(\mathbf{y}|\mathbf{X}))\\ &= -\frac{1}{2}\mathbf{y}^T(K_{NN}+\upsigma_{\text{noise}}^2I_N)^{-1}\mathbf{y} - \frac{1}{2}\log(|K_{NN}+\upsigma_{\text{noise}}^2I_N|) - \frac{N}{2}\log(2\pi),
\end{align*}
where $|\;\LargerCdot\;|$ denotes the determinant and $\log$ is the natural logarithm.\\
When we have a set of multi-output observations $\mathbf{Y}=[\mathbf{y}_1,\dots,\mathbf{y}_N]^T\in\mathbb R^{N\times D}$, a straightforward way to use GPs is using a GP for every output component.
\section{Deep Recurrent Gaussian Processes}
\label{sec:DeepRecurrentGaussianProcesses}
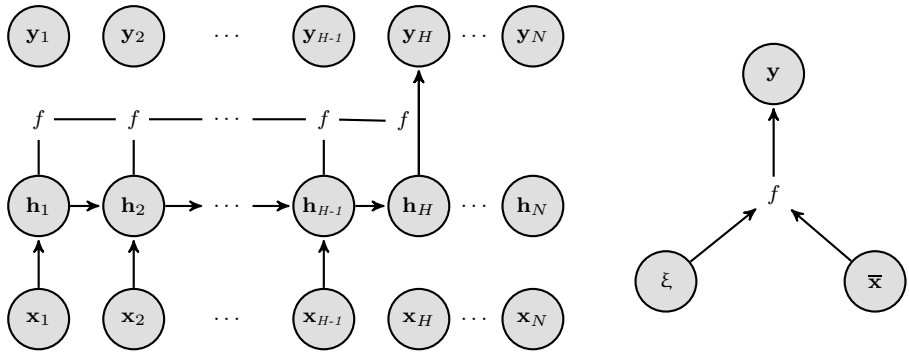
\begin{figure}
\centering
\begin{tikzpicture}[->,>=stealth',shorten >=1pt,auto,node distance=3cm,
        thick,main node/.style={circle,fill=lightgray!50,draw,minimum size=0.8cm,inner sep=0pt}]
\node[main node]  (1) at (0,0.25) {\footnotesize{$\mathbf{x}_1$}};
\node[main node] (2) at (1.25,0.25) {\footnotesize{$\mathbf{x}_2$}};
\node[] (3) at (2.5,0.25) {$\dots$};
\node[main node] (4) at (3.75,0.25) {\footnotesize{$\mathbf{x}_{\text{\tiny\textit{H-1}}}$}};
\node[main node] (5) at (5,0.25) {\footnotesize{$\mathbf{x}_H$}};
\node[] (25) at (5.75,0.25) {$\dots$};
\node[main node] (5) at (6.5,0.25) {\footnotesize{$\mathbf{x}_N$}};
\node[main node] (6) at (0,1.75) {\footnotesize{$\mathbf{h}_1$}};
\node[main node] (7) at (1.25,1.75) {\footnotesize{$\mathbf{h}_2$}};
\node[] (8) at (2.5,1.75) {$\dots$};
\node[main node] (9) at (3.75,1.75) {\footnotesize{$\mathbf{h}_{\text{\tiny\textit{H-1}}}$}};
\node[main node] (10) at (5,1.75) {\footnotesize{$\mathbf{h}_H$}};
\node[] (23) at (5.75,1.75) {$\dots$};
\node[main node] (24) at (6.5,1.75) {\footnotesize{$\mathbf{h}_N$}};
\node[main node] (11) at (0,4) {\footnotesize{$\mathbf{y}_1$}};
\node[main node] (12) at (1.25,4) {\footnotesize{$\mathbf{y}_2$}};
\node[] (13) at (2.5,4) {$\dots$};
\node[main node] (14) at (3.75,4) {\footnotesize{$\mathbf{y}_{\text{\tiny\textit{H-1}}}$}};
\node[main node] (15) at (5,4) {\footnotesize{$\mathbf{y}_H$}};
\node[] (21) at (5.75,4) {$\dots$};
\node[main node] (22) at (6.5,4) {\footnotesize{$\mathbf{y}_N$}};
\node[] (16) at (0,2.9) {$f$};
\node[] (17) at (1.25,2.9) {\footnotesize{$f$}};
\node[] (18) at (2.5,2.9) {$\dots$};
\node[] (19) at (3.75,2.9) {\footnotesize{$f$}};
\draw[->] (10) -- node (20) {\footnotesize{$f$}} (15);
\draw[-] (6) -- (16);
\draw[-] (7) -- (17);
\draw[] (8)  (18);
\draw[-] (9) -- (19);
\draw[-] (16) -- (17);
\draw[-] (17) -- (18);
\draw[-] (18) -- (19);
\draw[-] (19) -- (20);
\draw[->] (6) -- (7);
\draw[->] (7) -- (8);
\draw[->] (8) -- (9);
\draw[->] (9) -- (10);
\draw[->] (1) -- (6);
\draw[->] (2) -- (7);
\draw[] (3)  (8);
\draw[->] (4) -- (9);
\draw[->] (10) -- (15);
\node[main node] (26) at (8.25,0.75) {\footnotesize{$\upxi$}};
\node (27) at (11.75,1.5) {};
\node[main node] (28) at (11.0,0.75) {\footnotesize{$\overline{\mathbf{x}}$}};
\node[] (29) at (9.666,1.9) {\footnotesize{$f$}};
\node[main node] (30) at (9.666,3.5) {\footnotesize{$\mathbf{y}$}};
\draw[->] (26) -- (29);
\draw[->] (28) -- (29);
\draw[->] (29) -- (30);
\end{tikzpicture}
\caption{Graphical illustration of GP-LSTM. Here symbols without arrow are data sequences not used in the given time step. $\mathbf{x}_i$, $\mathbf{y}_i$ represent the data, $\mathbf{h}_i$ represents the learned latent representations (here multi-dimensional output case for observations and latent outputs) via LSTM for $i = 1,\dots,N$, where $N$ is the number of data and $H$ a time horizon, and they are mapped to the observed outputs through a GP. Illustration adjusted to our notation from~\citep{al2016learning}}\label{fig:DRGPL} 
\end{figure}

In this section we will introduce two DRGPs, first from~\citep{mattos2015recurrent}, which we will name DRGP-Nyström and which will be explained in more detail and second from~\citep{al2016learning}, which we name GP-LSTM.\\
DRGP-Nyström extends the GP-SSM framework to regression on sequences by using a recurrent construction, where the auto-regressive structure is not realized directly with the observed output-data, but with latent (non-observed) variables, the states, and is inserted in a VI procedure named \textsl{Recurrent Variational Bayes} (REVARB). The structure acts like a standard RNN, where every parametric layer is a GP. So a deep structure naturally arises and additionally uncertainty information can be carried through the hidden layers.\\
We define it for the 1-dimensional output observation case $y_i\in\mathbb R$ (it is directly expandable for the multi-dimensional output case $\mathbf{y}_i\in\mathbb R^{D}$, as mentioned at the end of Section \ref{sec:GaussianProcessesandGaussianProcessRegression}). We have $\mathbf{X}$ and $\mathbf{y}$ the sets of exogenous input-data and output observations defined as in Section \ref{sec:GaussianProcessesandGaussianProcessRegression} with $\mathbf{h}^{(l)} = [\mathrm{h}_{1+H_{\mathbf{x}}-H_{\mathrm{h}}}^{(l)},\dots,\mathrm{h}_N^{(l)}]^T\in\mathbb{R}^{N+H_{\mathrm{h}}-H_{\mathbf{x}}}$ the set of 1-dimensional latent variables (not-observed, also  expandable for the multi-dimensional case), $N\in\mathbb N$ is the number of observations, $H_{\mathrm{h}}$, $H_{\mathbf{x}}\in\mathbb N$ are time-horizons, for $l = 1,\dots,L$, where $L\in\mathbb N$ are the hidden layers. The detailed structure of DRGP-Nyström, where $i$ represents now the indexed time, is given by
\begin{align*}
\upzeta: \quad&\mathrm{h}_i^{(l)} = \mathrm{f}^{(l)}(\hat{\mathbf{h}}_{i}^{(l)})+\epsilon_i^{{h}^{(l)}},&&\text{with prior}\quad f^{(l)}_{\hat{\mathbf{H}}^{(l)}}|\hat{\boldsymbol{H}}^{(l)}\sim\mathcal{N}(\mathbf{0},K_{\hat{N}\hat{N}}^{(l)}), && l = 1,\dots,L\\
\uppsi: \quad&\mathrm{y}_i = \mathrm{f}^{(l)}(\hat{\mathbf{h}}_{i}^{(l)})+\epsilon_i^{y},&&\text{with prior}\quad f^{(l)}_{\hat{\mathbf{H}}^{(l)}}|\hat{\boldsymbol{H}}^{(l)}\sim\mathcal{N}(\mathbf{0},K_{\hat{N}\hat{N}}^{(l)}), && l = L+1
\end{align*}
with $\upzeta$, $\uppsi$ in \eqref{PHI}, \eqref{PSI}, $\epsilon_i^{\mathrm{h}^{(l)}}\sim \mathcal{N}(0,(\upsigma_{\text{noise}}^{(l)})^2)$, $\epsilon_i^{\mathrm{y}}\sim \mathcal{N}(0,\upsigma_{\text{noise}}^2)$ and $\hat{N}=N-H_{\mathbf{x}}$, for $i = H_{\mathbf{x}} + 1,\dots,N$. The matrix $K_{\hat{N}\hat{N}}^{(l)}$ represents a covariance matrix for a given covariance function $k$ and a set of input-data $\hat{\mathbf{H}}^{(l)}=[\hat{\mathbf{h}}_{H_{\mathbf{x}}+1}^{(l)},\dots,\hat{\mathbf{h}}_N^{(l)}]^T$ is specified as \\
\begin{flalign*}[left={\hat{\mathbf{h}}_i^{(l)} \stackrel{\mathrm{def}}
= \empheqlbrace}]
     && \begin{bmatrix}\overline{\mathbf{h}}_{i-1}^{(1)}\\\overline{\mathbf{x}}_{i-1}\end{bmatrix}&&\kern-1em\stackrel{\mathrm{def}}=&\left[\left[\mathrm{h}_{i-1}^{(1)},\dots,\mathrm{h}_{i-H_{\mathrm{h}}}^{(1)}\right],\left[\mathbf{x}_{i-1},\dots,\mathbf{x}_{i-H_{\mathbf{x}}}\right]\right]^T,&& l = 1\numberthis\label{INPUT}\\
     && \quad\begin{bmatrix}\overline{\mathbf{h}}_{i-1}^{(l)}\\\overline{\mathbf{h}}_{i}^{(l-1)}\end{bmatrix}&&\kern-1em\stackrel{\mathrm{def}}=&\left[\left[\mathrm{h}_{i-1}^{(l)},\dots,\mathrm{h}_{i-H_{\mathrm{h}}}^{(l)}\right],\left[\mathrm{h}_{i}^{(l-1)},\dots,\mathrm{h}_{i-H_{\mathrm{h}} + 1}^{(l-1)}\right]\right]^T,&&l = 2,\dots,L\\
     && \overline{\mathbf{h}}_{i}^{(L)}&&\kern-1em\stackrel{\mathrm{def}}=&\left[\mathrm{h}_{i}^{(L)},\dots,\mathrm{h}_{i-H_{\mathrm{h}} + 1}^{(L)}\right]^T,&&l = L + 1,
\end{flalign*}
where $\hat{\mathbf{h}}_i^{(1)}\in\mathbb R^{H_{\mathrm{h}}+H_{\mathbf{x}}Q}$, $\hat{\mathbf{h}}_i^{(l)}\in\mathbb R^{2H_h}$ for $l = 2,\dots,L$, $\hat{\mathbf{h}}_i^{(L+1)}\in\mathbb R^{H_{\mathrm{h}}}$, for\linebreak $i = H_{\mathbf{x}} + 1,\dots,N$. For simplification we set $H\stackrel{\mathrm{def}}=H_{\mathbf{x}}=H_{\mathrm{h}}$ in our experiments. In this form inference is intractable, because one is not able to get a closed analytical expression for the true posterior of $f^{(l)}$ for every layer or the ML. By introducing new variables and variational distributions, following the variational sparse framework proposed by~\citep{titsias2009variational,titsias2010bayesian}, a REVARB bound can be derived. The training is done through maximizing this REVARB-bound. An extension in~\citep{mattos2015recurrent} combines the DRGP-Nyström and RNN, which we will skip here, because we just have implemented the version described above with our new derived sparse variational bounds for our methods. This also means that our experimental comparisons in Section \ref{sec:experiments} involve the DRGP-Nyström without extension. Nevertheless, this extension would also work for our methods. A graphical illustration of the DRGP-Nyström is given in Figure \ref{fig:DRGPN}.\\
\begin{figure}
\centering
\begin{tikzpicture}[->,>=stealth',shorten >=1pt,auto,node distance=3cm,
        thick,main node/.style={circle,fill=lightgray!50,draw,minimum size=0.8cm,inner sep=0pt]}]
\node[main node] (1) at (0,0) {\footnotesize{$\mathbf{x}$}};
\node[main node] (2) at (2,0) {\footnotesize{$\overline{\mathbf{h}}^{(1)}$}};
\node[main node] (3) at (4,0) {\footnotesize{$\overline{\mathbf{h}}^{(2)}$}};
\node[] (4) at (6,0) {$\dots$};
\node[main node] (5) at (8,0) {\footnotesize{$\overline{\mathbf{h}}^{(L)}$}};
\node[main node] (6) at (10,0) {\footnotesize{$\mathrm{y}$}};
\draw[->] (1) -- node {\footnotesize{$f^{(1)}$}} (2);
\draw[->] (2) -- node {\footnotesize{$f^{(2)}$}} (3);
\draw[->] (3) -- node {\footnotesize{$f^{(3)}$}} (4);
\draw[->] (4) -- node {\footnotesize{$f^{(L)}$}} (5);
\draw[->] (5) -- node {\footnotesize{$f^{(L+1)}$}} (6);
\path[->] (2) edge [loop above] ()
(3) edge [loop above] ()
(5) edge [loop above] ();
\end{tikzpicture}
\caption{Graphical illustration of DRGP-Nyström. Between the hidden layers GPs are used for the mapping and also the last hidden representation $\overline{\mathbf{h}}^{(L)}$ is mapped to the observed output $\mathrm{y}$ via a GP. We see that just the latent variables $\overline{\mathbf{h}}^{(l)}$ are recurrent (illustrated via circle arrows). Illustration adjusted to our notation from~\citep{mattos2015recurrent}} \label{fig:DRGPN} 
\end{figure}
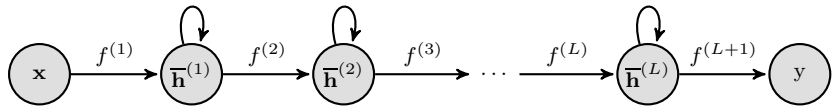GP-LSTM is a combination of GPs and LSTMs. LSTMs try to overcome vanishing gradients by placing a memory cell into each hidden unit. Special update rules for the hidden representations are used. So it adds stability to the network’s memory. LSTMs have proven to perform well on modeling sequential data. The GP-LSTM from~\citep{al2016learning} combines a GP with the advantages of LSTMs in the sense of defining structured recurrent deep covariance functions, also called deep kernels. These kernels entirely encapsulate the structural properties of LSTMs for combination with GPs. The specific covariance function $\tilde k$ is constructed in the sense that for an arbitrary deterministic transformation $\upxi:\mathbb R^{QH}\to \mathcal H, \bar{\mathbf{x}}=\left[\mathbf{x}_{i-1},\dots,\mathbf{x}_{i-H_{\mathbf{x}}}\right]\to\upxi(\bar{\mathbf{x}})$ (in this specific case this is the LSTM without the output layer or observation function), where $H=H_{\mathbf{x}}=H_{\mathbf{h}}\in\mathbb N$ is a time horizon, into a latent space $\mathcal H$ and a real-valued covariance function defined as $k:\mathcal H\times\mathcal H\to\mathbb R$, we have $\tilde k(\bar{\mathbf{x}}_i,\bar{\mathbf{x}}_j) \stackrel{\mathrm{def}}= k(\upxi(\bar{\mathbf{x}}_i),\upxi(\bar{\mathbf{x}}_j)), \bar{\mathbf{x}}_i,\bar{\mathbf{x}}_j\in\mathbb R^{QH}$. This definition guarantees that $\tilde k$ is a well defined covariance function. So GP-LSTM does not try to connect GPs in a recurrent fashion like DRGP-Nyström, but rather makes the model deep by using an arbitrary deterministic transformation $\upxi$, the LSTM. For scalability semi-stochastic optimization is used, as well as an existing algebraic structure of the kernels. The structure decomposes the relevant covariance matrices into Kronecker products of circulant matrices. Overall, this results in a training time $\mathcal O(N)$ per iteration and $\mathcal O(1)$ for test predictions. A graphical illustration of the GP-LSTM can be found in Figure \ref{fig:DRGPL}.
\section{Sparse Spectrum Gaussian Process and Variational Inference}
\label{sec:SparseSpectrumGaussianProcessandVariationalInference}
In the following sections we will recall the \textit{Sparse Spectrum Gaussian Process} (SSGP) first introduced by~\citep{quia2010sparse} and its improved version \textit{Variational Sparse Spectrum Gaussian Process} (VSSGP) by~\citep{gal2015improving}. Therefore, we first introduce SSGP and then have a closer look on the VI procedure in this context. We will not only introduce the VSSGP, but as new contribution we go a step backwards and also show how to variationally integrate out the input-space. In the case of simple GPR the input-data is given and it makes no sense to integrate out these variables, but in the layers of DRGPs we can profit from the fact that the inputs/outputs are actually latent variables. This allows us to propagate uncertainty through the layers, as with every variational input variable we have a related variational variance variable. Furthermore, this VI procedure is also applicable for the \textit{latent variable model} (LVM) in its special form introduced in~\citep{titsias2010bayesian}, where a reduced representation of high-dimensional data is searched for.

\subsection{Sparse Spectrum Gaussian Process}
\label{sec:SparseSpectrumGaussianProcess}
We introduce the SSGP following~\citep{gal2015improving} for the 1-dimensional output case. For a stationary covariance function $k$ on $\mathbb R^Q\times \mathbb R^Q$, that is a translation invariant positive definite kernel function, there exists a function\linebreak $\uprho :\mathbb R^Q\to\mathbb R,\boldsymbol\tau\mapsto \uprho(\boldsymbol\tau)$, such that $k(\mathbf{x},\mathbf{x}') =\uprho(\mathbf{x}-\mathbf{x'})$ for $\mathbf{x},\mathbf{x'}\in\mathbb R^Q$. Bochner’s theorem states that any stationary covariance function $k$ can be represented as the Fourier transform of a positive finite measure~\citep{stein2012interpolation}. Further, the positive finite measure $\boldsymbol\upmu$ is then \textit{proportional} to a probability measure $\mathrm{P}_{\boldsymbol{z}}$. If this probability measure has a density $p_{\boldsymbol{z}}$, there exists a power spectral density $\mathrm{S}$ of the measure $\boldsymbol\upmu$, which is also proportional to $p_{\boldsymbol{z}}$. The proportionality constant $\upsigma_{\text{power}}^2$ can then be derived as $\upsigma_{\text{power}}^2 = \uprho(\boldsymbol 0)$, as
\begin{align*}
&\uprho(\boldsymbol 0) = \int\limits_{\mathbb{R}^Q}e^{2\pi i\mathbf{z}^T\mathbf{0}}d\boldsymbol\upmu(\mathbf{z}) =\int\limits_{\mathbb{R}^Q}\mathrm{S}(\mathbf{z})d\mathbf{z} = \upsigma_{\text{power}}^2\int\limits_{\mathbb{R}^Q}p_{\boldsymbol{z}}(\mathbf{z})d\mathbf{z} = \upsigma_{\text{power}}^2.
\end{align*}
Then $\uprho(\boldsymbol\tau)$, using $\boldsymbol\tau=\mathbf{x}-\mathbf{x'}$, can be expressed as
\begin{align*}
\uprho(\boldsymbol\tau) &= \int\limits_{\mathbb{R}^Q} e^{2\pi i \mathbf{z}^T\boldsymbol\tau}d\boldsymbol\upmu(\mathbf{z})\\
& = \int\limits_{\mathbb{R}^Q} \mathrm{S}(\mathbf{z}) e^{2\pi i\mathbf{z}^T(\mathbf{x}-\mathbf{x}')}d\mathbf{z}\\
& = \upsigma_{\text{power}}^2\int\limits_{\mathbb{R}^Q} p_{\boldsymbol{z}}(\mathbf{z})e^{2\pi i\mathbf{z}^T\mathbf{x}}\left(e^{2\pi i\mathbf{z}^T\mathbf{x}'}\right)^*d\mathbf{z}\\
& \approx  \frac{\upsigma_{\text{power}}^2}{2M}\sum\limits_{m=1}^M\left(e^{2\pi i\mathbf{z}_m^T\mathbf{x}}\left(e^{2\pi i\mathbf{z}_m^T\mathbf{x}'}\right)^*+\left(e^{2\pi i\mathbf{z}_m^T\mathbf{x}}\right)^*e^{2\pi i\mathbf{z}_m^T\mathbf{x}'}\right)\numberthis\label{montecarlo1}\\
& = \frac{\upsigma_{\text{power}}^2}{M}\sum\limits_{m=1}^M\cos(2\pi \mathbf{z}_m^T(\mathbf{x}-\mathbf{x}'))\numberthis\label{next1}
\end{align*}
In \eqref{montecarlo1} an approximate integration is used sampling $M$ frequencies from $p_{\boldsymbol{z}}$ as a pair $\{\mathbf{z}_m,-\mathbf{z}_m\}$. As we use real valued covariance functions, the density is symmetric and we have a valid \textit{Monte Carlo Approximation} (MCA). This expression, by introducing $\mathbf{u}_m\in\mathbb R^Q$, $m=1,\dots,M$, can further be changed to
\begin{align*}
\eqref{next1} & = \frac{\upsigma_{\text{power}}^2}{M}\sum\limits_{m=1}^M\cos(2\pi \mathbf{z}_m^T((\mathbf{x}-\mathbf{u}_m)-(\mathbf{x}'-\mathbf{u}_m)),\\
& = \frac{\upsigma_{\text{power}}^2}{M}\sum\limits_{m=1}^M\int\limits_0^{2\pi}\frac{1}{2\pi}\sqrt{2}\cos(2\pi \mathbf{z}_m^T(\mathbf{x}-\mathbf{u}_m)+\mathrm{b})\sqrt{2}\numberthis\label{identity}\\
&\cos(2\pi \mathbf{z}_m^T(\mathbf{x}'-\mathbf{u}_m)+\mathrm{b})d\mathrm{b}\\
& \approx \frac{\upsigma_{\text{power}}^2}{M}\sum\limits_{m=1}^M\sqrt{2}\cos(2\pi \mathbf{z}_m^T(\mathbf{x}-\mathbf{u}_m)+\mathrm{b}_m)\sqrt{2}\numberthis\label{montecarlo2}\\
&\cos(2\pi \mathbf{z}_m^T(\mathbf{x}'-\mathbf{u}_m)+\mathrm{b}_m)\\
& \stackrel{\mathrm{def}}= \tilde k(\mathbf{x},\mathbf{x}').\numberthis\label{montecarlo3}
\end{align*}
In \eqref{montecarlo2} again a MCA is used sampling $b\sim\text{Unif}\left[0,2\pi\right]$ (uniform distribution) with one sample $\mathrm{b}_m$ per integral, while in \eqref{identity} a special identity, which can be found in \citep[Appendix A,~Identity 1]{gal2015improving}. We refer to $\mathbf{z}_m$ as the spectral points, $\mathrm{b}_m$ as the spectral phases and $\mathbf{u}_m$ as the pseudo-input points for\linebreak $m=1,\dots, M$.  The expression $\tilde K_{NN}\stackrel{\mathrm{def}}=(\tilde k(\mathbf{x}_i,\mathbf{x}_j))_{i,j=1}^N$ can be expressed as $\tilde K_{NN} = \tilde\Phi\tilde\Phi^T$, where $\tilde\Phi \stackrel{\mathrm{def}}= \left[\tilde\phi(\mathbf{x}_1,\text{\textbf{\textomega}}),\dots,\tilde\phi(\mathbf{x}_N,\text{\textbf{\textomega}})\right]^T\in\mathbb R^{N\times M}$ and
\begin{align*}
\tilde\phi(\mathbf{x},\text{\textbf{\textomega}})\stackrel{\mathrm{def}}= \sqrt{\frac{2\upsigma_{\text{power}}^2}{M}}\left[\cos(2\pi \mathbf{z}_1^T(\mathbf{x}-\mathbf{u}_1)+\mathrm{b}_1),\dots,\right.\numberthis\label{firstappr}\\
\left.\cos(2\pi \mathbf{z}_M^T(\mathbf{x}-\mathbf{u}_M)+\mathrm{b}_M)\right]^T\in\mathbb R^{M},
\end{align*}
as well as
\begin{align*}
\text{\textbf{\textomega}} \stackrel{\mathrm{def}}= [\text{\textbf{\textomega}}_1,\dots,\text{\textbf{\textomega}}_M]^T\stackrel{\mathrm{def}}= \begin{bmatrix}\begin{bmatrix}\mathbf{z}_1\\\mathrm{b}_1\end{bmatrix},\dots,\begin{bmatrix}\mathbf{z}_M\\\mathrm{b}_M\end{bmatrix}\end{bmatrix}^T\in\mathbb R^{M\times Q+1}.
\end{align*}
The definition\footnote{\label{note1}~\citep{gal2015improving} mentioned that this definition gives a better approximation than \eqref{next1}, which was used in~\citep{quia2010sparse}.} in \eqref{montecarlo3} is a sparse approximation of the given covariance function $k$. It results in a low-rank approximation for the corresponding covariance matrix $K_{NN}$ if $ M \ll N$, given the set of input-data $\mathbf{X} $. The density $p_{\boldsymbol{z}}$ is defined through the dual relation to the given covariance function through Bochner's Theorem.\\
In preview of our experiments in Section \ref{sec:experiments} and the following sections we choose a specific probability density like in~\citep[Proposition 2]{gal2015improving}, which approximates the \textit{spectral mixture} (SM) covariance function
\begin{align*}
k(\mathbf{x},\mathbf{x}')=\upsigma_{\text{power}}^2 \left(\prod_{q=1}^Q e^{-\frac{1}{2}\frac{(\mathrm{x}_q-\mathrm{x}_q')^2}{\mathrm{l}_q^2}}\right)\cos\left(2\pi\sum_{q=1}^Q\frac{(\mathrm{x}_q-\mathrm{x}_q')}{\mathrm{p}_q}\right),\numberthis\label{SMSS}
\end{align*}
with length scales $\mathrm{l}_q$, $\mathrm{p}_q\in\mathbb R$, $q=1,\dots, Q$.\\
Similar derivation as in the Equations \eqref{montecarlo1}-\eqref{montecarlo3} yield an additional scaling matrix $\mathfrak{L} \stackrel{\mathrm{def}}= \text{diag}([2\pi \mathrm{l}_q]_{q=1}^Q)\in\mathbb R^{Q\times Q}$ and a scaling vector $\mathbf{p} \stackrel{\mathrm{def}}= [\mathrm{p}_1^{-1},\dots,\mathrm{p}_Q^{-1}]^T\in\mathbb R^{Q}$ in Definition \eqref{firstappr}~\citep[Proposition 2]{gal2015improving}, which is
\begin{align*}
\phi(\mathbf{x},\text{\textbf{\textomega}}) \stackrel{\mathrm{def}} = \sqrt{\frac{2\upsigma_{\text{power}}^2}{M}}\left[\cos(2\pi(\mathfrak{L}^{-1}\mathbf{z}_1+\mathbf{p})^T(\mathbf{x}-\mathbf{u}_1) + \mathrm{b}_1),\dots,\right.\\
\left.\cos(2\pi(\mathfrak{L}^{-1}\mathbf{z}_M+\mathbf{p})^T(\mathbf{x}-\mathbf{u}_M) + \mathrm{b}_M)\right]^T\in\mathbb R^{M},
\end{align*}
again sampling $b\sim\text{Unif}\left[0,2\pi\right]$ and \textit{now} $\boldsymbol{z}\sim\mathcal{N}(\mathbf{0},I_Q)$, where $K_{NN}^{(\text{\tiny{SM}})} \stackrel{\mathrm{def}}= \Phi\Phi^T$ with $\Phi \stackrel{\mathrm{def}}= \left[\phi(\mathbf{x}_1,\text{\textbf{\textomega}}),\dots,\phi(\mathbf{x}_N,\text{\textbf{\textomega}})\right]^T\in\mathbb R^{N\times M}$. As $p_q \to \infty$, this reproduces the sparse approximation for the \textit{squared exponential} (SE) covariance function in the limit.\\
Note that the method for the SSGP case is a general one, as we do not have to take care about from which distribution to sample to receive tractable terms. For the VSSGP introduced in the next section we have to take care of the tractability of a KL-term for training. Nevertheless, an initialization of the spectral-points in SSGP and VSSGP other than $\mathbf{z}\sim\mathcal{N}(\mathbf{0},I_Q)$ can directly be interpreted as approximating other covariance functions as well. As in our methods the spectral-points are included in the training process, this can be interpreted as searching for the best approximate covariance function, based on an initial guess, the initialization sampled from $p_{\mathbf{z}}$.

\subsection{Variational Inference for VSSGP}
\label{sec:VariationalInferencerevisited}
In~\citep{gal2015improving} the SSGP was improved to VSSGP by variationally integrating out the spectral points and instead of optimizing the spectral points additionally optimizing the variational parameters. VI (general methodology can be found in the Appendix \ref{sec:VariationalInferencemethodology}) is a procedure to approximate a distribution, here the true posterior distribution, which is analytically intractable, by a variational distribution, which is easy to calculate. The log ML is also replaced by an approximation, a lower bound. We follow the scheme of~\citep[Chapter 4]{gal2015improving} for the multi-output case ${\mathbf{y}_i}\in\mathbb R^D$,  and the set ${\mathbf{Y}}=[\mathbf{y}^1,\dots,\mathbf{y}^D]\in\mathbb R^{N\times D}$, where $\mathbf{y}^d\in\mathbb R^N, d=1,\dots,D$, are now the data-vectors separated for the output dimensions $d$. By replacing the covariance function for each output component with the sparse SM covariance function and setting the prior
\begin{align*}
p_{\boldsymbol\omega}\stackrel{\mathrm{def}}=\prod_{m=1}^M p_{\boldsymbol{z}_m}p_{b_m},\text{ where }\boldsymbol{z}_m \sim  \mathcal{N}(\mathbf{0},I_Q),\quad b_m \sim \text{Unif}\left[0,2\pi\right]\numberthis\label{prior1}
\end{align*}
for $m=1,\dots,M$, where the product of the densities is defined as the point-wise product of functions, derived from the assumptions in the MCAs in \eqref{montecarlo1}, \eqref{montecarlo2}, we come to a form of the ML, which is
\begin{align*}
p_{\boldsymbol{Y}|\boldsymbol{X}}(\mathbf{Y}|\mathbf{X}) = \int\limits p_{\boldsymbol{Y}|\boldsymbol\omega,\boldsymbol{X}}(\mathbf{Y}|\text{\textbf{\textomega}},\mathbf{X})p_{\boldsymbol\omega}(\text{\textbf{\textomega}})d\text{\textbf{\textomega}}.
\end{align*}
Here we have $\boldsymbol{Y}|\boldsymbol\omega,\boldsymbol{X}\sim\mathcal{N}(\mathbf{0},K_{NN}^{(\text{\tiny{SM}})}+\upsigma_{\text{noise}}^2 I_N)$. Using an identity~\citep[Equations 2.113 - 2.115]{bishop2006pattern} like in~\citep{gal2015improving} and introducing new \linebreak variables $\mathbf{A} = [\mathbf{a}_1,\dots,\mathbf{a}_D] \in\mathbb R^{M\times D}$ with the prior assumptions 
\begin{align*}
p_{\boldsymbol{A}}\stackrel{\mathrm{def}}=\prod_{d=1}^D p_{\boldsymbol{a}_d},\text{ where } \boldsymbol{a}_d\sim\mathcal{N}(\mathbf{0},I_M),\numberthis\label{prior2}
\end{align*}
for $d=1,\dots,D,$ and $p_{\boldsymbol{Y}|\boldsymbol{A},\boldsymbol\omega,\boldsymbol{X}}=\prod\limits_{d=1}^D p_{\boldsymbol{y}^d|\boldsymbol{a}_d,\boldsymbol\omega,\boldsymbol{X}}$, where we have $\boldsymbol{y}^d|\boldsymbol{a}_d,\boldsymbol\omega,\boldsymbol{X}\sim\mathcal{N}(\Phi \mathbf{a}_d,\upsigma_{\text{noise}}^2 I_N)$, we can expand the previous equation to
\begin{align*}
p_{\boldsymbol{Y}|\boldsymbol{X}}(\mathbf{Y}|\mathbf{X}) = \int\limits p_{\boldsymbol{Y}|\boldsymbol{A},\boldsymbol\omega,\boldsymbol{X}}(\mathbf{Y}|\mathbf{A},\text{\textbf{\textomega}},\mathbf{X})p_{\boldsymbol{A}}(\mathbf{A})p_{\boldsymbol\omega}(\text{\textbf{\textomega}})d\mathbf{A} d\text{\textbf{\textomega}}.
\end{align*}
This eliminates the inverse $(K_{NN}^{(\text{\tiny{SM}})}+\upsigma_{\text{noise}}^2 I_N)^{-1}$ in $p_{\boldsymbol{Y}|\boldsymbol\omega,\boldsymbol{X}}$.\\
Now, to improve the SSGP to VSSGP, variational distributions are introduced in terms of
\begin{align}
q_{\boldsymbol\omega}&\stackrel{\mathrm{def}}=\prod_{m=1}^M q_{\mathbf{z}_m}q_{b_m},\text{ where }\boldsymbol{z}_m\sim\mathcal{N}(\boldsymbol\upalpha_m,\boldsymbol\upbeta_m),\quad b_m\sim\text{Unif}[\upgamma_m,\updelta_m],\numberthis\label{var1}
\end{align}
$\boldsymbol\upbeta_m\in\mathbb R^{Q\times Q}$ diagonal, $0 \leq \upgamma_m \leq \updelta_m \leq 2\pi$, for $m=1,\dots,M$, and 
\begin{align}
q_{\boldsymbol{A}}&\stackrel{\mathrm{def}}=\prod_{d=1}^D q_{\boldsymbol{a}_d},\text{ where }\boldsymbol{a}_d\sim\mathcal{N}(\mathbf{m}_d,\mathbf{s}_d),\numberthis\label{var2}
\end{align}
$\mathbf{s}_d\in\mathbb R^{M\times M}$ diagonal, for $d=1,\dots,D$, we can now use the VI procedure to come to the approximate models with different lower bounds to the log ML introduced by~\citep{gal2015improving}.\\
With $\mathcal{G}_{\text{\tiny{VSS}}} \stackrel{\mathrm{def}}= \log(p_{\boldsymbol{Y}|\boldsymbol{A},\boldsymbol\omega,\mathbf{X}}(\mathbf{Y}|\mathbf{A},\text{\textbf{\textomega}},\mathbf{X}))=\sum\limits_{d=1}^D\log(p_{\boldsymbol{y}_d|\boldsymbol{a}_d,\boldsymbol{X},\boldsymbol\omega}(\mathbf{y}_d|\mathbf{a}_d,\mathbf{X},\text{\textbf{\textomega}}))$,\linebreak $Q_{\text{\tiny{VSS}}} \stackrel{\mathrm{def}}=q_{\boldsymbol{A}}q_{\boldsymbol\omega}$ and $\mathbf{KL}$ the Kullback-Leibler divergence (see Appendix \ref{sec:KullbackLeiblerdivergence}), one of the lower bounds is
\begin{align}
\log(p_{\boldsymbol{Y}|\boldsymbol{X}}(\mathbf{Y}|\mathbf{X})) & \geq \mathbf{E}[{\mathcal{G}}_{\text{\tiny{VSS}}}]_{Q_{\text{\tiny{VSS}}}} - \mathbf{KL}(q_{\boldsymbol{A}}||p_{\boldsymbol{A}}) - \mathbf{KL}(q_{\boldsymbol\omega}||p_{\boldsymbol\omega})\numberthis\label{identity00}.
\end{align}
By proving the existence of an optimal distribution for $\boldsymbol{A}$ in the sense of a functional local optimum of the bound and filling the optimal expectation and variance into the bound above, one gets the other one. For details we refer to~\citep[Section 4.1, Equations 10, 11]{gal2015improving}.

\subsection{Variational Inference for DGP}
\label{sec:VariationalInferenceforDGP}
As a first contribution of this article, in this section we will go a step backwards and aim to marginalize also the input space. To prevent misunderstanding, we will write from now on $\mathbf{H} = [\mathbf{h}_1,\dots,\mathbf{h}_N]^T\in\mathbb R^{N\times Q}$ instead of $\mathbf{X}=[\mathbf{x}_1,\dots,\mathbf{x}_N]^T\in\mathbb R^{N\times Q}$, as we want to signify, that $\mathbf{X}$ in the previous Section \ref{sec:VariationalInferencerevisited} represented the set of measured input data. $\mathbf{H}$ is now a set of latent variables (later on representing the hidden states). This extra step will make it possible to propagate uncertainty between the hidden layers of a DGP, as we gain an extra variance parameter for the inputs. Therefore, we introduce a prior
\begin{align}
p_{\boldsymbol{H}} \stackrel{\mathrm{def}}= \prod\limits_{i=1}^N p_{\boldsymbol{h}_i},\text{ where }\boldsymbol{h}_i\sim\mathcal{N}(\mathbf{0},I_Q),\numberthis\label{prior3}
\end{align}
for $i=1,\dots,N$, and with the joint probability model
\begin{align*}
p_{\boldsymbol{Y},\boldsymbol{A},\boldsymbol\omega,\boldsymbol{H}}(\mathbf{Y},\mathbf{A},\text{\textbf{\textomega}},\mathbf{H}) = p_{\boldsymbol{Y}|\boldsymbol{A},\boldsymbol\omega,\boldsymbol{H}}(\mathbf{Y}|\mathbf{A},\text{\textbf{\textomega}},\mathbf{H})p_{\boldsymbol{A}}(\mathbf{A})p_{\boldsymbol\omega}(\text{\textbf{\textomega}})p_{\boldsymbol{H}}(\mathbf{H})
\end{align*}
we get
\begin{align}
p_{\boldsymbol{Y}}(\mathbf{Y})&= \int\limits p_{\boldsymbol{Y},\boldsymbol{A},\boldsymbol\omega,\boldsymbol{H}}(\mathbf{Y},\mathbf{A},\text{\textbf{\textomega}},\mathbf{H})d\mathbf{A}d\text{\textbf{\textomega}}d\mathbf{H}\numberthis\label{identity000}\\
&= \int\limits p_{\boldsymbol{Y}|\boldsymbol{A},\boldsymbol\omega,\boldsymbol{H}}(\mathbf{Y}|\mathbf{A},\text{\textbf{\textomega}},\mathbf{H})p_{\boldsymbol{A}}(\mathbf{A})p_{\boldsymbol\omega}(\text{\textbf{\textomega}})p_{\boldsymbol{H}}(\mathbf{H})d\mathbf{A}d\text{\textbf{\textomega}}d\mathbf{H} ,\numberthis\label{identity0}
\end{align}
where $\log(p_{\boldsymbol{Y}|\boldsymbol{A},\boldsymbol\omega,\boldsymbol{H}}(\mathbf{Y}|\mathbf{A},\text{\textbf{\textomega}},\mathbf{H}))=\sum\limits_{d=1}^D\log(p_{\boldsymbol{y}^d|\boldsymbol{a}_d,\boldsymbol\omega,\boldsymbol{H}}(\mathbf{y}^d|\mathbf{a}_d,\text{\textbf{\textomega}},\mathbf{H}))$ and where we also have $\boldsymbol{y}^d|\boldsymbol{a}_d,\boldsymbol\omega,\boldsymbol{H}\sim\mathcal{N}(\Phi \mathbf{a}_d,\upsigma_{\text{noise}}^2 I_N)$. This is also the ML of the LVM. Introducing variational distributions in terms of 
\begin{align}
q_{\boldsymbol{H}} \stackrel{\mathrm{def}}= \prod\limits_{i=1}^N q_{\boldsymbol{h}_i},\text{ where }{\boldsymbol{h}_i} \sim \mathcal{N}(\boldsymbol\upmu_i,\boldsymbol\uplambda_i),\numberthis\label{identity11}
\end{align}
$\boldsymbol\uplambda_i\in\mathbb R^{Q\times Q}$ diagonal, for $i=1,\dots,N$, and $q_{\boldsymbol{A}}$, $q_{\boldsymbol\omega}$ as in \eqref{var1}, \eqref{var2}, we can now use the VI procedure to variationally integrate out $\mathbf{A}, \text{\textbf{\textomega}}$ and $\mathbf{H}$.\\
The resulting statistics $\Psi_1 = \mathbf{E}\left[\Phi\right]_{q_{\boldsymbol\omega}q_{\boldsymbol{H}}}\in\mathbb R^{N\times M}$ and $\Psi_2 = \mathbf{E}\left[\Phi^T\Phi\right]_{q_{\boldsymbol\omega}q_{\boldsymbol{H}}}\in\mathbb R^{M\times M}$ are calculated as 
\begin{align*}
(\Psi_1)_{nm} & = \mathfrak{S}_m^1 \mathrm{Z}_{nm}e^{-\frac{1}{2}\hat{\boldsymbol\upalpha}_m^T \mathrm{C}_{nm}\hat{\boldsymbol\upalpha}_m}\cos(\hat{\boldsymbol\alpha}_m^T(\mathbf{c}_{nm}-\mathbf{u}_m)+\mathrm{b}_m),
\end{align*}
for $m, = 1,\dots, M$, $n, = 1,\dots, N$ and $\Psi_2 = \sum\limits_{n=1}^N \Psi_2^n$ with
\begin{align*}
(\Psi_2^n)_{mm'} &= \Sigma_{mm'}^2\mathrm{Z}_{mm'}^n \left(e^{-\frac{1}{2}\bar{\boldsymbol\upalpha}_{mm'}^T {\mathrm{D}_{mm'}^n}\bar{\boldsymbol\upalpha}_{mm'}}\cos(\bar{\boldsymbol\upalpha}_{mm'}^T\mathbf{d}_{mm'}^n - \bar{\uptau}_{mm'} + \bar{\mathrm{b}}_{mm'})\right.\\
& \left. + e^{-\frac{1}{2}\overset{+}{\boldsymbol\upalpha}_{mm'}^T {\mathrm{D}_{mm'}^n}\overset{+}{\boldsymbol\upalpha}_{mm'}}\cos(\overset{+}{\boldsymbol\upalpha}_{mm'}^T\mathbf{d}_{mm'}^n + \overset{+}{\uptau}_{mm'} +  \overset{+}{\mathrm{b}}_{mm'})\right),
\end{align*}
for $m,m' = 1,\dots, M$, $m\neq m'$, and 
\begin{align*}
(\Psi_2^n)_{mm'} &= \frac{\upsigma_{\text{power}}^2}{M} \left(1+\tilde{\Sigma}_m^2 \tilde{\mathrm{Z}}_{nm}e^{-2\hat{\boldsymbol\upalpha}_m^T\tilde{\mathrm{C}}_{nm}\hat{\boldsymbol\upalpha}_m}\cos(2(\hat{\boldsymbol\upalpha}_m^T(\tilde{\mathbf{c}}_{nm}-\mathbf{u}_m)+\mathrm{b}_m))\right),
\end{align*}
for $m,m' = 1,\dots, M$, $m=m'$, where we made the same simplification as in\linebreak ~\citep{gal2015improving}, in particular randomizing the phases following the MCA and integrating over all other variational parameters. We denote this version DRGP-VSS. For details on the appearing variables and the derivation, as well as the general form, we refer to the Appendix \ref{sec:Computationofthestatistics}.\\
If we skip the variational approximation over the spectral-points, introduced in Section \ref{sec:VariationalInferencerevisited}, and are just variational over $\mathbf{A}$, $\mathbf{H}$, we derive another simplification, which we call DRGP-SS. This one is derived with the statistics
\begin{align*}
(\Psi_1)_{nm} & = \sqrt{\frac{2\upsigma_{\text{power}}^2}{M}} e^{-\frac{1}{2}\hat{\mathbf{z}}_m^T\boldsymbol\uplambda_n\hat{\mathbf{z}}_m}\cos(\hat{\mathbf{z}}_m^T(\boldsymbol\upmu_n-\mathbf{u}_m)+\mathrm{b}_m)
\end{align*}
for $m, = 1,\dots, M$, $n, = 1,\dots, N$ and $\Psi_2 = \sum\limits_{n=1}^N \Psi_2^n$ with
\begin{align*}
(\Psi_2^n)_{mm'} & = \frac{\upsigma_{\text{power}}^2}{M}\left(e^{-\frac{1}{2}\bar{\mathbf{z}}_{mm'}^T\boldsymbol\uplambda_n\bar{\mathbf{z}}_{mm'}}\cos(\bar{\mathbf{z}}_{mm'}^T\boldsymbol\upmu_n - \bar{\uprho}_{mm'} + \bar{\mathrm{b}}_{mm'})\right.\\
& \left. + e^{-\frac{1}{2}\overset{+}{\mathbf{z}}_{mm'}^T\boldsymbol\uplambda_n\overset{+}{\mathbf{z}}_{mm'}}\cos(\overset{+}{\mathbf{z}}_{mm'}^T\boldsymbol\upmu_n + \overset{+}{\uprho}_{mm'} + \overset{+}{\mathrm{b}}_{mm'}))\right),
\end{align*}
for $m,m' = 1,\dots, M$, where we also refer to the Appendix \ref{sec:Computationofthestatistics} for the details.\\
We can state the following Proposition (Proof can be found in Appendix \ref{sec:Variationalbound}).\\
\\
{\bf Proposition 1.} {\it Let 
\begin{align*}
p_{\boldsymbol{Y}}(\mathbf{Y})&= \int\limits p_{\boldsymbol{Y},\boldsymbol{A},\boldsymbol\omega,\boldsymbol{H}}(\mathbf{Y},\mathbf{A},\text{\textbf{\textomega}},\mathbf{H})d\mathbf{A}d\text{\textbf{\textomega}}d\mathbf{H}\\
&= \int\limits p_{\boldsymbol{Y}|\boldsymbol{A},\boldsymbol\omega,\boldsymbol{H}}(\mathbf{Y}|\mathbf{A},\text{\textbf{\textomega}},\mathbf{H})p_{\boldsymbol{A}}(\mathbf{A})p_{\boldsymbol\omega}(\text{\textbf{\textomega}})p_{\boldsymbol{H}}(\mathbf{H})d\mathbf{A}d\text{\textbf{\textomega}}d\mathbf{H} 
\end{align*}
be the ML of our model with sparse approximation SM as in Section \ref{sec:SparseSpectrumGaussianProcess}, priors \eqref{prior1}, \eqref{prior2}, \eqref{prior3} variational distributions \eqref{var1}, \eqref{var2}, \eqref{identity11} and $P_{\text{\tiny{DGP}}}\stackrel{\mathrm{def}}= p_{\boldsymbol{A}}p_{\boldsymbol\omega}p_{\boldsymbol{H}}$, $Q_{\text{\tiny{DGP}}} \stackrel{\mathrm{def}}=q_{\boldsymbol{A}}q_{\boldsymbol\omega}q_{\boldsymbol{H}}$. Then with the VI procedure, $\mathcal{G}_{\text{\tiny{DGP}}} \stackrel{\mathrm{def}}= \log(p_{\boldsymbol{Y}|\boldsymbol{A},\boldsymbol\omega,\boldsymbol{H}}(\mathbf{Y}|\mathbf{A},\text{\textbf{\textomega}},\mathbf{H}))$, $\mathbf{M} = [\mathbf{m}_1, \dots  ,\mathbf{m}_D]\in\mathbb R^{M\times D}$ and  $\mathbf{S} = \sum\limits_{i=1}^D \mathbf{s}_d$ we get the bound
\begin{align*} 
\log(p_{\boldsymbol{Y}}(\mathbf{Y}))&\geq \mathbf{E}[\mathcal{G}_{\text{\tiny{DGP}}}]_{Q_{\text{\tiny{DGP}}}} - \mathbf{KL}(Q_{\text{\tiny{DGP}}}||P_{\text{\tiny{DGP}}}),\numberthis\label{identity1}\\
& = -\frac{ND}{2}\log(2\pi \upsigma_{\text{noise}}^2) - \frac{\text{\textup{tr}}\left(\mathbf{Y}^T\mathbf{Y}\right)}{2\upsigma_{\text{noise}}^2}  ,\numberthis\label{identity2}\\
& + \frac{\text{\textup{tr}}\left(\mathbf{Y}^T\Psi_1 \mathbf{M}\right)}{\upsigma_{\text{noise}}^2}- \frac{\text{\textup{tr}}\left(\Psi_2(\mathbf{S} + \mathbf{M}\mathbf{M}^T)\right)}{2\upsigma_{\text{noise}}^2}\\
& - \mathbf{KL}(q_{\boldsymbol{A}}||p_{\boldsymbol{A}}) - \mathbf{KL}(q_{\boldsymbol\omega}||p_{\boldsymbol\omega})- \mathbf{KL}(q_{\boldsymbol{H}}||p_{\boldsymbol{H}})\\
&\stackrel{\mathrm{def}}=\mathbfcal{L}.
\end{align*}
}\\
In the DRGP-SS case $\mathbf{KL}(q_{\boldsymbol\omega}||p_{\boldsymbol\omega})$ vanishes in \eqref{identity1}, \eqref{identity2}.
The KL terms can be analytically calculated. The explicit expressions can be found in the Appendix \ref{sec:KullbackLeiblerdivergence}.
Here, the low-rank approximation $\mathbf{E}\left[\Phi\Phi^T\right]_{q_{\boldsymbol\omega}q_{\boldsymbol{H}}}$ for the covariance matrix $K_{NN}$ is computed from a non-stationary approximate covariance function even when $k$ is stationary, similar as~\citep[see Discussion 1.]{gal2015improving} pointed out. The derived bound for the general case is optimized with respect to
\begin{align*}
 \boldsymbol\Theta&=\{\boldsymbol\upmu_i,\boldsymbol\uplambda_i\}_{i=1}^N\cup\{\boldsymbol\upalpha_m,\boldsymbol\upbeta_m,\mathbf{u}_m,\updelta_m,\upgamma_m\}_{m=1}^M\cup\{(\mathbf{m}_d,\mathbf{s}_d)\}_{d=1}^D\\
&\cup\{\upsigma_{\text{power}},\upsigma_{\text{noise}},\mathfrak{L},\mathbf{p}\}.
\end{align*}
For DRGP-VSS we optimize the same parameters but replace $\{\updelta_m,\upgamma_m\}_{m=1}^M$ with $\{\mathrm{b}_m\}_{m=1}^M$ and for DRGP-SS we do the same as for DRGP-VSS but additionally replacing $\{\boldsymbol\upalpha_m,\boldsymbol\upbeta_m\}_{m=1}^M$ with $\{\mathbf{z}_m\}_{m=1}^M$. The product of probability densities $Q_{\text{\tiny{DGP}}}=q_{\boldsymbol{A}}q_{\boldsymbol\omega}q_{\boldsymbol{H}}$ acts as an approximation to the true posterior $p_{\boldsymbol{A},\boldsymbol\omega,\boldsymbol{H}|\boldsymbol{Y}}$.

\subsection{Optimal variational distribution over $\mathbf{A}$}
\label{sec:Optimalvariationaldistribution}
Similar as in~\citep{gal2015improving} we can derive analytically an optimal variational distribution $q^{\text{opt}}_{\boldsymbol{A}}$ for the variables $\boldsymbol{A}$ and the lower bound $\mathbfcal{L}$ in the sense of a local functional optima $\mathbfcal{L}(q^{\text{opt}}_{\boldsymbol{A}} )$, which simplifies the above optimization procedure by eliminating the variational parameters  $\mathbf{S}$ and $\mathbf{M}$. This comes with an additional cost of time complexity in the training compared to the previous lower bound. This will be explained in the next sections. For proving the optimality of $q^{\text{opt}}_{\boldsymbol{A}}$  we have to show that the matrix $\Psi_2$ is always positive definite. As $\Phi^T\Phi$ is positive definite and we integrate always over variables with given probability densities, which are positive for every argument, this can be seen as follows.
\begin{align*}
\mathbf{e}^T\Psi_2\mathbf{e}&= \mathbf{e}^T\mathbf{E}\left[\Phi^T\Phi\right]_{q_{\boldsymbol\omega}q_{\boldsymbol{H}}}\mathbf{e}\\
&= \mathbf{E}\left[\mathbf{e}^T\Phi^T\Phi\mathbf{e}\right]_{q_{\boldsymbol\omega}q_{\boldsymbol{H}}}\\
&= \int\underbrace{\mathbf{e}^T\Phi^T\Phi\mathbf{e}}_{> 0, \text{ for all } \mathbf{e}\in\mathbb R^M, \mathbf{e}\neq \mathbf{0}}\underbrace{q_{\boldsymbol\omega}(\text{\textbf{\textomega}})q_{\boldsymbol{H}}(\mathbf{H})}_{> 0}d\text{\textbf{\textomega}}d\mathbf{H}>0,\numberthis\label{optimal}
\end{align*}
$\text{for all } \mathbf{e}\in\mathbb R^M, \mathbf{e}\neq \mathbf{0}$ and $\eqref{optimal}=\mathbf{0},\text{ for }\mathbf{e}= \mathbf{0}$.
The optimal bound is stated in the next Proposition (Proof can be found in Appendix \ref{sec:Optimalvariationaldistributionproof}, which is a more detailed version of the proof of~\citep{gal2015improving}).\\
\\
{\bf Proposition 2.} {\it
Let
\begin{align*}
\mathbfcal{L}(\boldsymbol\Theta) & = -\frac{ND}{2}\log(2\pi \upsigma_{\text{noise}}^2) - \frac{\text{\textup{tr}}\left(\mathbf{Y}^T\mathbf{Y}\right)}{2\upsigma_{\text{noise}}^2} + \frac{\text{\textup{tr}}\left(\mathbf{Y}^T\Psi_1 \mathbf{M}\right)}{\upsigma_{\text{noise}}^2} - \frac{\text{\textup{tr}}\left(\Psi_2(\mathbf{S} + \mathbf{M}\mathbf{M}^T)\right)}{2\upsigma_{\text{noise}}^2}\\
& - \mathbf{KL}(q_{\boldsymbol{A}}||p_{\boldsymbol{A}}) - \mathbf{KL}(q_{\boldsymbol\omega}||p_{\boldsymbol\omega})- \mathbf{KL}(q_{\boldsymbol{H}}||p_{\boldsymbol{H}}),
\end{align*}
be the cost function for our approximate model, then for $\mathbfcal{L}$ and for $\boldsymbol{A}$ an optimal distribution in the sense of a local functional optima $\mathbfcal{L}(q^{\text{opt}}_{\boldsymbol{A}})$, with priors \eqref{prior1}, \eqref{prior2}, \eqref{prior3}, variational distributions \eqref{var1}, \eqref{var2}, \eqref{identity11} is given by
\begin{align*}
q^{\text{opt}}_{\boldsymbol{A}}\stackrel{\mathrm{def}}= \prod_{d=1}^{D}q_{\boldsymbol{a}_d},\text{ where }\boldsymbol{a}_d \sim\mathcal{N}(A^{-1}\Psi_1^T\mathbf{y}_d,\upsigma_{\text{noise}}^2 A^{-1}),
\end{align*}
with $A = \Psi_2 + \upsigma_{\text{noise}}^2 I_M$, for $d=1,\dots,D$.
}\\
\\
Finally, we come to the optimal cost function, which follows in the next Proposition (Proof can be found in Appendix \ref{sec:Optimallowerbound}).\\
\\
{\bf Proposition 3.} {\it The cost function
\begin{align*}
\mathbfcal{L}(\boldsymbol\Theta) & = -\frac{ND}{2}\log(2\pi \upsigma_{\text{noise}}^2) - \frac{\text{\textup{tr}}\left(\mathbf{Y}^T\mathbf{Y}\right)}{2\upsigma_{\text{noise}}^2} + \frac{\text{\textup{tr}}\left(\mathbf{Y}^T\Psi_1 \mathbf{M}\right)}{\upsigma_{\text{noise}}^2} - \frac{\text{\textup{tr}}\left(\Psi_2(\mathbf{S} + \mathbf{M}\mathbf{M}^T)\right)}{2\upsigma_{\text{noise}}^2}\\
& - \mathbf{KL}(q_{\boldsymbol{A}}||p_{\boldsymbol{A}}) - \mathbf{KL}(q_{\boldsymbol\omega}||p_{\boldsymbol\omega})- \mathbf{KL}(q_{\boldsymbol{H}}||p_{\boldsymbol{H}}),
\end{align*}
reduces to the following optimal cost function, when using Proposition 2.,
\begin{align*}
\mathbfcal{L}^{\text{opt}}(\boldsymbol\Theta^{\text{opt}}) & =-\frac{(N-M)D}{2}\log(\upsigma_{\text{noise}}^2)- \frac{ND}{2}\log(2\pi) \\
&- \frac{\text{\textup{tr}}\left(\mathbf{Y}^T\mathbf{Y}\right)}{2\upsigma_{\text{noise}}^2} + \frac{\text{\textup{tr}}\left(\mathbf{Y}^T\Psi_1 A^{-1}\Psi_1^T\mathbf{Y}\right)}{2\upsigma_{\text{noise}}^2} -\frac{D\log(\left|A^{-1}\right|)}{2}\\
&- \mathbf{KL}(q_{\boldsymbol\omega}||p_{\boldsymbol\omega})- \mathbf{KL}(q_{\boldsymbol{H}}||p_{\boldsymbol{H}}),
\end{align*}
where $\boldsymbol\Theta^{\text{opt}}$ is the reduced set of parameters to optimize in the training.
}\\
\\
$\mathbfcal{L}^{\text{opt}}$ for the general case is now optimized with respect to
\begin{align*}
\boldsymbol\Theta^{\text{opt}}=\{\boldsymbol\upmu_i,\boldsymbol\uplambda_i\}_{i=1}^N\cup\{\boldsymbol\upalpha_m,\boldsymbol\upbeta_m,\mathbf{u}_m,\updelta_m,\upgamma_m\}_{m=1}^M\cup\{\upsigma_{\text{power}},\upsigma_{\text{noise}},\mathfrak{L},\mathbf{p}\}.
\end{align*}
For DRGP-VSS we optimize the same parameters but replace $\{\updelta_m,\upgamma_m\}_{m=1}^M$ with $\{\mathrm{b}_m\}_{m=1}^M$ and for DRGP-SS we do the same as for DRGP-VSS, but additionally replace $\{\boldsymbol\upalpha_m,\boldsymbol\upbeta_m\}_{m=1}^M$ with $\{\mathbf{z}_m\}_{m=1}^M$ and canceling $\mathbf{KL}(q_{\boldsymbol\omega}||p_{\boldsymbol\omega})$.

\section{Deep Recurrent Gaussian Process with Variational Sparse Spectrum Approximation}
\label{sec:DeepRecurrentGaussianProcesswithVariationalSparseSpectrumApproximation}
In this section we want to link our newly derived Variational Sparse Spectrum Approximation with the DRGP framework introduced in~\citep{mattos2015recurrent}. Therefore, we show the exact derivation in the following Section \ref{sec:RecurrentVariationalBayesforVariationalSparseSpectrumGaussianProcess}. In Section \ref{sec:DistributedInferenceforoptimal} we focus on how to reduce the complexity for each iteration in the training in order to make it practical for application.

\subsection{Recurrent Variational Bayes (REVARB) with Variational Sparse Spectrum Approximation}
\label{sec:RecurrentVariationalBayesforVariationalSparseSpectrumGaussianProcess}
Choosing the same recurrent structure as defined in Section \ref{sec:DeepRecurrentGaussianProcesses} for DRGP-Nyström, direct inference is intractable, because it is not possible to get a closed analytical expression for the true posterior of $f^{(l)}$ for every layer or for the ML. As mentioned in the beginning, this motivates the variational framework, called REVARB framework introduced in~\citep{mattos2015recurrent}. We exactly follow \citep[Section~4]{mattos2015recurrent}, but instead of using the sparse Nyström approximation of~\citep{titsias2009variational,titsias2010bayesian}, we will use the one developed and defined\linebreak\clearpage\noindent in our last Section \ref{sec:VariationalInferenceforDGP}.
 For the remaining part of our article we introduce 
\begin{align*}
\mathbf{y}_{H_{\mathbf{x}}+1:}&= [y_{H_{\mathbf{x}} + 1},\dots,y_N] \in\mathbb{R}^{\hat{N}}\\
\mathbf{h}_{H_{\mathbf{x}}+1:}^{(l)}&= [\mathrm{h}_{H_{\mathbf{x}} + 1}^{(l)},\dots,\mathrm{h}_N^{(l)}] \in\mathbb{R}^{\hat{N}}\\
\mathbf{H}_L &= [\mathbf{h}^{(1)},\dots,\mathbf{h}^{(L)}]^T\in\mathbb R^{L\times \hat{N}+H_{\mathrm{h}}},\\
\tilde{\mathbf{H}}_L &= [\tilde{\mathbf{h}}^{(1)},\dots,\tilde{\mathbf{h}}^{(L)}]^T\in\mathbb R^{L\times H_{{\mathrm{h}}}},\\
\tilde{\mathbf{h}}^{(l)} &= [\mathrm{h}_{1+H_{\mathbf{x}}-H_{\mathrm{h}}}^{(l)},\dots,\mathrm{h}_{H_{\mathrm{h}}}^{(l)}]^T\in\mathbb R^{H_{\mathrm{h}}},\\
\mathbf{A}_L  &= [\mathbf{a}^{(1)},\dots,\mathbf{a}^{(L+1)}]^T\in\mathbb R^{L+1\times M},\\ \text{\textbf{\textomega}}_L  &= [\text{\textbf{\textomega}}^{(1)},\dots, \text{\textbf{\textomega}}^{(L+1)}]^T\in\mathbb R^{L+1\times M\times Q+1},\\\text{\textbf{\textomega}}^{(l)} &= \begin{bmatrix}\begin{bmatrix}\mathbf{z}_1^{(l)}\\\mathrm{b}_1^{(l)}\end{bmatrix},\dots,\begin{bmatrix}\mathbf{z}_M^{(l)}\\\mathrm{b}_M^{(l)}\end{bmatrix}\end{bmatrix}^T\in\mathbb R^{M\times Q+1}.
\end{align*}
for $l = 1,\dots,L+1$. In detail, considering a model with $L$ hidden layers and 1-dimensional outputs, the joint probability density of all the random variables conditioned on $\boldsymbol{X}$ is given by
\begin{align*}
& p_{\boldsymbol{y}_{H_{\mathbf{x}}+1:},\boldsymbol{a}^{(L+1)},\boldsymbol\omega^{(L+1)},\left[\boldsymbol{a}^{(l)},\boldsymbol\omega^{(l)},\boldsymbol{h}^{(l)}\right]_{l=1}^L|\boldsymbol{X}}\\
& =p_{\boldsymbol{y}_{H_{\mathbf{x}}+1:}|\boldsymbol{a}^{(L+1)},\boldsymbol\omega^{(L+1)},\hat{\boldsymbol{H}}^{(L+1)}}p_{\boldsymbol{a}^{(L+1)}}p_{\boldsymbol\omega^{(L+1)}}\numberthis\label{commond}\\
&\prod_{l=1}^{L}p_{\boldsymbol{h}^{(l)}_{H_{\mathbf{x}}+1:}|\boldsymbol{a}^{(l)},\boldsymbol\omega^{(l)},\hat{\boldsymbol{H}}^{(l)}}p_{\boldsymbol{a}^{(l)}}p_{\boldsymbol\omega^{(l)}}p_{\tilde{\boldsymbol{h}}^{(l)}}.
\end{align*}
The priors and the variational distributions are assumed in terms of
\begin{align*}
&P_{\text{\scalebox{.8}{\tiny{REVARB}}}}  \stackrel{\mathrm{def}}= p_{\tilde{\boldsymbol{H}}_L}p_{\boldsymbol{A}_L}p_{\boldsymbol\omega_L}, && Q_{\text{\scalebox{.8}{\tiny{REVARB}}}}  \stackrel{\mathrm{def}}= q_{\boldsymbol{H}_L}q_{\boldsymbol{A}_L}q_{\boldsymbol\omega_L},\numberthis\label{revarb}&\hfill
\end{align*}
where
\begin{align*}
&p_{\tilde{\boldsymbol{H}}_L}  = \prod\limits_{l=1}^{L} p_{\tilde{\boldsymbol{h}}^{(l)}},\quad\tilde{\boldsymbol{h}}^{(l)}\sim\mathcal{N}(\mathbf{0},I_{H_{\mathrm{h}}}),\\
&p_{\boldsymbol{A}_L} = \prod\limits_{l=1}^{L+1}p_{\boldsymbol{a}^{(l)}},\quad\boldsymbol{a}^{(l)}\sim\mathcal{N}(\mathbf{0},I_{M}),\hfill\\
&p_{\boldsymbol\omega_L}  = \prod\limits_{l=1}^{L+1}p_{\boldsymbol\omega^{(l)}}= \prod\limits_{l=1}^{L+1}\prod\limits_{m=1}^M p_{\boldsymbol{z}_m^{(l)}}p_{b_m^{(l)}},\\
&\boldsymbol{z}_m^{(l)}\sim \mathcal{N}(\mathbf{0},I_Q),\quad b_m^{(l)}\sim\text{Unif}\left[0,2\pi\right], \hfill\\
&q_{\boldsymbol{H}_L}  = \prod\limits_{l=1}^{L}\prod\limits_{i=1+H_{\mathbf{x}}-H_{\mathrm{h}}}^{N} q_{h_i^{(l)}},\quad h_i^{(l)}\sim\mathcal{N}(\upmu^{(l)}_i,\uplambda^{(l)}_i),\\
&q_{\boldsymbol{A}_L}=\prod\limits_{l=1}^{L+1}\prod\limits_{m=1}^{M}q_{a_m^{(l)}},\quad a_m^{(l)}\sim\mathcal{N}(\mathrm{m}^{(l)}_i,\mathrm{s}^{(l)}_i), \hfill\\
&q_{\boldsymbol\omega_L} = \prod\limits_{l=1}^{L+1}q_{\boldsymbol\omega^{(l)}}=\prod\limits_{l=1}^{L+1}\prod\limits_{m=1}^M q_{\boldsymbol{z}_m^{(l)}}q_{b_m^{(l)}},\\
&\boldsymbol{z}_m^{(l)} \sim\mathcal{N}(\mathbf{\upalpha}_m^{(l)},\mathbf{\upbeta}_m^{(l)}),\quad b_m^{(l)}\sim\text{Unif}[\upgamma_m^{(l)},\updelta_m^{(l)}],\hfill
\end{align*}
\\
where $\mathbf{\upbeta}_m^{(l)}\in \;\mathbb R^{Q\times Q}\;\text{is diagonal}$, for $i=1+H_{\mathbf{x}}-H_{\mathrm{h}},\dots,N$, $m=1\dots,M$, $l=1,\dots,L+1$. The next Proposition states the \textit{Recurrent Variational Bayes} (optimal) lower bound for \textit{(V)SS} (REVARB-(V)SS) which is optimized during the training of our models (Proof, exact expressions can be found in Appendix \ref{sec:REVARBboundforVSSGP}).\\
\\
{\bf Proposition 4.} {\it The REVARB-(V)SS bound $\mathbfcal{L}_{\text{\scalebox{.8}{\tiny{REVARB}}}}$ to optimize, with priors and variational distributions defined in \eqref{revarb}, and
\begin{align*}
\mathcal{G}_{\text{\scalebox{.8}{\tiny{REVARB}}}} &\stackrel{\mathrm{def}}= \log(p_{\boldsymbol{y}_{H_{\mathbf{x}}+1:}|\boldsymbol{a}^{(L+1)},\boldsymbol\omega^{(L+1)},\hat{\boldsymbol{H}}^{(L+1)}}(\mathbf{y}_{H_{\mathbf{x}}+1:}|\mathbf{a}^{(L+1)},\text{\textbf{\textomega}}^{(L+1)},\hat{\mathbf{H}}^{(L+1)}))\\
&+\sum_{l=1}^{L}\log(p_{\boldsymbol{h}^{(l)}_{H_{\mathbf{x}}+1:}|\boldsymbol{a}^{(l)},\boldsymbol\omega^{(l)},\hat{\boldsymbol{H}}^{(l)}}(\mathbf{h}^{(l)}_{H_{\mathbf{x}}+1:}|\mathbf{a}^{(l)},\text{\textbf{\textomega}}^{(l)},\hat{\mathbf{H}}^{(l)}))
\end{align*}
is
\begin{align*}
\log(p_{\boldsymbol{y}_{H_{\mathbf{x}}+1:}|\boldsymbol{X}}(\mathbf{y}_{H_{\mathbf{x}}+1:}|\mathbf{X})) \geq \mathbfcal{L}_{\text{\scalebox{.8}{\tiny{REVARB}}}}=\mathbf{E}[\mathcal{G}_{\text{\scalebox{.8}{\tiny{REVARB}}}}]_{Q_{\text{\scalebox{.8}{\tiny{REVARB}}}}} - \mathbf{KL}(Q_{\text{\scalebox{.8}{\tiny{REVARB}}}}||P_{\text{\scalebox{.8}{\tiny{REVARB}}}}).
\end{align*}
Additionally, the optimal bound $\mathbfcal{L}^{\text{opt}}_{\text{\scalebox{.8}{\tiny{REVARB}}}}$ can be obtained immediately analogous to Proposition 3. and the fact, that the bound decomposes into a sum of independent terms for $\boldsymbol{A}_L$.
}\\
\\
The optimal bound here is different to the variational bound derived in~\citep{cutajar2016practical}, who optimized the variational weights along with the other hyperparameters, did not propagate the uncertainty and did not assume a recurrent structure.
To overcome implementation issues, it is convenient to also introduce priors for $\boldsymbol{X}$ in \eqref{commond} with
\begin{align*}
p_{\boldsymbol{y}_{H_{\mathbf{x}}+1:}}(\mathbf{y}_{H_{\mathbf{x}}+1:}) &= \int\limits p_{\boldsymbol{y}_{H_{\mathbf{x}}+1:},\boldsymbol{X}}(\mathbf{y}_{H_{\mathbf{x}}+1:},\mathbf{X})d\mathbf{X}\\
 &= \int\limits p_{\boldsymbol{y}_{H_{\mathbf{x}}+1:}|\boldsymbol{X}}(\mathbf{y}_{H_{\mathbf{x}}+1:}|\mathbf{X})p_{\boldsymbol{X}}(\mathbf{X}) d\mathbf{X},
\end{align*}
and variational distributions $p_{\boldsymbol{x}_i}$, $q_{\boldsymbol{x}_i}$, where $\boldsymbol{x}_i\sim\mathcal{N}(\tilde{\boldsymbol\upmu}_i,\tilde{\boldsymbol\uplambda}_i)$, $i = 1,...,N$ in both cases. We can interpret these as Delta-series for $\tilde{\boldsymbol\uplambda}_i\to\mathbf{0}$ with $\tilde{\boldsymbol\upmu}_i=\mathbf{x}_i$, $\mathbf{x}_i$ the measured input-data for $i=1,\dots,N$. Then it is possible to use the statistics calculated in the previous Section \ref{sec:VariationalInferenceforDGP} as well for the input-data $\mathbf{X}$ by setting the mean equal to the measured input-data and letting the variance converge to zero. We refer to the Appendix \ref{sec:REVARBboundforVSSGP} for further details. The amount of parameters to optimize is $2NL$, as~\citep{mattos2015recurrent} pointed out.\\
The upcoming statistics for each layer, $\Psi_1^{(l)}$ and $\Psi_2^{(l)}$, are computed as in the previous Section \ref{sec:VariationalInferenceforDGP} with the structured input-data defined in the Equations \eqref{INPUT} and $\mathbf{X}$, $\{\boldsymbol\upmu_i^{(l)},\boldsymbol\uplambda_i^{(l)}\}_{i=1+H_{\mathbf{x}}-H_{\mathrm{h}}}^N$, $\{\boldsymbol\upalpha_m^{(l)},\boldsymbol\upbeta_m^{(l)},\mathbf{u}_m^{(l)},\updelta_m^{(l)},\upgamma_m^{(l)}\}_{m=1}^M$, $\{\upsigma_{\text{power}}^{(l)},\upsigma_{\text{noise}}^{(l)},\mathfrak{L}^{(l)},\mathbf{p}^{(l)}\}$ for $l=1,\dots,L+1$, for the general case. Again for DRGP-VVS, we replace $\{\updelta_m^{(l)},\upgamma_m^{(l)}\}_{m=1}^M$ with $\{\mathrm{b}_m^{(l)}\}_{m=1}^M$ and for DRGP-SS we do the same, but replace $\{\boldsymbol\upalpha_m^{(l)},\boldsymbol\upbeta_m^{(l)}\}_{m=1}^M$ with  $\{\text{\textbf{\textomega}}_m^{(l)}\}_{m=1}^M$.\\ Predictions for new $\hat{\mathbf{h}}_{*}^{(l)}$ in the REVARB-(V)SS framework with the approximate predictive distributions in terms of
\begin{align*}
q_{f^{(1)}_{\hat{\mathbf{h}}_{\ast}^{(1)}}|\overline{\boldsymbol{x}}_{*}}&= \int p_{f^{(1)}_{\hat{\mathbf{h}}_{\ast}^{(1)}}|\boldsymbol{a}^{(1)}, \boldsymbol\omega^{(1)}, \hat{\boldsymbol{h}}_{\ast}^{(1)}}(\text{\Large\LargerCdot}|\mathbf{a}^{(1)}, \text{\textbf{\textomega}}^{(1)}, \hat{\mathbf{h}}_{\ast}^{(1)})\numberthis\label{pred1}\\
&q_{\boldsymbol{a}^{(1)}}(\mathbf{a}^{(1)})q_{\boldsymbol\omega^{(1)}}(\text{\textbf{\textomega}}^{(1)})q_{\boldsymbol{h}^{(1)}}(\mathbf{h}^{(1)})q_{\boldsymbol{h}_{*}^{(1)}}(\mathbf{h}_{*}^{(1)})d\mathbf{a}^{(1)}d\text{\textbf{\textomega}}^{(1)}d\mathbf{h}^{(1)}d\mathbf{h}_{*}^{(1)},
\end{align*}
and 
\begin{align*}
q_{f^{(l)}_{\hat{\mathbf{h}}_{\ast}^{(l)}}}&= \int p_{f^{(l)}_{\hat{\mathbf{h}}_{\ast}^{(l)}}|\boldsymbol{a}^{(l)}, \boldsymbol\omega^{(l)}, \hat{\boldsymbol{h}}_{\ast}^{(l)}}({\text{\Large\LargerCdot}}| \mathbf{a}^{(l)}, \text{\textbf{\textomega}}^{(l)}, \hat{\mathbf{h}}_{\ast}^{(l)})\numberthis\label{pred2}\\
&q_{\boldsymbol{a}^{(l)}}(\mathbf{a}^{(l)})q_{\boldsymbol\omega^{(l)}}(\text{\textbf{\textomega}}^{(l)})q_{\boldsymbol{h}^{(l)}}(\mathbf{h}^{(l)})q_{\boldsymbol{h}^{(l-1)}}(\mathbf{h}^{(l-1)})\\
&q_{\boldsymbol{h}_{*}^{(l)}}(\mathbf{h}_{*}^{(l)})q_{\boldsymbol{h}_{*}^{(l-1)}}(\mathbf{h}_{*}^{(l-1)})d\mathbf{a}^{(l)}d\text{\textbf{\textomega}}^{(l)}d\mathbf{h}^{(l)}d\mathbf{h}^{(l-1)}d\mathbf{h}_{*}^{(l)}d\mathbf{h}_{*}^{(l-1)},
\end{align*}
for $2\leq l\leq L$, and
\begin{align*}
q_{f^{(L+1)}_{\hat{\mathbf{h}}_{\ast}^{(L+1)}}}& = \int p_{f^{(L+1)}_{\hat{\mathbf{h}}_{\ast}^{(L+1)}}|\boldsymbol{a}^{(L+1)}, \boldsymbol\omega^{(L+1)}, \hat{\boldsymbol{h}}_{\ast}^{(L+1)}}({\text{\Large\LargerCdot}}| \mathbf{a}^{(L+1)}, \text{\textbf{\textomega}}^{(L+1)}, \hat{\mathbf{h}}_{\ast}^{(L+1)})\numberthis\label{pred3}\\
&q_{\boldsymbol{a}^{(L+1)}}(\mathbf{a}^{(L+1)})q_{\boldsymbol\omega^{(L+1)}}(\text{\textbf{\textomega}}^{(L+1)})q_{\boldsymbol{h}^{(L)}}(\mathbf{h}^{(L)})\\
&q_{\boldsymbol{h}_{*}^{(L+1)}}(\mathbf{h}_{*}^{(L)})d\mathbf{a}^{(L+1)}d\text{\textbf{\textomega}}^{(L+1)}d\mathbf{h}^{(L)}d\mathbf{h}_{*}^{(L)},
\end{align*}
are performed iteratively with approximate uncertainty propagation between each layer. $q_{\boldsymbol{h}_{*}^{(l)}}$ is Gaussian with mean and variance derived from previous predictions $\mathbf{h}_{*}^{(l)}$ for $l=1,\dots,L$. We choose the abbreviation $q_{f^{(l)}_{\hat{\mathbf{h}}_{\ast}^{(l)}}}$ for all $l=1,\dots,L+1$ in the remaining part of our article. We introduce $\boldsymbol\upmu^{(l)}_{H_{\mathbf{x}}+1:}=[\upmu_{H_{\mathrm{h}} + 1},\dots,\upmu_N] \in\mathbb{R}^{\hat{N}}$ and $\Psi_{1\ast}^{(l)}$, $\Psi_{2\ast}^{(l)}$ means calculating the statistics for one new $\hat{\mathbf{h}}_{*}^{(l)}$. We derive the following Proposition (Proof can be found in Appendix \ref{sec:Meanandvarianceofthepredictivedistribution}).\\
\\
{\bf Proposition 5.} {\it Predictions for each layer $l$ and new $\hat{\mathbf{h}}_{*}^{(l)}$ are performed with
\begin{align*}
\mathbf{E}\left[f^{(l)}_{\hat{\mathbf{h}}_{\ast}^{(l)}}\right]_{q_{f^{(l)}_{\hat{\mathbf{h}}_{\ast}^{(l)}}}}  &= \Psi_{1\ast}^{(l)}\mathbf{m}^{(l)},\\
\mathbf{V}\left[f^{(l)}_{\hat{\mathbf{h}}_{\ast}^{(l)}}\right]_{q_{f^{(l)}_{\hat{\mathbf{h}}_{\ast}^{(l)}}}} & = \left(\textbf{m}^{(l)}\right)^T\left(\Psi_{2\ast}^{(l)}-\left(\Psi_{1\ast}^{(l)}\right)^T\Psi_{1\ast}^{(l)}\right)\textbf{m}^{(l)}+ \text{\textup{tr}}\left(\mathbf{s}^{(l)}\Psi_{2\ast}^{(l)}\right),
\end{align*}
where
\begin{align*}
\mathbf{m}^{(l)} \stackrel{\mathrm{opt}}= \left(A^{(l)}\right)^{-1}\left(\Psi_1^{(l)}\right)^T\boldsymbol\upmu^{(l)}_{H_{\mathbf{x}}+1:},\quad \mathbf{s}^{(l)}\stackrel{\mathrm{opt}}= \upsigma_{\text{noise}}^{(l)}\left(A^{(l)}\right)^{-1},
\end{align*}
for $1,\dots,L$, and fully analog for $l=L+1$ by replacing $\boldsymbol\upmu^{(l)}_{H_{\mathbf{x}}+1:}$ with $\mathbf{y}_{H_{\mathbf{x}}+1:}$.}\\
\\
To predict $\mathrm{h}_{\ast}^{(l)}$ or $\mathrm{y}_{\ast}$ , we calculate
\begin{align*}
\upmu_{*}^{(l)} = \mathbf{E}\left[f^{(l)}_{\hat{\mathbf{h}}_{\ast}^{(l)}}\right]_{q_{f^{(l)}_{\hat{\mathbf{h}}_{\ast}^{(l)}}}}\text{ and }\uplambda_{*}^{(l)}=\mathbf{V}\left[f^{(l)}_{\hat{\mathbf{h}}_{\ast}^{(l)}}\right]_{q_{f^{(l)}_{\hat{\mathbf{h}}_{\ast}^{(l)}}}} + \upsigma_{\text{noise}}^{(l)}.
\end{align*}
Here, $\upmu_{*}^{(l)}$ serves as an approximation for $\mathrm{h}_{\ast}^{(l)}$ in the case $l=1,\dots,L$ only, and because in the variational framework we only need $\upmu_{*}^{(l)}$ and the variance $\uplambda_{*}^{(l)}$ for the next layer's prediction $(l+1)$, we have all we need to predict through every layer. In the case $l=L+1$, we again replace $\upmu_{*}^{(l)}$ with $\mathrm{y}^{\ast}$ and $\uplambda_{*}^{(l)}$ with $(\upsigma^{*})^2$ (no approximation; writing $\mathbf{y}^{\ast}$ and $\Sigma^{*}$ for the sets of predictions). In Algorithm \ref{TrainingVSS} and \ref{PredictionVSS} we show the necessary steps for training and prediction of DRGP-VSS in the optimal bound case. For DRGP-SS all can be obtained analogously by replacing $\{\boldsymbol\upalpha_m,\boldsymbol\upbeta_m\}_{m=1}^M$ with $\{\mathbf{z}_m\}_{m=1}^M$.
\begin{algorithm}
\caption{DRGP-VSS Training} \label{TrainingVSS}
\algorithmicrequire{ Number of hidden layers $L$, wide of the time horizons $H_{\mathbf{x}}$, $H_{\mathrm{h}}$, initialization of the parameter set\\ $\boldsymbol\Theta^{\text{opt}}_0=\bigcup_{l=1}^{L+1}{\{\boldsymbol\upmu_i^{(l)},\boldsymbol\uplambda_i^{(l)}\}_{i=1+H_{\mathbf{x}}-H_{\mathrm{h}}}^N\cup\{\boldsymbol\upalpha_m^{(l)},\boldsymbol\upbeta_m^{(l)},\mathbf{u}_m^{(l)},\mathrm{b}_m^{(l)}\}_{m=1}^M\cup\{\upsigma_{\text{power}}^{(l)},\upsigma_{\text{noise}}^{(l)},\mathfrak{L}^{(l)},\mathbf{p}^{(l)}\}}$, set of input-data $\mathbf{X}$, set of output-data $\mathbf{y}$, amount of iterations $I_1$, $I_2$, $I_2>I_1$, optimizer}\\
\algorithmicensure{ Trained parameter-set $\boldsymbol\Theta^{\text{opt}}_I$}
    \begin{algorithmic}[1]
      \Function{Training}{$L$,$H_{\mathbf{x}}$,$H_{\mathrm{h}}$,$\mathbf{X}$,$\mathbf{y}$,$\boldsymbol\Theta^{\text{opt}}_0$,$I$}
				\For{$j = 0$ to $I_2$}\Comment{Optimization with l-bfgs or gradient descent}
					\If{$j<I_1$}
						\State Fix $\upsigma_{\text{power}}^{(l)},\upsigma_{\text{noise}}^{(l)}$ (optional $\boldsymbol\upalpha_m^{(l)}$, $\boldsymbol\upbeta_m^{(l)}$, $\mathbf{u}_m^{(l)}$) for the optimization-step
						\State to independently train the latent states $\boldsymbol\upmu^{(l)},\boldsymbol\uplambda^{(l)}\in\mathbb R^N$.	
					\EndIf
					\State Compute model structure with $L$,$H_{\mathbf{x}}$,$H_{\mathrm{h}}$,$\mathbf{X}$,$\mathbf{y}$,$\boldsymbol\Theta^{\text{opt}}_j$. \Comment{see $\hat{\mathbf{h}}_{*}^{(l)}$ in Section \ref{sec:DeepRecurrentGaussianProcesses}}
					\State Calculate gradients, approximate Hessian matrix of REVARB-(V)SS w.r.t $\boldsymbol\Theta^{\text{opt}}_j$
					\State and perform $j$-th optimization-step.
				\EndFor
				\State \textbf{return} $\boldsymbol\Theta^{\text{opt}}_I$
       \EndFunction
\end{algorithmic}
\end{algorithm}
\begin{algorithm}
\caption{DRGP-VSS Prediction}\label{PredictionVSS}
\algorithmicrequire{ Set of new input-data $\mathbf{X}_{\ast}$, trained $\boldsymbol\Theta^{\text{opt}}_I$, number of hidden layers $L$, size of time horizons $H_{\mathbf{x}}$, $H_{\mathrm{h}}$}\\
\algorithmicensure{ Set of predicted output values $\mathbf{y}^{\ast}$ and set of variance predictions $\Sigma^{*}$}
    \begin{algorithmic}[1]
      \Function{Prediction}{$\mathbf{X}_{\ast}$,$\boldsymbol\Theta^{\text{opt}}_I$,$L$,$H_{\mathbf{x}}$,$H_{\mathrm{h}}$}
				\For{$j = 1$ to $length(\mathbf{X}_{\ast})$}\\
					\Comment{Iterative prediction with uncertainty propagation between each layer}
					\State Compute the approximation of $\hat{\mathbf{h}}_{*}^{(1)}$\Comment{see $\hat{\mathbf{h}}_{*}^{(l)}$ in Section \ref{sec:DeepRecurrentGaussianProcesses}}  
					\State for given structure in training with $\textnormal{\fontfamily{phv}\selectfont\straighttheta}_I$,$L$,$H_{\mathbf{x}}$,$H_{\mathrm{h}}$,$\mathbf{X}_{*}$ and past predictions $\upmu_{*}^{(1)}$, $\uplambda_{*}^{(1)}$.
				  \State Perform prediction with $\hat{\mathbf{h}}_{*}^{(1)}$ and Proposition 5 for first layer.
					\For{$l = 2$ to $L+1$} 
						\State Compute the approximation of $\hat{\mathbf{h}}_{*}^{(l)}$ \Comment{see $\hat{\mathbf{h}}_{*}^{(l)}$ in Section \ref{sec:DeepRecurrentGaussianProcesses}} 
						\State for given structure in training with $\boldsymbol\Theta^{\text{opt}}_I$,$L$ $H_{\mathbf{x}},H_{\mathrm{h}}$ and past predictions
						\State $\upmu_{*}^{(l)}$,$\uplambda_{*}^{(l)}$,$\upmu_{*}^{(l-1)}$,$\uplambda_{*}^{(l-1)}$.
						\State Perform prediction with $\hat{\mathbf{h}}_{*}^{(l)}$ and Proposition 5 for the $l$-th layer.
					\EndFor
				\EndFor
				\State \textbf{return} $\mathbf{y}^{\ast}$, $\Sigma^{*}$
			\EndFunction
\end{algorithmic}
\end{algorithm}

\subsection{Distributed Inference for optimal REVARB-(V)SS}
\label{sec:DistributedInferenceforoptimal}

Calculating the \textit{optimal} REVARB-(V)SS requires $\mathcal O(NM^2Q_{\text{max}}L)$, where\linebreak $Q_{\text{max}} = \displaystyle\max_{l=1\dots,L+1}{Q^{(l)}}$, $Q^{(l)}\stackrel{\mathrm{def}}=\dim(\hat{\mathbf{h}_{i}}^{(l)})$ and $\hat{\mathbf{h}}_i^{(l)}$ is coming from the Equations \eqref{INPUT} for a fixed chosen $i$ and $l=1,\dots,L+1$. In this section we show, how we can reduce the complexity of inference in the optimal REVARB-(V)SS setting with distributed inference to $\mathcal O(M^3)$. Similar arguments hold for REVARB-(V)SS (non-optimal) with even more capability to reduce down the complexity.\\
 $\mathbfcal{L}^{\text{opt}}_{\text{\scalebox{.8}{\tiny{REVARB}}}}$ in Appendix \ref{sec:REVARBboundforVSSGP} Equation \eqref{proof3}, separated for each hidden layer and the output layer ($\boldsymbol\upmu_{H_{\mathbf{x}}+1:}$ replaced by $\mathbf{y}_{H_{\mathbf{x}}+1:}$), can be written as $\mathbfcal{L}^{\text{opt}}_{\text{\scalebox{.8}{\tiny{REVARB}}}}= \sum\limits_{l=1}^{L+1}B_l,$
where we ignore the KL term $\mathbf{KL}(q_{\boldsymbol\omega_L}||p_{\boldsymbol\omega_L})$ and the terms
\begin{align*} 
&\sum_{i=1+H_{\mathbf{x}}-H_{\mathrm{h}}}^N \int_{\mathrm{h}_i^{(l)}}q_{h_i^{(l)}}\left(\mathrm{h}_i^{(l)}\right)\log\left(q_{h_i^{(l)}}\left(\mathrm{h}_i^{(l)}\right)\right)d \mathrm{h}_i^{(l)},\\
&\sum_{i=1+H_{\mathbf{x}}-H_{\mathrm{h}}}^{H_{\mathrm{h}}}\int_{\mathrm{h}_i^{(l)}}q_{h_i^{(l)}}\left(\mathrm{h}_i^{(l)}\right)\log\left(p_{h_i^{(l)}}\left(\mathrm{h}_i^{(l)}\right)\right)d \mathrm{h}_i^{(l)},
\end{align*}
and where
\begin{align*} 
B_l & =\frac{\hat{N}-M}{2}\log\left(\left(\upsigma_{\text{noise}}^{(l)}\right)^2\right)-\frac{\boldsymbol\upmu^T_{H_{\mathbf{x}}+1:}\boldsymbol\upmu_{H_{\mathbf{x}}+1:}}{2\left(\upsigma_{\text{noise}}^{(l)}\right)^2}\\
& + \frac{\boldsymbol\upmu_{H_{\mathbf{x}}+1:}^T\Psi_{1}^{(l)} \left(A^{(l)}\right)^{-1}(\Psi_{1}^{(l)})^T\boldsymbol\upmu_{H_{\mathbf{x}}+1:}}{2\left(\upsigma_{\text{noise}}^{(l)}\right)^2}-\frac{\log(|(A^{(l)})^{-1}|)}{2} -\frac{\hat{N}}{2}\log(2\pi),
\end{align*}
which can be separated further into $B_l= \sum_{n=1}^{\hat{N}}B_{ln}$, a sum of $\hat{N}=N-H_{\mathbf{x}}$ independent terms, ignoring $\frac{\log(|\left(A^{(l)}\right)^{-1}|)}{2}$, and adding for completeness $\frac{M}{2}\log((\upsigma_{\text{noise}}^{(l)})^2)$
\begin{align*} 
B_{ln} & =\frac{1}{2}\log\left(\frac{\left(\upsigma_{\text{noise}}^{(l)}\right)^2}{2\pi}\right)-\frac{\upmu_{n}\upmu_{n}}{2\left(\upsigma_{\text{noise}}^{(l)}\right)^2} + \frac{\upmu_{n}(\Psi_1^{(l)})_{n\LargerCdot} (A^{(l)})^{-1}(\Psi_1^{(l)})_{n\LargerCdot}^T\upmu_{n}}{2\left(\upsigma_{\text{noise}}^{(l)}\right)^2},
\end{align*}
where $(\Psi_1^{(l)})_{n\LargerCdot}$ means, taking the $n$-th column of $\Psi_1^{(l)}$ and
\begin{align*} 
A^{(l)} = \Psi_2^{(l)} + (\upsigma_{\text{noise}}^{(l)})^2 I_M = \left(\sum\limits_{n=1}^{\hat{N}} \left(\Psi_2^n\right)^{(l)}\right)+ (\upsigma_{\text{noise}}^{(l)})^2 I_M,
\end{align*}
for $l=1\dots,L+1$, $n=1\dots,\hat{N}$. \\
These terms and the sums of $\Psi_2^{(l)}$ can be computed on different nodes in parallel without communication. Only the $L+1$ inversions and determinants of $A^{(l)}$ now are responsibly for the complexity, which can also be computed on $L+1$ nodes. Summing this bound over $n$ and $l$, we receive the total complexity of $\mathcal O(M^3)$ per single iteration with $\lceil\frac{\hat{N}Q_{\text{max}}}{M}\rceil$ nodes, where the calculation of one $\left(\Psi_2^n\right)^{(l)}$ has complexity $\mathcal O(M^2Q^{(l)})$ with $Q^{(l)}\leq M$.

\section{Experiments}
\label{sec:experiments}
In this section we want to compare our methods DRGP-SS, DRGP-VSS (with optimal bound) against other well known sparse GPs and the full GP with NARX structure (see Appendix \ref{sec:NARXstructureforusewith}), the GP-SSM of~\citep{svensson2015computationally}, the DRGP-Nyström of~\citep{mattos2015recurrent} and the GP-LSTM of~\citep{al2016learning}. The full GP is named GP-full, the FITC approximation of~\citep{snelson2006sparse} is named GP-FITC, the DTC approximation of~\citep{williams2000using} is named GP-DTC, the SSGP approximation of~\citep{quia2010sparse} is named GP-SS, the VSSGP approximation of~\citep{gal2015improving} is named GP-VSS. The main setting is simulation. This means that no past measured output observations (but perhaps predicted output observations) are used to predict current or next output observations. For all experiments we will use the \textit{root mean squared error} (RMSE) for output test-data $\mathbf{y}^{\text{test}}=[\mathrm{y}_1^{\text{test}},\dots,\mathrm{y}_{N_{\text{test}}}^{\text{test}}]^T\in\mathbb{R}^{N_{\text{test}}}$
\begin{align*}
\text{RMSE} = \sqrt{\frac{1}{N_{\text{test}}}\sum_{i=1}^{N_{\text{test}}}(\mathrm{y}_i^{\text{test}}-\mathrm{y}_{i}^{\ast})^2}
\end{align*}
as measure for prediction accuracy.

\subsection{Implementation}

Our methods DRGP-(V)SS were implemented in Python, using the library Theano and Matlab R2016b. For the optimization/training we used Python, Theano. Theano allows us to take full advantage of the automatic differentiation to calculate the gradients. For simulation and visualization we used Matlab R2016b. For reproducibility of the results, we provide the code online at\\
http://github.com/RomanFoell/DRGP-VSS.\\
We further implemented in Matlab R2016b the methods DRGP-Nyström, GP-SS, GP-DTC, GP-FITC, GP-full 
and used these implementations for the experiments. For GP-VSS and GP-LSTM we used the published code\footnote{\label{GPVSS}GP-VSS code available from https://github.com/yaringal/VSSGP.}\footnote{\label{DRGPLSTM}DRGP-LSTM code available from https://github.com/alshedivat/keras-gp.}. For GP-SSM  the published code is only applicable for small range of state dimension and a small time horizon, so we just show the results from their paper.
\subsection{Data-set description}
\label{sec:Datasetdescription}

In this section we introduce the data-sets we used in our experiments. We chose a large number of data-sets in training size going from 250 until 12500 data-points in order to show the performance for a wide range. We will begin with the smallest, the \textit{Drive} data-set, which was first introduced by~\citep{wigren2010input}. It is based on a system which has two electric motors that drive a pulley using a flexible belt. The input is the sum of voltages applied to the motors and the output is the speed of the belt. The data-set \textit{Dryer}\footnote{\label{dryer}Received from http://homes.esat.kuleuven.be/~tokka/daisydata.html.} describes a system, where air is fanned through a tube and heated at an inlet. The input is the voltage over the heating device (a mesh of resistor wires). The output is the air temperature measured by a thermocouple. The third data-set \textit{Ballbeam}\footref{dryer}\footnote{\label{note2}Description can be found under http://forums.ni.com/t5/NI-myRIO/myBall-Beam-Classic-Control-Experiment/ta-p/3498079.} describes a system, where the input is the angle of a beam and the output the position of a ball. \textit{Actuator} is the name of the fourth data-set, which was described by~\citep{sjoberg1995nonlinear} and which stems from an hydraulic actuator that controls a robot arm, where the input is the size of the actuator’s valve opening and the output is its oil pressure. The \textit{Damper} data-set, introduced by~\citep{wang2009identification}, poses the problem of modeling the input–output behavior of a magneto-rheological fluid damper and is also used as a case study in the System Identification Toolbox of Mathworks Matlab. The data-set \textit{Power Load}\footnote{Originally received from Global Energy Forecasting Kaggle competitions organized in 2012.}, used in~\citep{al2016learning}, consists of data, where the power load should be predicted from the historical temperature data. This data-set was used for 1-step ahead prediction, where past measured output observations are used to predict current or next output observations, but we will use it here for free simulation. We down-sampled by starting with the first sample and choosing every 4th data-point, because the original data-set with a size of 38064 samples and a chosen time-horizon of 48 is too large for our implementation, which is not parallelized so far. The newly provided data-set \textit{Emission}\footnote{\label{note3}Available from http://github.com/RomanFoell/DRGP-VSS.} contains an emission-level of nitrogen oxide from a driving car as output and as inputs the indicated torque, boost pressure, EGR (exhaust gas recirculation) rate, injection, rail pressure and speed. The numerical characteristics of all data-sets are summarized in Table \ref{tab:datas}. The separation of the data-sets Drive, Actuator, Damper, Power Load in training- and test-data was given by the papers we use for comparison. We separated the other data-sets ourselves.
\begin{table}
\centering
\caption{Summary of data-sets for system identification tasks}
\label{tab:datas}
\begin{tabular}{r|r|r|r|p{1.6cm}|p{1.6cm}}
\hline
\textbf{parameters}&  &  &  & \multirow{3}{\linewidth}{input\\ dimension} & \multirow{3}{\linewidth}{output\\ dimension} \\
$\backslash$ & $N$ & $N_{train}$ & $N_{test}$ &  &  \\
\textbf{data-sets}&  &  &  &  &   \\
\hline
Drive & 500 & 250 & 250  & \hfill 1 & \hfill 1 \\
Dryer & 1000 & 500 & 500  & \hfill 1 & \hfill 1  \\
Ballbeam & 1000 & 500 & 500  & \hfill 1 & \hfill 1 \\
Actuator & 1024 & 512 & 512 & \hfill 2 & \hfill 1  \\
Damper & 3499 & 2000 & 1499 & \hfill 1 & \hfill 1 \\
Power Load & 9518 & 7139 & 2379 & \hfill 11 & \hfill 1 \\
Emission & 12500 & 10000 & 2500 & \hfill 6 & \hfill 1 \\
\hline
\end{tabular}
\end{table}

\subsection{Nonlinear System Identification}
\label{sec:NonlinearSystemIdentification}

\begin{figure}
\centering
	\quad\quad\quad\quad\quad\quad\quad\quad\subfloat{\includegraphics[width=0.1\textwidth]{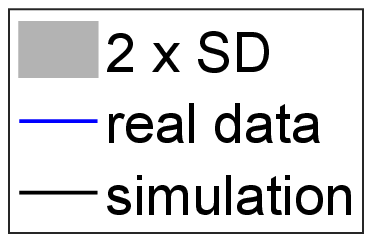}}\hfill\quad\quad\;
	\subfloat{\includegraphics[width=0.125\textwidth]{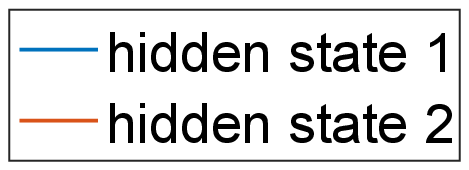}}\quad\quad\quad\quad\quad\quad\quad\quad\quad\quad\\
\centering
	\subfloat[\tiny DRGP-VSS Drive]{\includegraphics[width=0.25\textwidth]{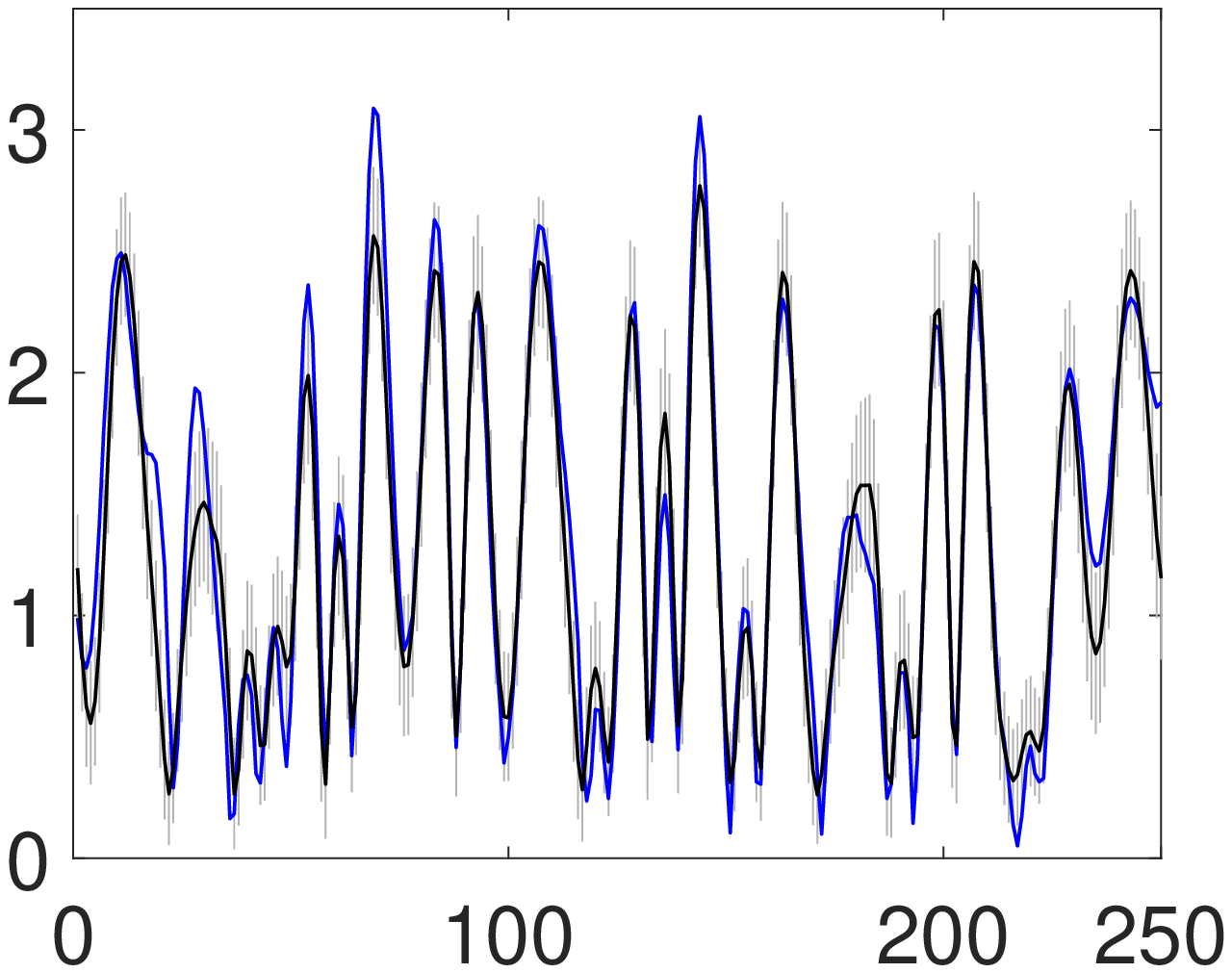}}
  \subfloat[\tiny DRGP-VSS Actuator]{\includegraphics[width=0.25\textwidth]{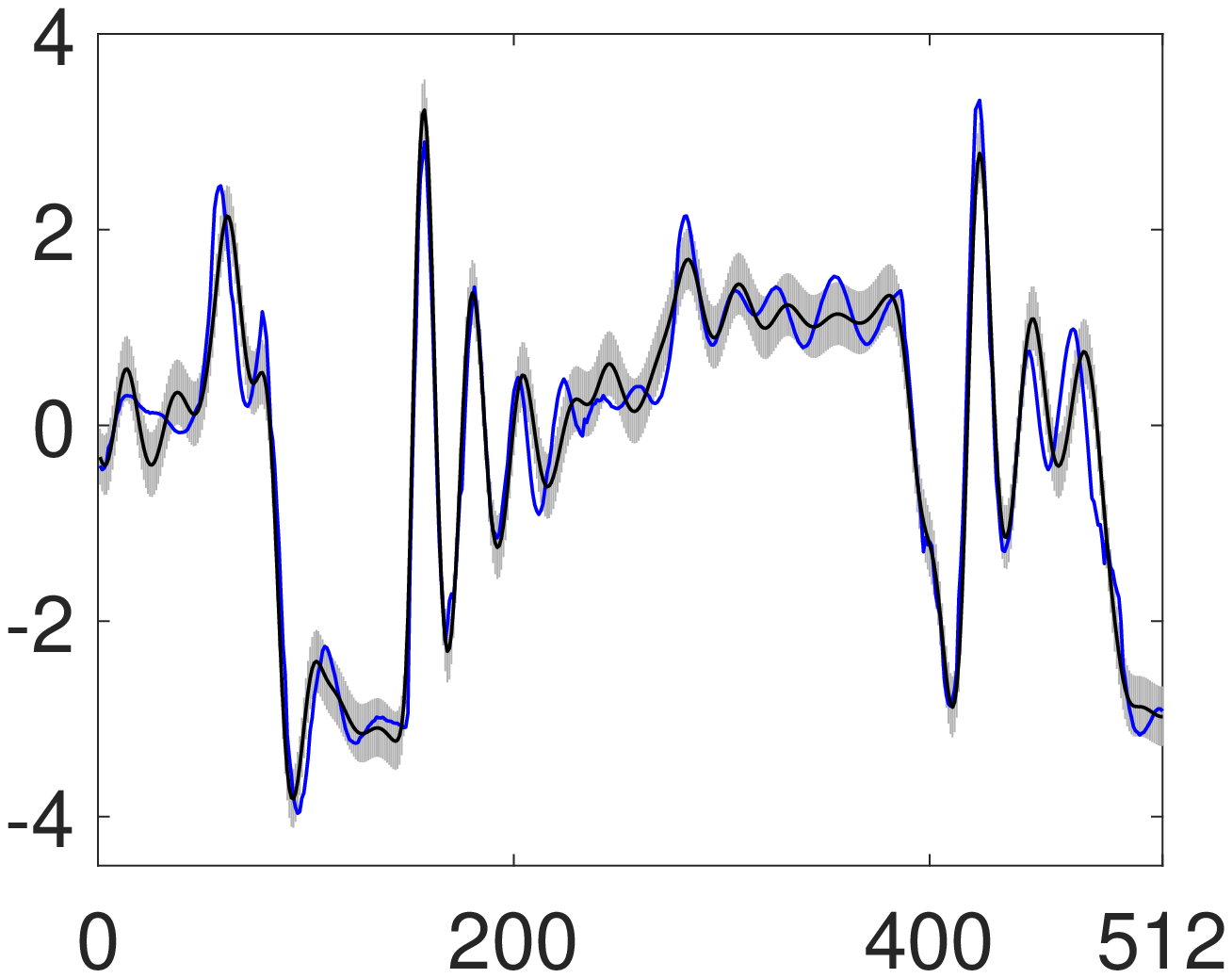}}
		\subfloat[\tiny DRGP-VSS Drive]{\includegraphics[width=0.25\textwidth]{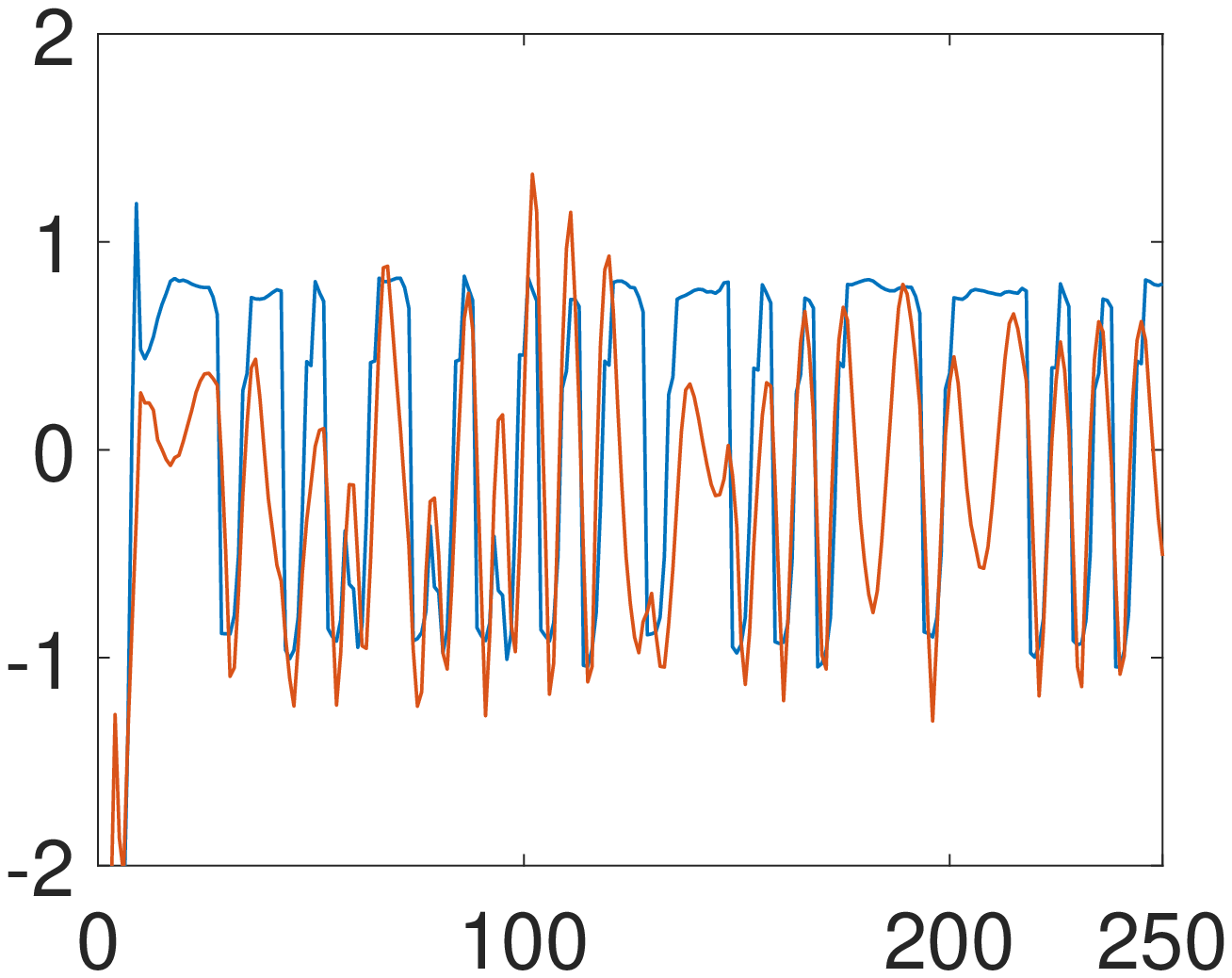}}
	\subfloat[\tiny DRGP-VSS Actuator]{\includegraphics[width=0.25\textwidth]{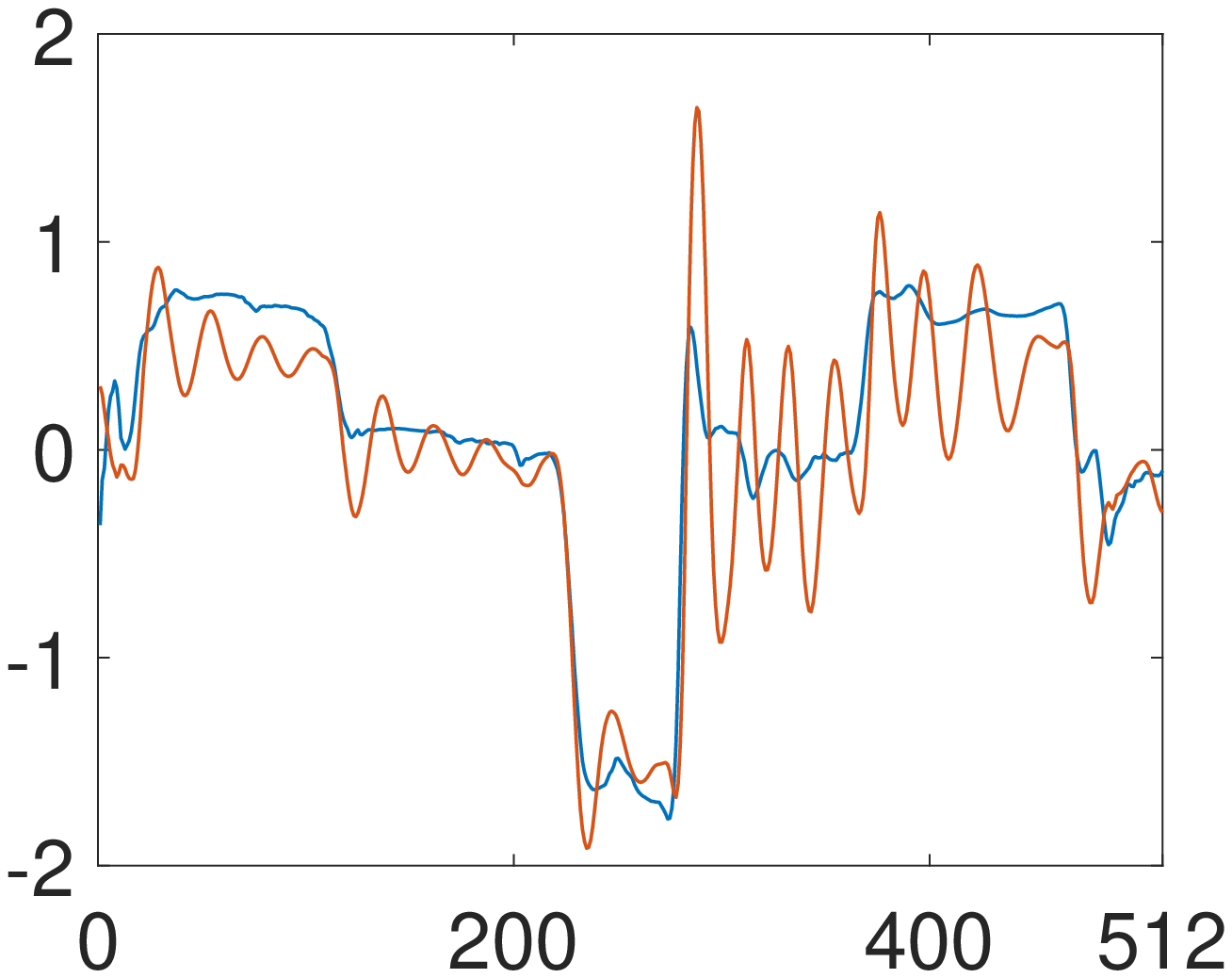}}\\
\centering
	\subfloat[\tiny DRGP-SS Drive]{\includegraphics[width=0.25\textwidth]{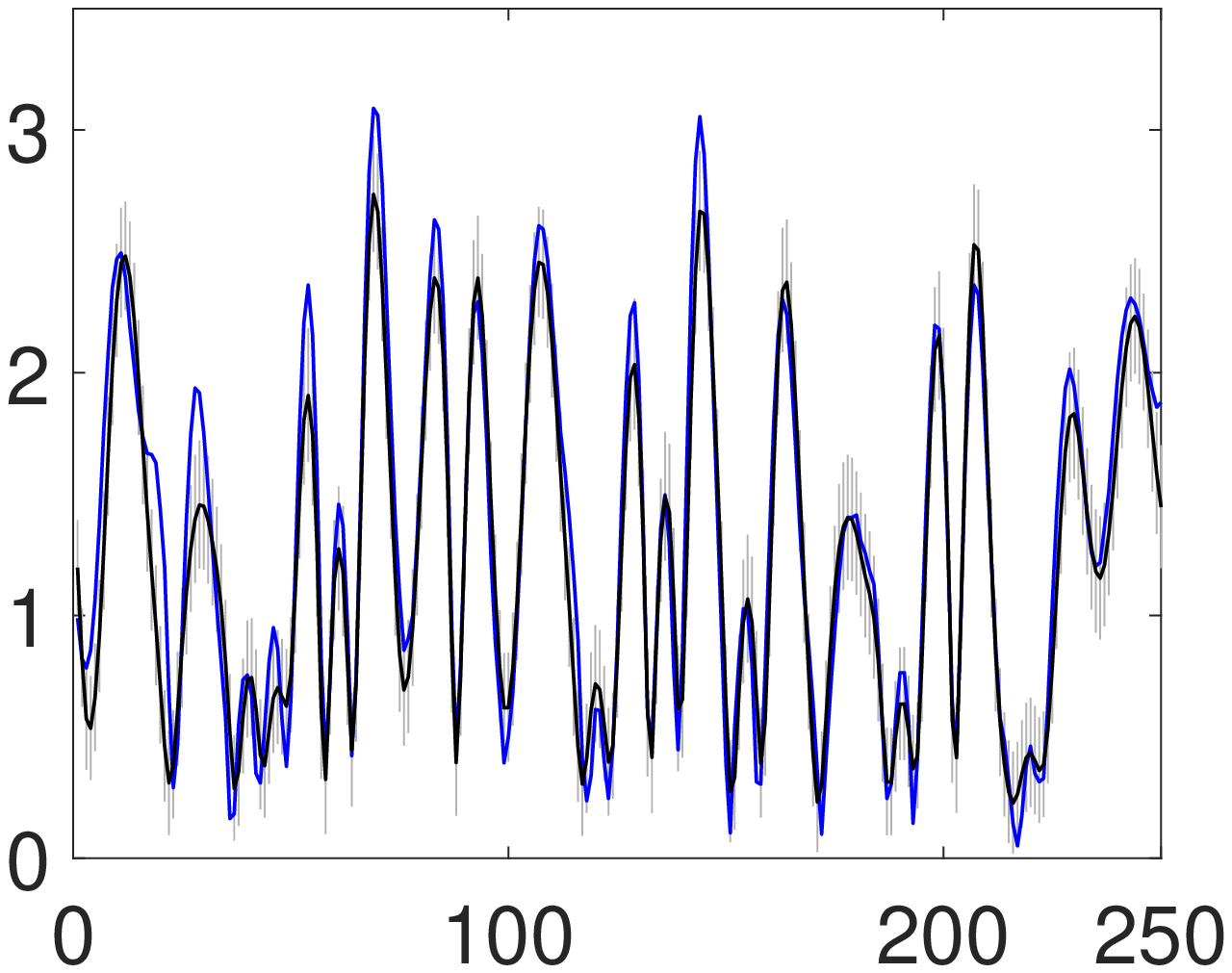}}
  \subfloat[\tiny DRGP-SS Actuator]{\includegraphics[width=0.25\textwidth]{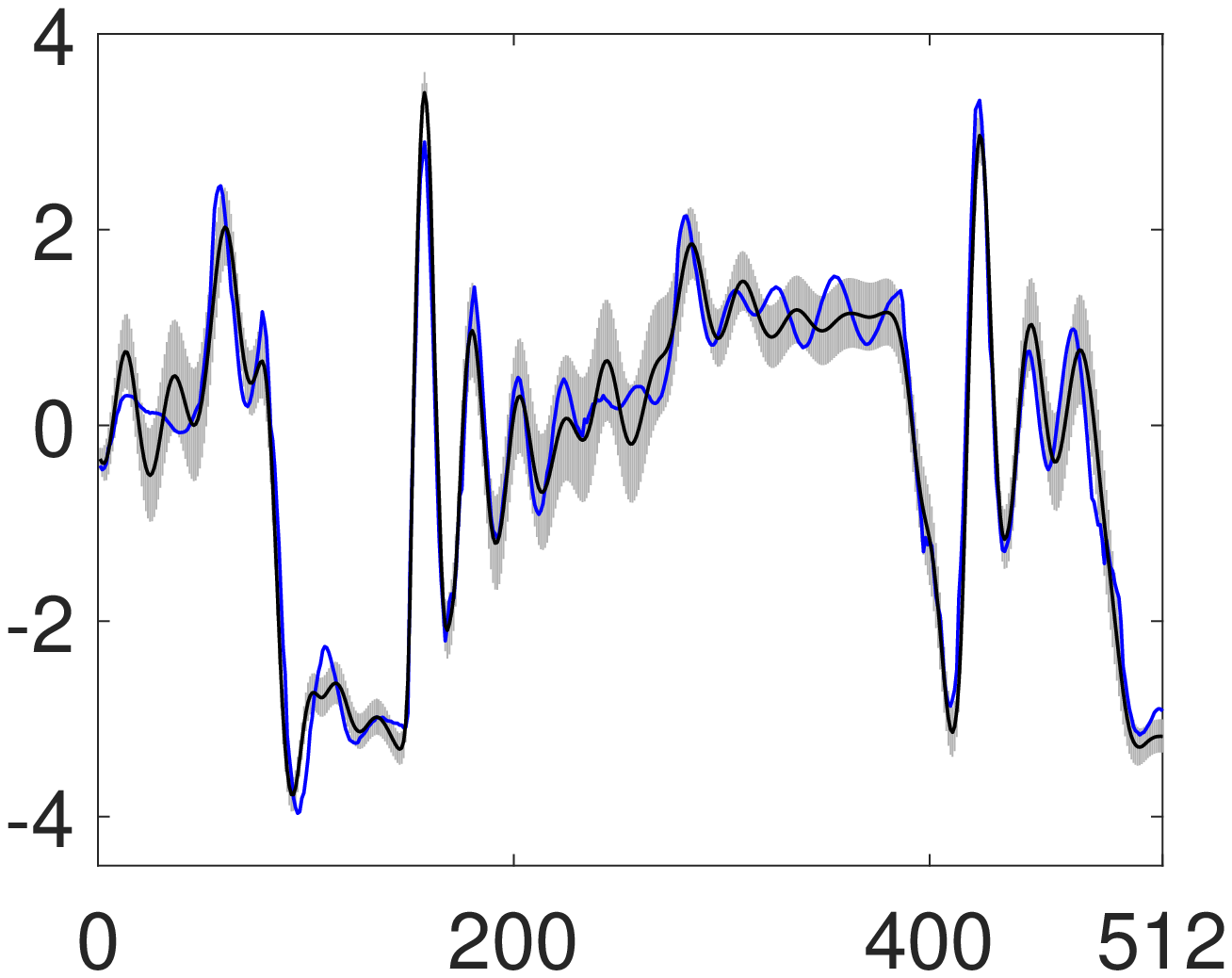}}
		\subfloat[\tiny DRGP-SS Drive]{\includegraphics[width=0.25\textwidth]{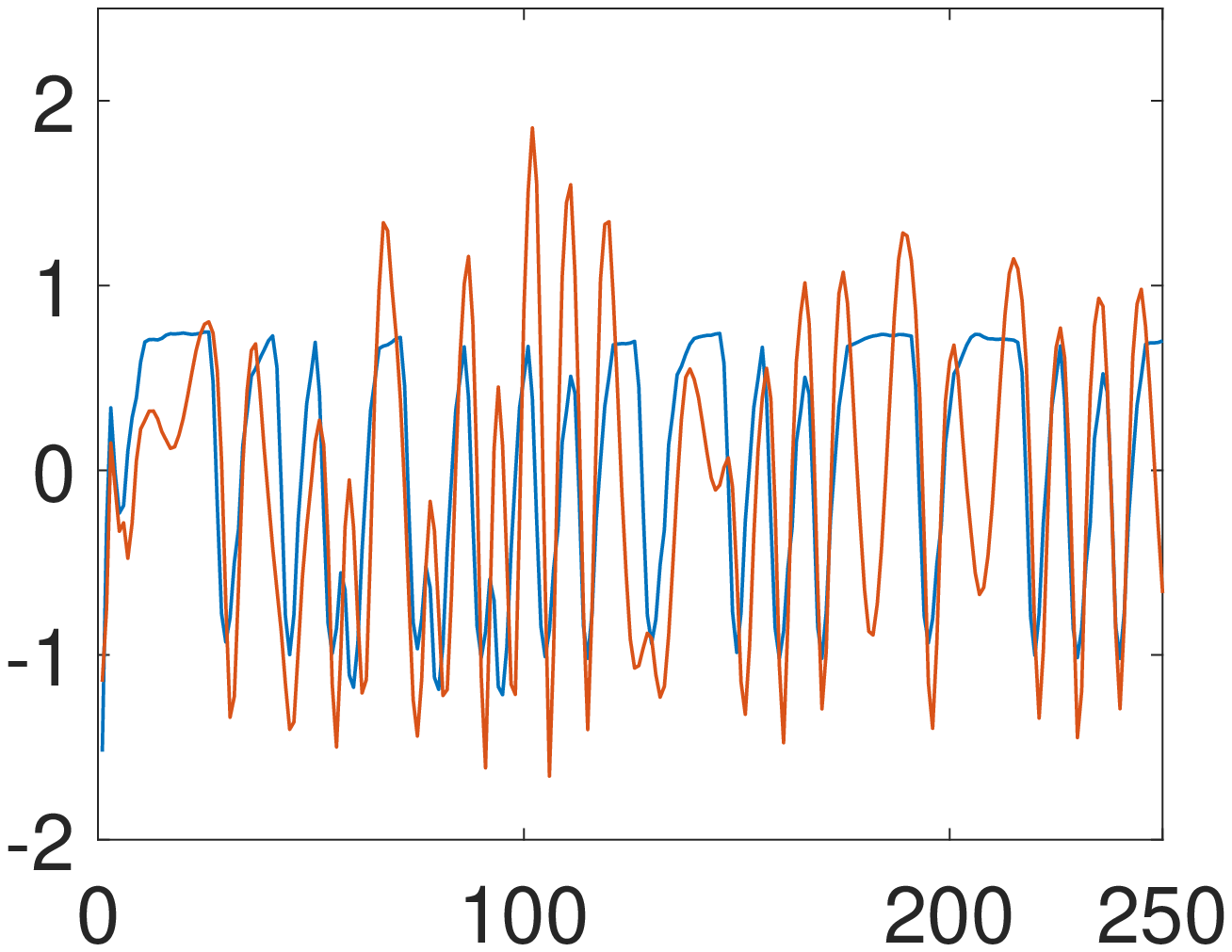}}
	\subfloat[\tiny DRGP-SS Actuator]{\includegraphics[width=0.25\textwidth]{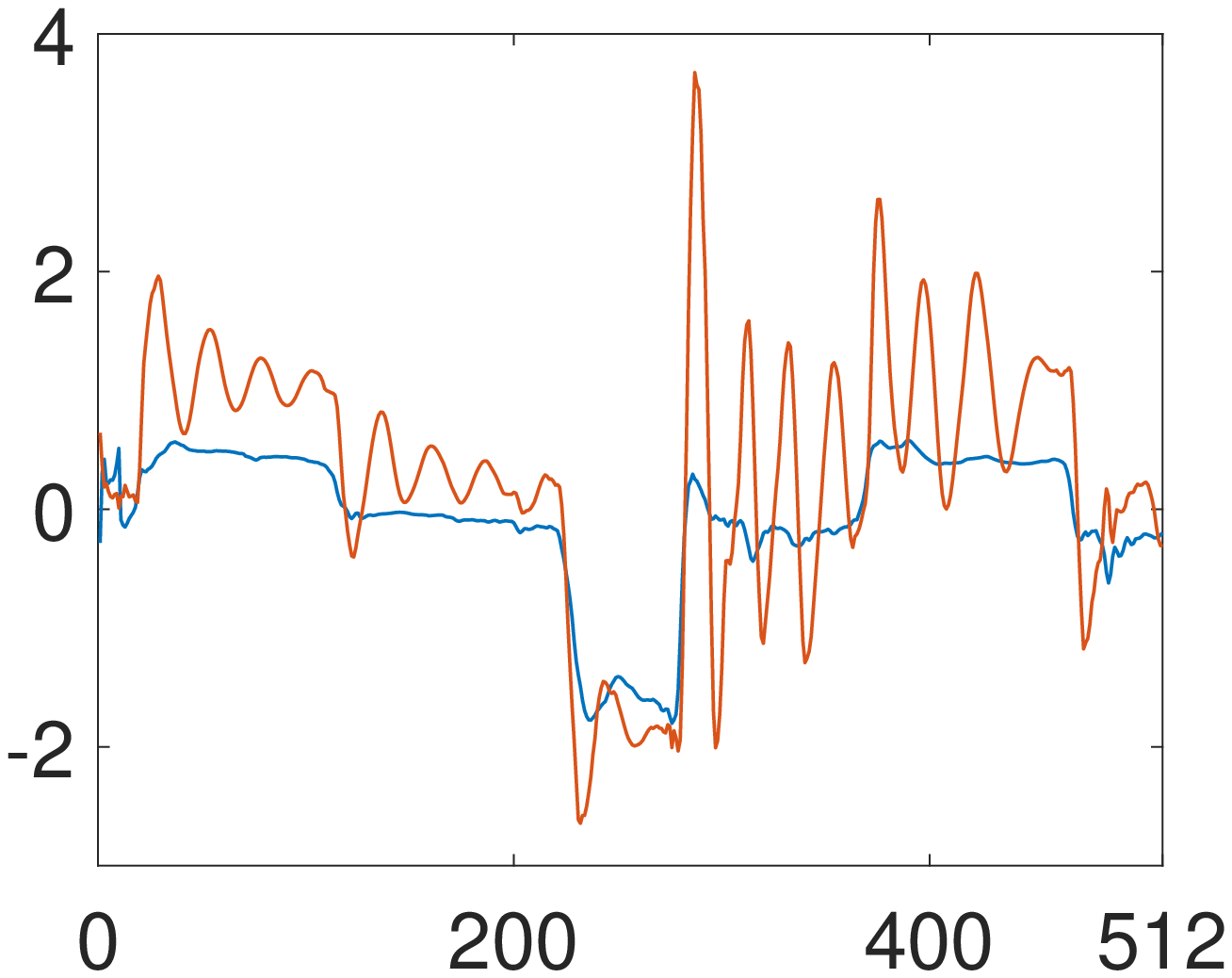}}\\
\caption{Simulation results visualized for the data-sets Drive and Actuator for the methods DRGP-VSS and DRGP-SS and their learned latent states. For simulation results on the left we have blue: real data, black: simulation, grey: $\pm 2$ times Standard Deviation (SD) and on the right the hidden states, blue: first layer, and red: second layer}
\label{fig:VIS}	
\end{figure}

For the data-sets Drive and Actuator we chose for our methods DRGP-(V)SS the setting  $L=2$ hidden layers, $M=100$ spectral-points and time-horizon $H_{\mathrm{h}} = H_{\mathbf{x}} = 10$, which was also used in~\citep{mattos2015recurrent} and~\citep{al2016learning} for free simulation (using pseudo-input points for spectral-points). For these two data-sets we filled the results from their papers into Table \ref{tab:RMSERMSE}. Further we chose for our methods DRGP-(V)SS and DRGP-Nyström on the data-sets Ballbeam and Dryer $L=1$, $M=100$ and $H_{\mathrm{h}} = H_{\mathbf{x}} = 10$. For the data-set Damper we chose $L=2$, $M=125$ and $H_{\mathrm{h}} = H_{\mathbf{x}} = 10$. For the data-set Power Load we chose $L=1$, $M=125$ and $H_{\mathrm{h}} = H_{\mathbf{x}} = 12$. For the data-set Emission we chose $L=1$, $M=125$ and $H_{\mathrm{h}} = H_{\mathbf{x}} = 10$. We show additional results for the data-sets Drive, Actuator and Emission with skipped recurrent part for the first layer in Table \ref{tab:RMSE2}. Here we want to examine the effect of noise in the inputs of the first layer on the RMSE.\\
The other sparse GPs and the full GP were trained with NARX structure $H_{\mathbf{x}} = H_{\mathrm{y}}$ with the same time horizon as for our DRGPs and with the same amount of pseudo-input points or spectral points. Again, for the data-sets Drive, Actuator and Emission we show the results with deleted auto-regressive part in Table \ref{tab:RMSE2}, so setting $H_{\mathrm{y}}=0$ and just modeling the data with exogenous output. For GP-SSM we just show the results of the data-set Damper from the paper~\citep{svensson2015computationally}.\\
For GP-LSTM we chose the same setting for the amount of hidden layers, pseudo-input points and time horizon as for our DRGPs. As activation function we chose $\tanh$ and as setting free simulation. We tested with 8, 16, 32, 64, 128 hidden units (every hidden layer of a RNN is specified with a hidden unit parameter) for all training data-sets. For the data-sets with training size smaller or equal to 2000 we chose the version GP-LSTM in~\citep{al2016learning} and for the ones larger than 2000 the scalable version MSGP-LSTM. We did not pre-train the weights.\\
All GPs which use the Nyström approximation were initialized for the pseudo-inputs points with a random subset of size $M$ from the input-data $\hat{\mathbf{H}}^{(l)}$ and trained with SE covariance function. For the ones which use the sparse spectrum approximation, which includes our methods, we trained with a spectral-point initialization sampled from $\mathcal{N}(\mathbf{0},I_Q)$, an initialization for the pseudo-input points with a random subset of size $M$ from the input-data or setting them all to zero. For our methods DRGP-(V)SS and GP-VSS we fixed the length scales $\mathrm{p}_q^{(l)} = \infty$, for all $q, l$. So all GPs with sparse spectrum approximation were also initialized as SE covariance function (see Equation \eqref{SMSS}). For all GPs we used automatic relevance determination, so each input dimension has its own length scale. For our methods DRGP-(V)SS and DRGP-Nyström, the noise parameters and the hyperparameters were initialized by $\upsigma_{\text{noise}}^{(l)}=0.1$, $\upsigma_{\text{power}}^{(l)}=1$ and the length scales by either $\mathrm{l}_q^{(l)}= \sqrt{\max(\hat{\mathbf{H}}_q^{(l)}) - \min(\hat{\mathbf{H}}_q^{(l)})}$ or $\mathrm{l}_q^{(l)}= \max(\hat{\mathbf{H}}_q^{(l)}) - \min(\hat{\mathbf{H}}_q^{(l)})$, for all $q, l$, where $\hat{\mathbf{H}}^{l}_q$ is the data-vector containing the $q$-th input-dimension values of every input-data point $\hat{\mathbf{h}}_i^{(l)}$, for all $i$. Furthermore, we initialized the latent hidden states with the output-observation data values. The other standard GPs were also initialized with $\upsigma_{\text{noise}}=0.1$, $\upsigma_{\text{power}}=1$ and the same initialization for length scales with respect to the input data with NARX structure. For the method of~\citep{al2016learning} we used the initialization for the weights provided by Keras, a Deep Learning library for Theano and TensorFlow and $\upsigma_{\text{noise}}=0.1$, $\upsigma_{\text{power}}=1$. For the length scale initialization we chose the same values for all input-dimensions. For all our implementations we used the positive transformation $\mathrm{x}'=\log(1+\exp(\mathrm{x}))^2$ for the calculation of the gradients in order for the parameters constrained to be positive.
All methods were trained on the normalized data $\frac{\dots-\mu}{\sigma^2}$, for every dimension independently, several times (same amount per data-set: the initializations are still not deterministic, e.g. for pseudo-inputs points and spectral points) with about 50 to 100 iterations with L-BFGS, either from Matlab R2016b with fmincon, or Python 2.7.12 with scipy optimize, and the best results in RMSE value on the test-data are shown. For our methods DRGP-(V)SS and DRGP-Nyström we fixed $\upsigma_{\text{noise}}^{(l)}$, $\upsigma_{\text{power}}^{(l)}$ for all $l$ (optional the spectral points/pseudo-input points) during the first iterations to independently train the latent states. For all other GPs we also tested with fixed and not fixed $\upsigma_{\text{noise}}=0.1$, except GP-LSTM. For DRGP-VSS we also fixed $\boldsymbol\upbeta_m^{(l)}$ for all $m$, $l$ to small value around $0.001$, as well as $\mathrm{b}_m$ for all $m$, $l$ sampling from $\text{Unif}\left[0,2\pi\right]$ (this seems to work better in practice). The limitations for $\boldsymbol\upbeta_m^{(l)}$ also holds for VSS as well. We want to signify at this point that setting $\mathbf{u}_m=\mathbf{0}$ for all $m=1,\dots,M$ worked sometimes better than choosing a subset from the input-data. This seems to be different to~\citep{gal2015improving}, who pointed out: ’These are necessary to the approximation. Without these points (or equivalently, setting these to $\mathbf{0}$), the features would decay quickly for data points far from the origin (the fixed point $\mathbf{0}$).’ 
The results are shown in the Tables \ref{tab:RMSERMSE} and \ref{tab:RMSE2}. In Figure \ref{fig:VIS} we visualized the simulation results for the data-sets Drive and Actuator for our methods DRGP-(V)SS.
\begin{table}
\centering
\caption{Summary of RMSE values for the free simulation results on system identification test data. Best values per data-set are bold. All values are calculated on the original data, unless the data-set Power Load, where the RMSE is shown for the normalized data. Here we have full recurrence for our methods DRGP-SS, DRGP-VSS and DRGP-Nyström and with auto-regressive part for standard sparse GPs and full GP}
\label{tab:RMSERMSE}
\scalebox{0.85}{
\begin{tabular}{r|r|r|r|r|r|r|r}
\hline
\textbf{data-sets}&  &  &  &  &  &  &  \\
$\backslash$ & Emission & Power Load & Damper & Actuator & Ballbeam & Dryer &  Drive \\
\textbf{methods} &  &  &  &  &  &  & \\
\hline
DRGP-VSS & \textit{0.104} & \textbf{\textit{0.457}} & \textit{5.825} & \textit{0.357} & \textit{0.084} &  \textit{0.109} & \textit{0.229}\\
DRGP-SS & \textit{0.108} & \textit{0.497} & $\textbf{\textit{5.277}}$ & $\textbf{\textit{0.329}}$ & \textit{ 0.081} & \textit{0.108} & $\textbf{\textit{0.226}}$\\
DRGP-Nyström & 0.109 & 0.493 & 6.344  & 0.368 & 0.082 & 0.109 & 0.249\\
GP-LSTM &  \textbf{0.096} & 0.529 & 9.083 & 0.430 & \textbf{0.062} & 0.108 & 0.320\\
GP-SSM & N/A & N/A & 8.170 & N/A & N/A & N/A &  N/A\\
GP-VSS & 0.130 & 0.514 & 6.554 & 0.449 & 0.120 & 0.112 & 0.401\\
GP-SS & 0.128 & 0.539 & 6.730 & 0.439 & 0.077 &  0.106 & 0.358\\
GP-DTC & 0.137 & 0.566 & 7.474 & 0.458 & 0.122 & \textbf{0.105} & 0.408\\
GP-FITC & 0.126 & 0.536 & 6.754 & 0.433 & 0.084 & 0.108 & 0.403\\
GP-full & 0.122 & 0.696 & 9.890 & 0.449 & 0.128 & 0.106 & 0.444\\
\hline
\end{tabular}}
\end{table}
\begin{table}
\centering
\caption{Summary of RMSE values for the free simulation results on system identification test data. Best values are bold. Here we have missing recurrence in the first layer for our methods DRGP-SS, DRGP-VSS and DRGP-Nyström and missing auto-regressive part for standard sparse GPs and full GP. GP-LSTM and GP-SSM not listed (we used their code for calculation, which is not adaptable for this recurrence setting or overall not possible to calculate)}
\label{tab:RMSE2}
\scalebox{0.85}{
\begin{tabular}{r|r|r|r}
\hline
\textbf{data-sets}&  &  &   \\
$\backslash$ & Emission & Actuator  & Drive \\
\textbf{methods} &  &  &  \\
\hline
DRGP-VSS & \textit{0.062} & \textbf{\textit{0.388}} & \textit{0.268} \\
DRGP-SS & \textit{0.062} & \textit{0.563} &  \textbf{\textit{0.253}} \\
DRGP-Nyström & 0.059 & 0.415 & 0.289 \\
GP-VSS & 0.058 & 0.767 & 0.549\\
GP-SS & 0.060 & 0.777 & 0.556\\
GP-DTC & 0.061 & 0.864 & 0.540\\
GP-FITC & \textbf{0.057}& 0.860 & 0.539\\
GP-full & 0.066 &  1.037 &  0.542\\
\hline
\end{tabular}}
\end{table}

\subsubsection{Discussion of the  Nonlinear System Identification}

We tested our methods DRGP-(V)SS on a large number of data-sets and compared it with some state of the art methods DRGP-Nyström and GP-LSTM, the full GP and some well known sparse GPs. In Table \ref{tab:RMSERMSE} (full recurrence) we can see on most data-sets improved results for DRGP-SS and DRGP-VSS over the other methods. Both our methods perform sometimes better than the other, where DRGP-SS slightly dominates over DRGP-VSS in most cases. This might be caused by the additional regularization properties for DRGP-VSS in terms of the extra variational spectral point variance and the KL-term, which forces DRGP-VSS to not diverge far away from the SE kernel. Compared to DRGP-Nyström, our methods produced better results on nearly all data-sets. For GP-LSTM we have to mention, that free simulation in their paper was defined differently to ours. GP-LSTM still has an auto-regressive part, where its own past output predictions are reused. Our setting does recurrence just for the latent hidden states and no past predictions are reused. This might be a reason for their better results achieved on some data-sets. For the data-set Dryer we got competitive RMSE values for all methods. This might be a hint that this data-set with the chosen setting is quite easy to model. On the data-set Ballbeam GP-LSTM got the best result, followed by GP-SS, GP-FITC, our methods and DRGP-Nyström. Comparing our results of free simulation of the data-set Power Load with the results of this data-set in~\citep{al2016learning} for 1-step ahead prediction (of course this is no fair comparison), we are clearly not that good (received value of 0.158). Perhaps with pre-trained initializations for the weights for GP-LSTM, which was the case in~\citep{al2016learning} for 1-step ahead prediction, we could get better results for free simulation for GP-LSTM. In Appendix \ref{sec:AdditionalResults} Figure \ref{fig:Power} we show the result for our method DRGP-SS on this data-set. For the results on the Emission data-set in Table \ref{tab:RMSERMSE}, we see that the DRGPs and GP-LSTM outperform the standard GPs headed by GP-LSTM. Though it seems that DRGP-Nyström, DRGP-(V)SS and GP-LSTM still have problems finding similar fittings as with skipped recurrent part in the first layer in Table \ref{tab:RMSE2}. But they are able to achieve a much better fitting compared to the others. This is explained most likely through the capability of optimizing the recurrent latent state in the first layer. For the data-set Emission, past emission values correlate with the current emission value for the data-sets frequency only a little. So recurrence of this variable should actually be not important. We also tried some other initializations for the latent hidden states on this data-set (0.1 times output, PCA of the input) for our methods DRGP-(V)SS and DRGP-Nyström, but without any improvements. We think that more research on good initializations of the hidden states is necessary.\\
For the Emission data-set and the results in Table \ref{tab:RMSE2} we observe that all methods show similar results. This might be also an indication that this data-set with this specific setting of recurrence is overall easy to model. For the data-sets Actuator in Table \ref{tab:RMSE2} we see that the results for our methods DRGP-(V)SS and DRGP-Nyström are quite good compared to Table \ref{tab:RMSERMSE}. Looking for the results of this data-set comparing Table \ref{tab:RMSE2} with Table \ref{tab:RMSERMSE}, all methods suffer from the missing recurrence. The Actuator data-set's past outputs seem important, which might be effected by inertia of the whole system. For the Actuator data-set, we have to say that this data-set is not easy to model with GPs in general. DRGP-Nyström is better in finding a good latent state representation than our methods which might come from the additional regularization terms in the REVARB bound, which lead to more stabilization in the overall optimization process than for our methods. The method of~\citep{al2016learning} got for the most similar setting to ours in Table \ref{tab:RMSE2}, which they call regression (still there exist a recurrent part in the first layer), good results, but with a much larger required time lag $H_{\mathrm{h}} = H_{\mathbf{x}} =32$. For the data-set Drive in Table \ref{tab:RMSE2} we see overall worse results compared to Table \ref{tab:RMSERMSE} with DRGP-SS performing best.

\subsection{Optimization and parameter study of DRGP-(V)SS}
\begin{figure}
{\includegraphics[width=0.1\textwidth]{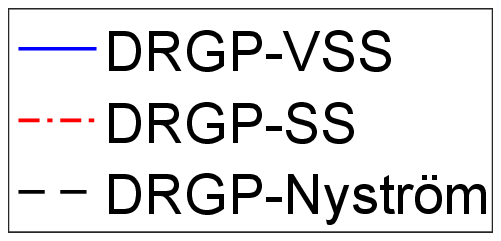}}\hfill
\centering
	\subfloat[\tiny Test RMSE vs\endgraf\hspace{0.5cm} iterations]{\includegraphics[width=0.25\textwidth]{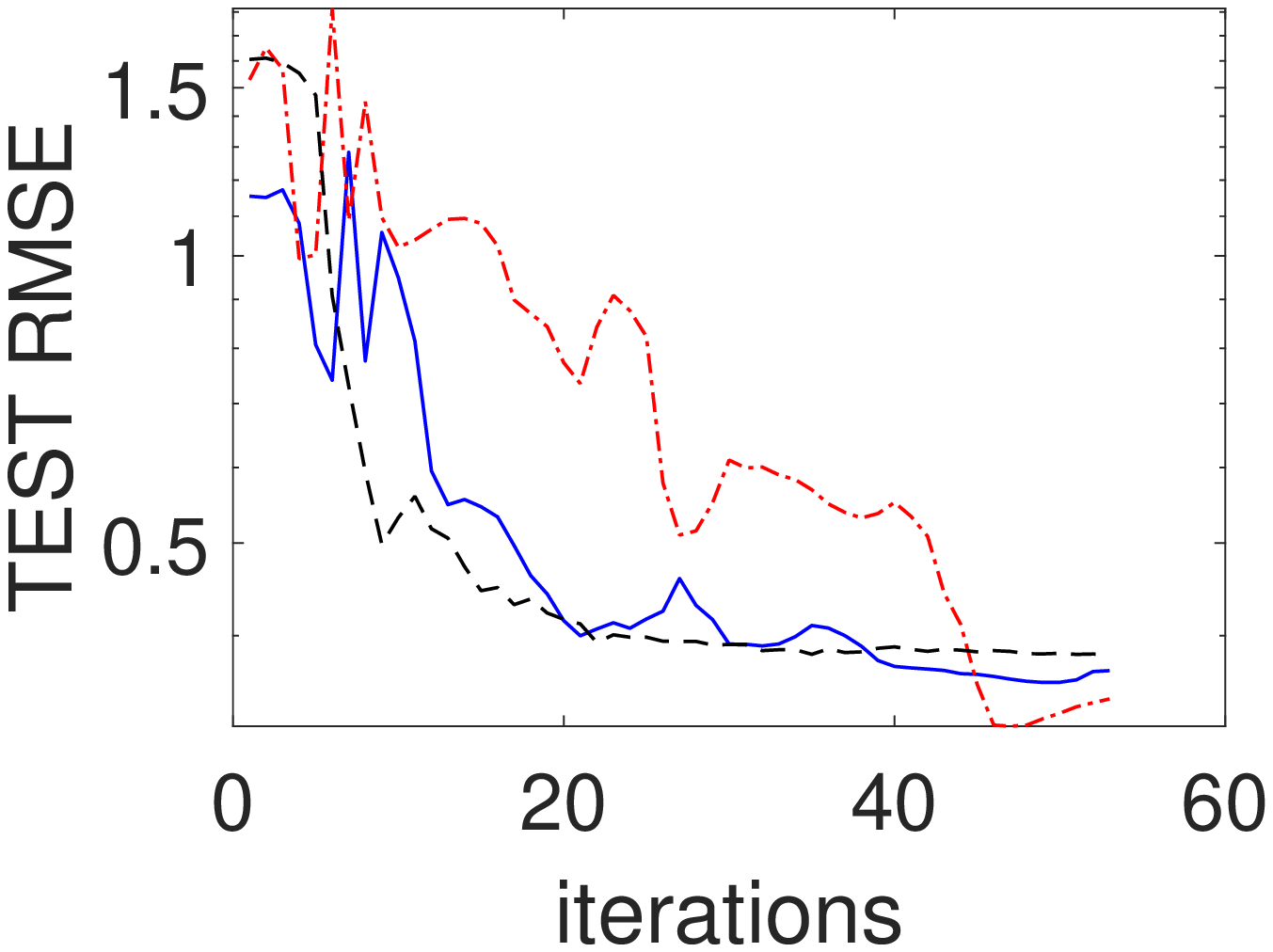}}
  \subfloat[\tiny NLBV vs\endgraf\hspace{0.5cm} iterations]{\includegraphics[width=0.25\textwidth]{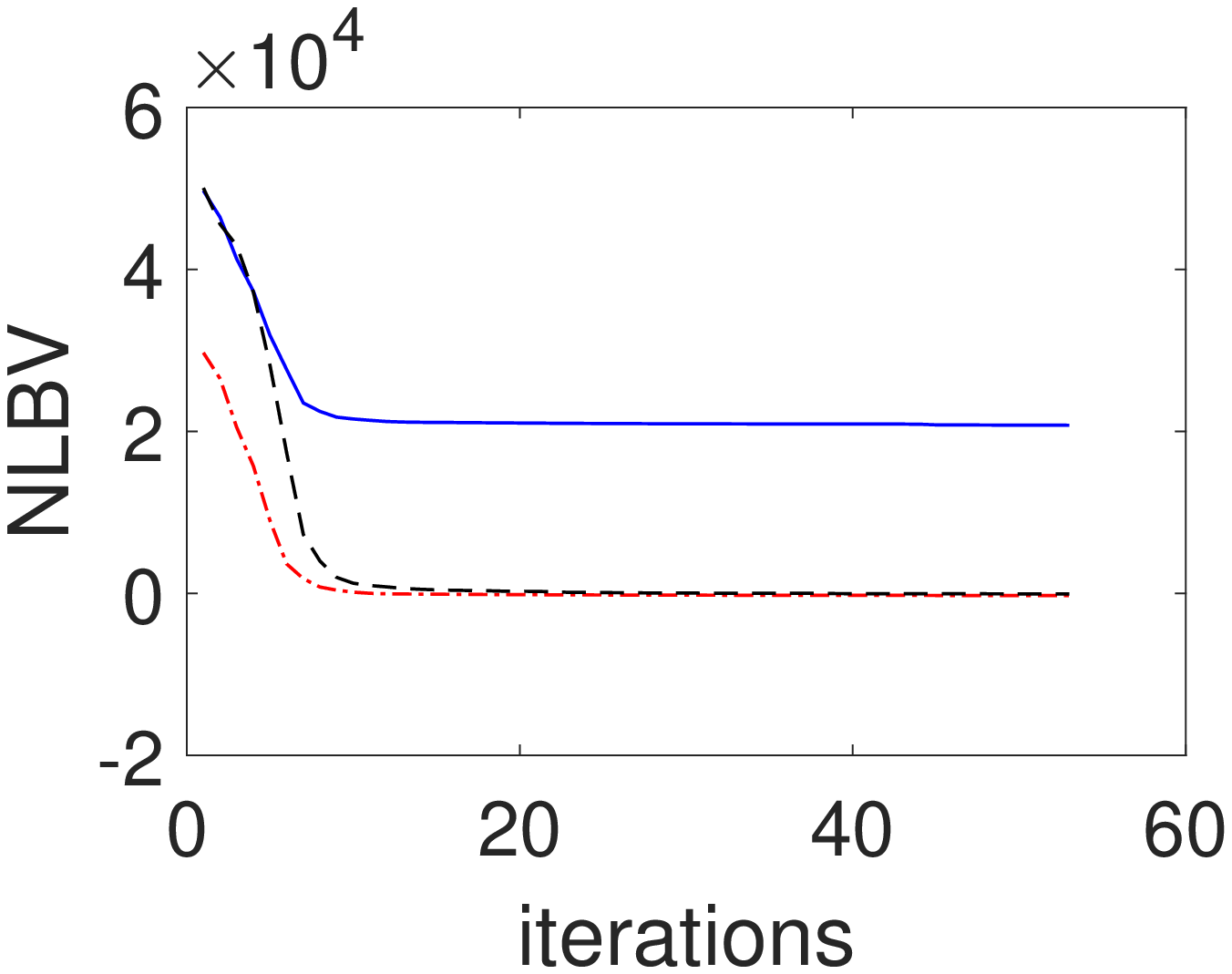}}
	\subfloat[\tiny Test RMSE vs\endgraf\hspace{0.5cm} time horizon]{\includegraphics[width=0.25\textwidth]{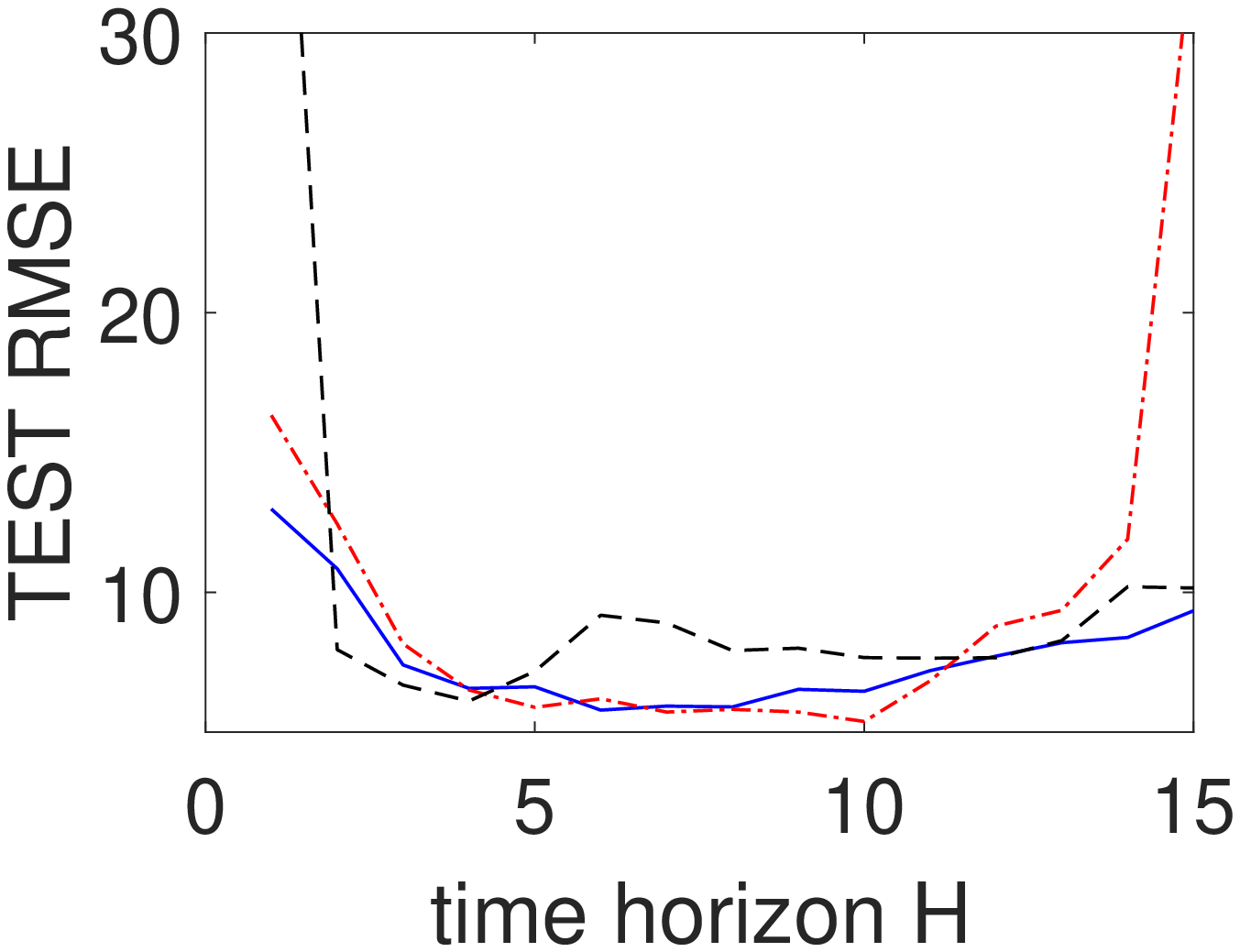}}
	\subfloat[\tiny Test RMSE vs\endgraf\hspace{0.5cm} layers L]{\includegraphics[width=0.25\textwidth]{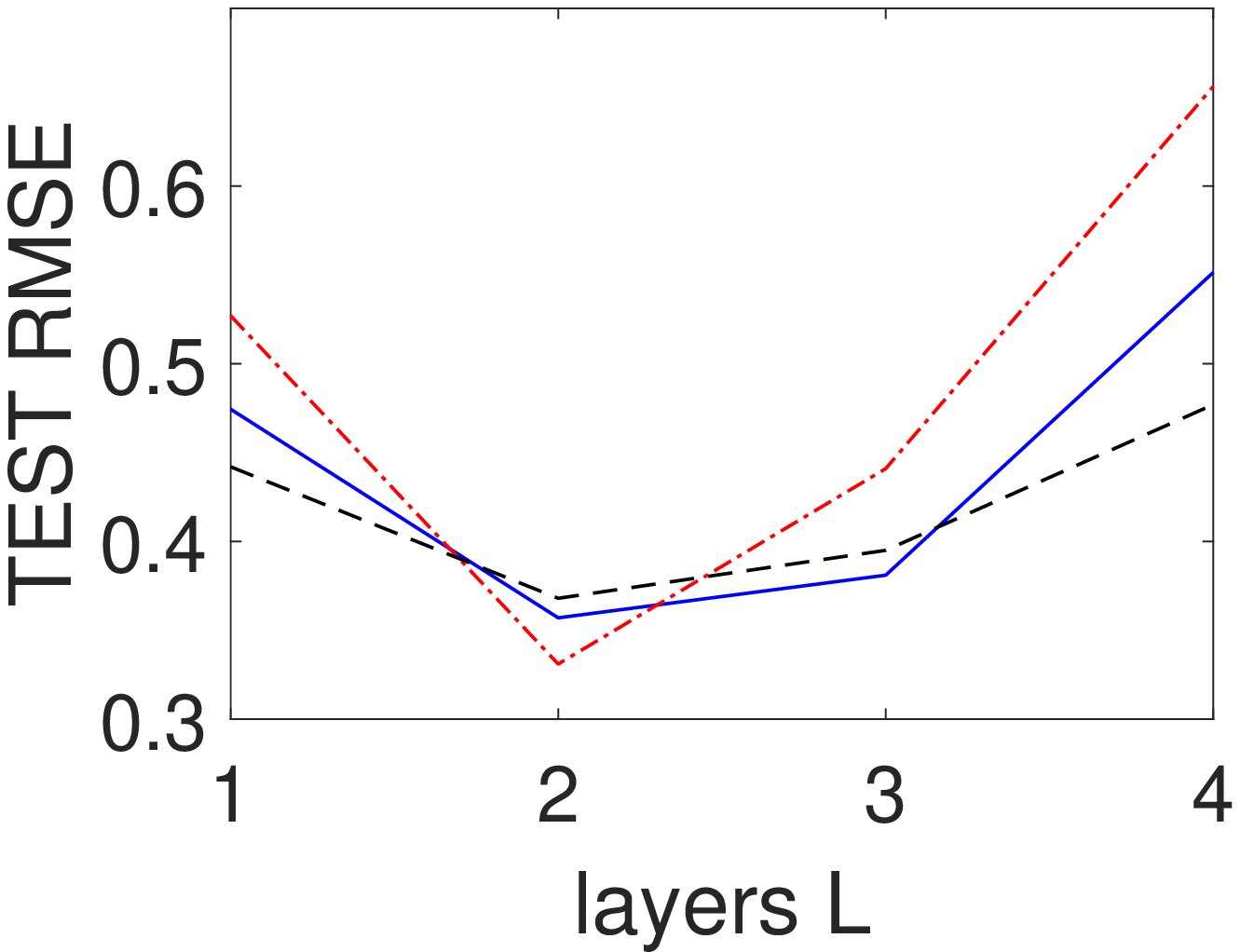}}\\
\caption{Results on the data-sets Actuator and Damper. Figure (a) shows the RMSE on test data for every iteration from 1 to 70 for $L = 2$ and $H = H_{\mathbf{x}} = H_{\mathrm{h}} = 10$ on the data-set Actuator. Figure (b) show the NLBV for every iteration from 1 to 70 for $L = 2$ on the data-set Actuator. Figure (c) shows the Test RMSE versus the time horizon $H = H_{\mathbf{x}} = H_{\mathrm{h}}$ on the Damper data-set with $L = 2$. Figure (d) shows the Test RMSE versus the layers L on the Actuator data-set for $H = H_{\mathbf{x}} = H_{\mathrm{h}} = 10$}
\label{fig:VIS2}	
\end{figure}

In Figure \ref{fig:VIS2} (a), (b) we can see the evolution of the optimization process in terms of Test RMSE and \textit{negative lower bound value} (NLBV) for every iteration on the data-set Actuator. In Figure \ref{fig:VIS2} (c) we can see the Test RMSE values for different settings of the time horizon $H = H_{\mathbf{x}} = H_{\mathrm{h}}$ on the data-set Damper. Figure \ref{fig:VIS2} (d) shows results for different values of layers $L$ on the Actuator data-set. For Figure \ref{fig:VIS2} (c) we trained 3 times for every value of the time horizon and took the average. For Figure \ref{fig:VIS2} (d) we trained several times for every layer (same amount) and took the best value. For all experiments we used the same setting as described in Section \ref{sec:NonlinearSystemIdentification}.

\subsubsection{Discussion Optimization and parameter study}

From Figure \ref{fig:VIS2} (a), where we see the Test RMSE evolution over the iterations of the data-set Actuator, one can deduce that all methods arrive at their best results after 50-60 iterations with some fluctuations over the whole optimization procedure especially for DRGP-SS, but also for DRGP-VSS. DRGP-Nyström seems to be more stable in this sense then our methods. 
A reason might be the effect of the extra regularization terms in the REVARB bound. In~\citep{bui2016unifying} investigations show similar properties. A second reason might be that the sparse spectral representation of the covariance function has more oscillations on the whole axes (although it is decaying to zero) than the sparse Nyström approximation, and this nature causes more difficulties in the optimization procedure. The evolution of the NLBV in \ref{fig:VIS2} (b) seems reasonable. DRGP-Nyström and DRGP-SS show a similar behavior. DRGP-VSS does not come down as far as the other two, which stems from the extra KL-term. In Figure \ref{fig:VIS2} (c) we can see the evolution of the Test RMSE over different time horizons on the data-set Damper. DRGP-SS is a little bit more robust for the chosen time horizon compared to DRGP-VSS for wider horizons, but then starts to get worse even faster. This is in conformity with the robustness test in Figure \ref{fig:ROBUST1}. DRGP-Nyström is even more sensitive to wider horizons. In Figure \ref{fig:VIS2} (d) we compare the DRGPs for different hidden layers on the Actuator data-set. Here we see that our methods seem more sensitive for the right choice of hidden layers on this data-set especially for DRGP-SS. But we can see that they all share the same minimum, which is reached for $L=2$ hidden layers.

\subsection{Robustness of DRGP-(V)SS}
\begin{figure}
\scalebox{0.81}{\begin{minipage}{0.65\textwidth}
\captionof{table}{Data-set Drive, descriptive values Mean, Standard Deviation (SD), Best, Worst for our methods DRGP-SS, DRGP-VSS for different settings of time horizons $H = H_{\mathbf{x}} = H_{\mathrm{h}} = 6, 10, 12$}
\label{tab:ROBUST2}
\begin{tabular}{r|r|r|r|r}
\hline
\textbf{measure}&  &  &  & \\
$\backslash$ & Mean & SD & Best & Worst \\
\textbf{methods} &  &  &  & \\
\hline
DRGP-Nyström, H=6 &  0.408 & 0.041 & 0.373 & 0.498 \\
DRGP-VSS, H=6 & 0.390 &  0.045 &  0.295 & 0.455 \\
DRGP-SS, H=6 & 0.383 & 0.039 & 0.303 & 0.430 \\
DRGP-Nyström, H=10 & 0.365 & 0.031 &  0.311 & 0.404 \\
DRGP-VSS, H=10 & 0.268 & 0.024 & 0.229 & 0.308 \\
DRGP-SS, H=10 &  0.309 &  0.051 & 0.226 & 0.368 \\
DRGP-Nyström, H=12 &  0.361 & 0.011 & 0.349 & 0.377 \\
DRGP-VSS, H=12 & 0.248 &  0.023 & 0.219 & 0.301 \\
DRGP-SS, H=12 & 0.306 & 0.056 &  0.225 & 0.424 \\
\hline
\end{tabular}
\end{minipage}}
\hspace{0.75cm}
\begin{minipage}{0.41\textwidth}
\includegraphics[width=1\textwidth]{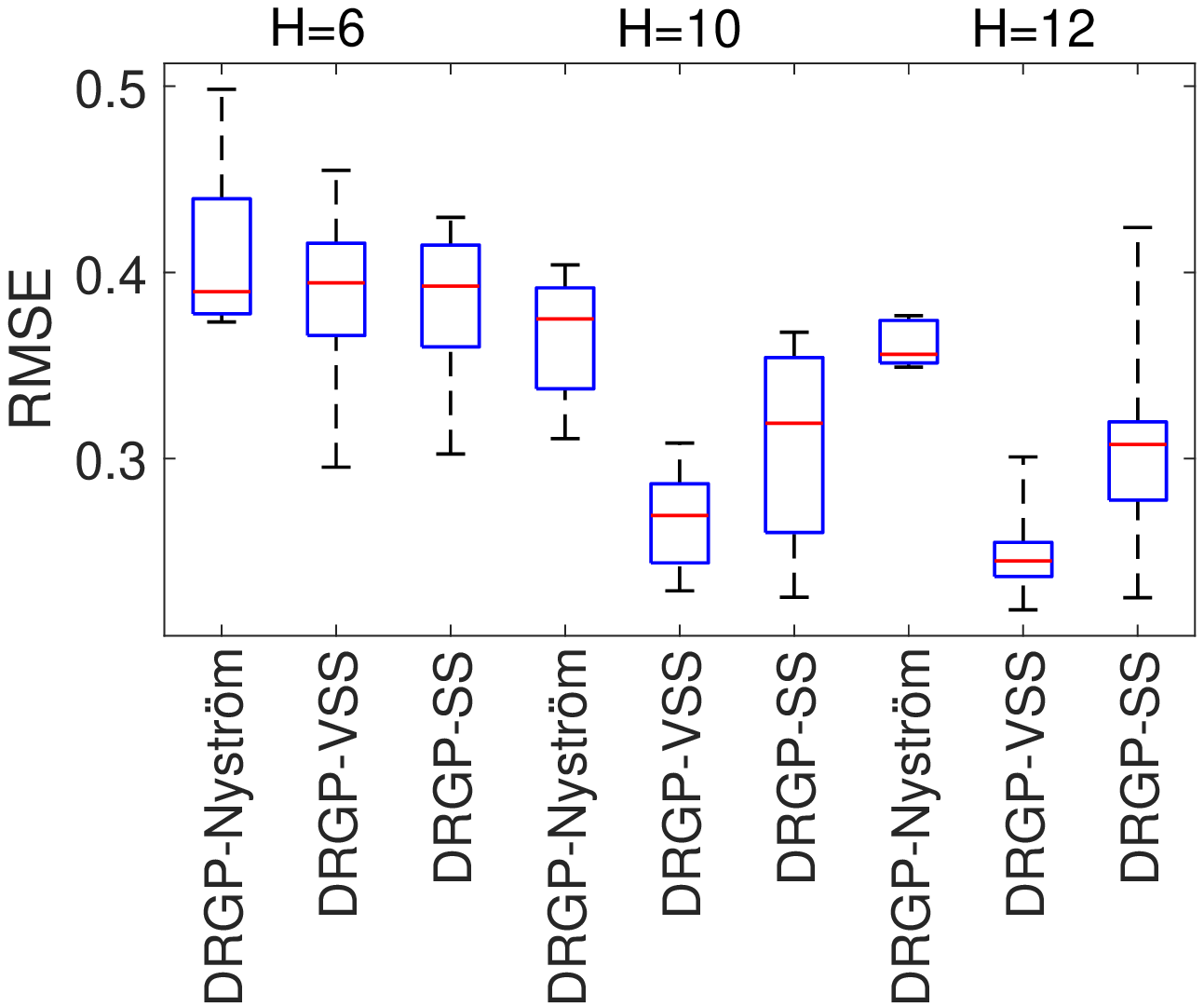}
\captionof{figure}{Data-set Drive boxplot with whiskers from minimum to maximum}
\label{fig:ROBUST2}
\end{minipage}
\end{figure}

\begin{figure}
\scalebox{0.81}{\begin{minipage}{0.65\textwidth}
\captionof{table}{Data-set Damper descriptive values Mean, Standard Deviation (SD), Best, Worst for our methods DRGP-SS, DRGP-VSS for different settings of time horizons $H = H_{\mathbf{x}} = H_{\mathrm{h}} = 2, 6, 10$}
\label{tab:ROBUST1}
\begin{tabular}{r|r|r|r|r}
\hline
\textbf{measure}&  &  &  & \\
$\backslash$ & Mean & SD & Best & Worst \\
\textbf{methods} &  &  &  & \\
\hline
DRGP-Nyström, H=2 & 7.566 & 0.506 & 6.649 & 8.193 \\
DRGP-VSS, H=2 & 10.146 & 1.546 & 8.152 & 12.807 \\
DRGP-SS, H=2 & 12.761 & 0.989 & 11.338 & 14.710 \\
DRGP-Nyström, H=6 & 8.666 & 0.642 & 7.893 & 9.634 \\
DRGP-VSS, H=6 & 5.913 & 0.280 & 5.263 & 6.174 \\
DRGP-SS, H=6 & 5.820 & 0.338 & 5.141 & 6.271 \\
DRGP-Nyström, H=10 & 7.893 & 0.760 & 7.123 & 9.205 \\
DRGP-VSS, H=10 & 6.962 & 0.762 & 5.825 & 8.179 \\
DRGP-SS, H=10 & 5.814 & 0.478 & 5.277 & 6.863 \\
\hline
\end{tabular}
\end{minipage}}
\hspace{0.75cm}
\begin{minipage}{0.41\textwidth}
\includegraphics[width=1\textwidth]{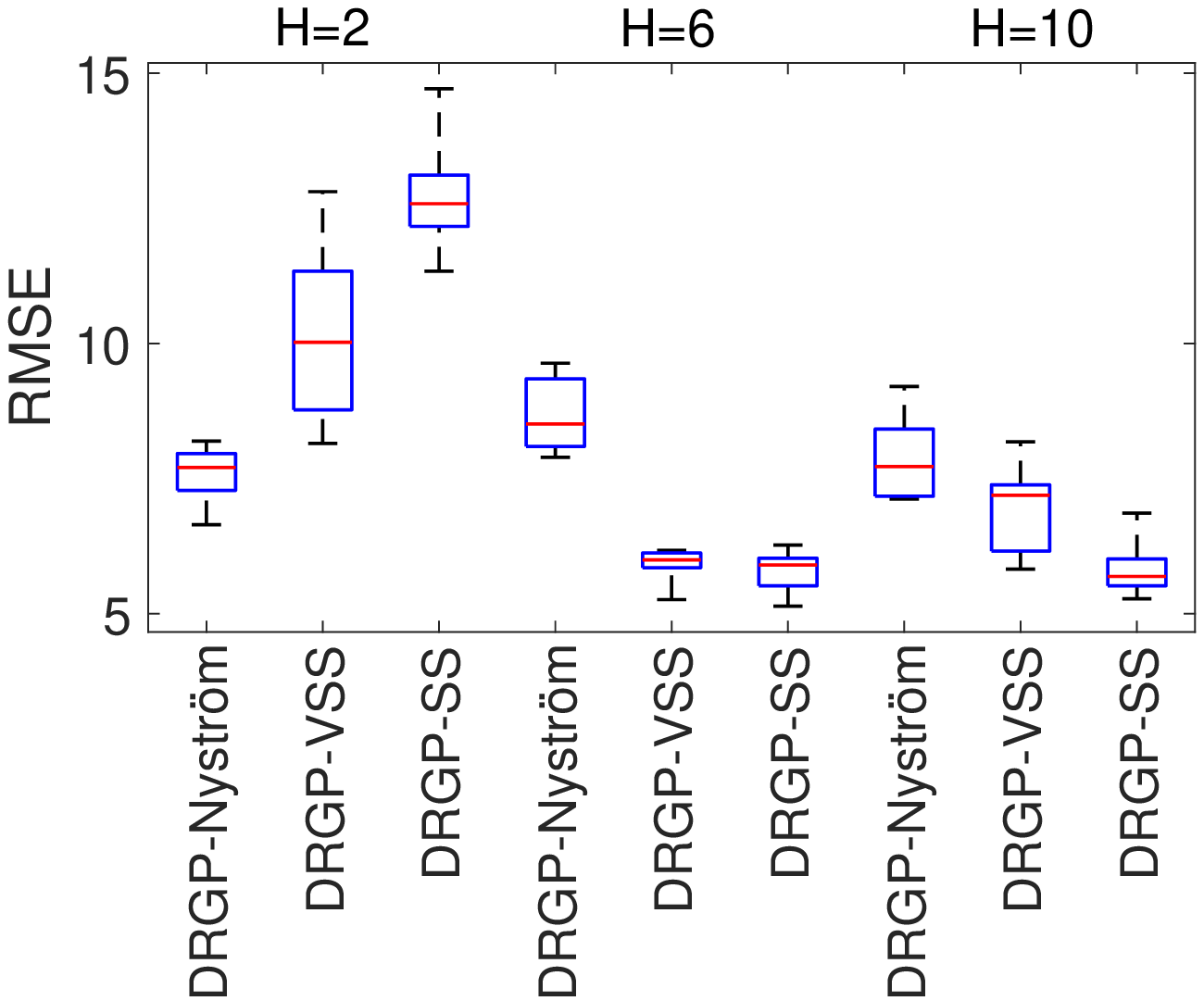}
\captionof{figure}{Data-set Damper boxplot with whiskers from minimum to maximum}
\label{fig:ROBUST1}
\end{minipage}
\end{figure}

In order to evaluate the reproducing quality of our results on some data-sets, we provide a robustness test on the data-sets Drive and Damper with 10 runs for every method with different time horizons. We show in the Tables \ref{tab:ROBUST2}, \ref{tab:ROBUST1} the measures Mean, Standard Deviation (SD), Best and Worst for Damper with \linebreak$H = H_{\mathbf{x}} = H_{\mathrm{h}} = 2, 6, 10$ and for Drive with $H = H_{\mathbf{x}} = H_{\mathrm{h}} = 6, 10, 12$ of the RMSE of our methods. In Figure \ref{fig:ROBUST2}, \ref{fig:ROBUST1} we show the box-plots of the RMSE values. For all experiments we used the same setting as described in Section \ref{sec:NonlinearSystemIdentification}.\\ 
\\

\subsubsection{Discussion of the Robustness}

For the data-set Drive we can see for both our methods good reproducing properties for $H= 10, 12$. The time horizons, where our methods performs best, seem to be between $H=10$ and $H=12$. The reproducing properties of DRGP-VSS, in terms of close or even better to the value in Table \ref{tab:RMSERMSE}, is significantly better for this data-set compared to DRGP-SS. DRGP-VSS is better in terms of finding a better hidden state representation, which is less prone to over-fitting. This can been seen from the learned hidden state representations that are not smooth or far away from the initialization hidden state representations $\mathbf{y}$. DRGP-Nyström has similar robustness properties as DRGP-VSS, but clearly is outperformed by DRGP-VSS. For this chosen setting, DRGP-Nyström received not as good results as in Table \ref{tab:RMSERMSE}, where the result was taken from~\citep{mattos2015recurrent}.
For the Damper data-set we can see that DRGP-SS is more stable than DRGP-VSS in terms of higher time horizons, whereas it is more unstable for smaller ones. This can also be seen in Figure \ref{fig:VIS2} (c). This might be caused again by the additional regularization properties in terms of the extra variational spectral point variance and the KL-term for DRGP-VSS. For the other time horizons we can see similar robustness quality which is good in terms of the standard deviation. DRGP-Nyström shows similar reproducing properties on this data-set as our methods for higher horizons and better ones for smaller horizons.

\section{Conclusion}
In this article we introduced two new DRGPs based on the SS approximation introduced by~\citep{quia2010sparse} and the improved VSS approximation by~\citep{gal2015improving}. We also integrated variationally over the input-space and proved the existence of an optimal variational distribution for $\boldsymbol{A}$. We could show that our methods gave better results on nearly all data-sets used in this article compared to DRGP-Nyström from~\citep{mattos2015recurrent}, whereas DRGP-Nyström seems to be more stable. We could also show better results on most data-sets compared to GP-LSTM~\citep{al2016learning}. We showed how to reduce the training time complexity to $\mathcal O(M^3)$ with DVI when having the explicit gradients. For the future, we plan to implement our methods with DVI for applications on big data-sets. Our method is directly applicable for the multi-output case $\mathbf{Y}\in\mathbb{R}^{N\times D}$. Further research goes in the directions of testing other initialization techniques or settings for the overall optimization problem. Normalization techniques for the hidden state as well as specific priors could give additional boost in performance and robustness. One interesting technique is the variational resampling technique used in~\citep{cutajar2016practical}. This would also work for our methods. Another interesting topic would be mixing our methods with the DRGP-Nyström by using for specific layers different sparse approximations, to get overall better and more robust results. Furthermore,~\citep{hoang2017generalized} introduced an improved version of the VSS approximation, which might be adaptable for our case. Our sparse approximation in Section \ref{sec:SparseSpectrumGaussianProcessandVariationalInference} is also practical for dimensionality reduction as shown in~\citep{titsias2010bayesian} and can be further expanded to a deep version in this application. 


\begin{acknowledgements}
We would like to acknowledge support for this project
from the ETAS GmbH.
\end{acknowledgements}


\clearpage

\appendix
\section{Appendix}
\subsection{Kullback Leibler divergence}
\label{sec:KullbackLeiblerdivergence}
Assume we have a set of random vectors $\boldsymbol{X}\in\mathbb R^{N\times Q}$. For two continuous densities $q_{\boldsymbol{X}}$ and $p_{\boldsymbol{X}}$ the Kullback-Leibler divergence of $p_{\mathbf{X}}$ from $q_{\mathbf{X}}$ is given by
\begin{align*}
\mathbf{KL}(q_{\boldsymbol{X}}||p_{\boldsymbol{X}})=\int q_{\boldsymbol{X}}(\mathbf{X})\log\left(\frac{q_{\boldsymbol{X}}(\mathbf{X})}{p_{\boldsymbol{X}}(\mathbf{X})}\right)d\mathbf{X}.
\end{align*}
For two multivariate Gaussian densities $q_{\boldsymbol{X}}$, where $\boldsymbol{X}\sim\prod_{j=1}^Q\mathcal{N}(\Gamma_j,\Delta_j)$, and $p_{\boldsymbol{X}}$, where $\boldsymbol{X} \sim\prod_{j=1}^Q\mathcal{N}(\Xi_j,\Pi_j)$ we have (see~\citep[Appendix 1. Equation A.2]{damianou2015deep})
\begin{align*}
\mathbf{KL}(q_{\boldsymbol{X}}||p_{\boldsymbol{X}}) &= \frac{1}{2}\sum\limits_{j=1}^Q\log(\left|\Pi_j\right|)+\frac{1}{2}\sum\limits_{j=1}^Q\text{tr}\left(\Pi_j^{-1}\Delta_j+(\Xi_j-\Gamma_j)^T\Pi_j^{-1}(\Xi_j-\Gamma_j)\right)\\
& - \frac{1}{2}\sum\limits_{j=1}^Q\log(\left|\Delta_j\right|) - \frac{NQ}{2}.
\end{align*}
For the specific cases in Section \ref{sec:VariationalInferencerevisited}, \ref{sec:VariationalInferenceforDGP} we can derive
\begin{align*}
\mathbf{KL}(q_{\boldsymbol{A}}||p_{\boldsymbol{A}}) =& \frac{1}{2}\sum\limits_{d=1}^D\text{tr}\left(\mathbf{s}_d+\mathbf{m}_d\mathbf{m}_d^T\right) - \frac{1}{2}\sum\limits_{d=1}^D\log(\left|\mathbf{s}_d\right|) - \frac{DM}{2},\numberthis\label{KLA}\\
\mathbf{KL}(q_{\boldsymbol\omega}||p_{\boldsymbol\omega}) =& \frac{1}{2}\sum\limits_{m=1}^M\text{tr}\left(\boldsymbol\upbeta_m+\boldsymbol\upalpha_m\boldsymbol\upalpha_m^T\right) - \frac{1}{2}\sum\limits_{m=1}^M\log\left(|\boldsymbol\upbeta_m\right|) - \frac{MQ}{2},\\
\mathbf{KL}(q_{\boldsymbol{H}}||p_{\boldsymbol{H}}) =& \frac{1}{2}\sum\limits_{i=1}^N\text{tr}\left(\boldsymbol\uplambda_i+\boldsymbol\upmu_i\boldsymbol\upmu_i^T\right) - \frac{1}{2}\sum\limits_{i=1}^N\log(\left|\boldsymbol\uplambda_i\right|) - \frac{NQ}{2}.
\end{align*}
\subsection{Variational Inference methodology}
\label{sec:VariationalInferencemethodology}
Let $\mathbf{Y}\in\mathbb R^{N\times D}$ be a set of observations and $\boldsymbol{X}$ a set of random vectors as in \ref{sec:KullbackLeiblerdivergence}. The observations as well as the samples of $\boldsymbol{X}$ are generated by $p_{\boldsymbol{Y},\boldsymbol{X}}$. The log ML is given by
\begin{align*}
\log(p_{\boldsymbol{Y}}(\mathbf{Y})) = \log\left(\int\limits p_{\boldsymbol{Y},\boldsymbol{X}}(\mathbf{Y},\mathbf{X}) d\mathbf{X}\right) = \log\left(\int\limits p_{\boldsymbol{Y}|\boldsymbol{X}}(\mathbf{Y}|\mathbf{X})p_{\boldsymbol{X}}(\mathbf{X}) d\mathbf{X}\right)
\end{align*}
and it is approximated via
\begin{align*}
\log(p_{\boldsymbol{Y}}(\mathbf{Y})) & = \log\left(\int\limits p_{\boldsymbol{Y}|\boldsymbol{X}}(\mathbf{Y}|\mathbf{X})p_{\boldsymbol{X}}(\mathbf{X}) d\mathbf{X}\right)\\
& = \log\left(\int\limits \frac{q_{\boldsymbol{X}}(\mathbf{X})}{q_{\boldsymbol{X}}(\mathbf{X})}p_{\boldsymbol{Y}|\boldsymbol{X}}(\mathbf{Y}|\mathbf{X})p_{\boldsymbol{X}}(\mathbf{X}) d\mathbf{X}\right)\\
& \stackrel{\mathrm{JI}}\geq \int\limits q_{\boldsymbol{X}}(\mathbf{X})\log\left(\frac{p_{\boldsymbol{Y}|\boldsymbol{X}}(\mathbf{Y}|\mathbf{X})p_{\boldsymbol{X}}(\mathbf{X})}{q_{\boldsymbol{X}}(\mathbf{X})}\right) d\mathbf{X}\\
& = \int\limits q_{\boldsymbol{X}}(\mathbf{X})\log\left(p_{\boldsymbol{Y}|\boldsymbol{X}}(\mathbf{Y}|\mathbf{X})\right) d\mathbf{X} - \int\limits q_{\boldsymbol{X}}(\mathbf{X})\log\left(\frac{q_{\boldsymbol{X}}(\mathbf{X})}{p_{\boldsymbol{X}}(\mathbf{X})}\right) d\mathbf{X}\\
& = E[\mathcal{G}]_{q_{\boldsymbol{X}}} - \mathbf{KL}(q_{\boldsymbol{X}}||p_{\boldsymbol{X}})
\end{align*}
for a defined variational distribution in terms of $q_{\mathbf{X}}$ and $\mathcal{G}\stackrel{\mathrm{def}}=\log\left(p_{\boldsymbol{Y}|\boldsymbol{X}}(\mathbf{Y}|\mathbf{X})\right)$. In JI we used Jensen's inequality. We show next the equivalence of maximizing the log lower bound with respect to the variational parameters and the minimization of the Kullback Leibler divergence $\mathbf{KL}(q_{\boldsymbol{X}}||p_{\boldsymbol{X}|\boldsymbol{Y}})$.
We then use $q_{\boldsymbol{X}}$ as approximation to the true posterior $p_{\boldsymbol{X}|\boldsymbol{Y}}$. 
\begin{align*}
E[\mathcal{G}]_{q_{\boldsymbol{X}}} - \mathbf{KL}(q_{\boldsymbol{X}}||p_{\boldsymbol{X}}) & = \int\limits q_{\boldsymbol{X}}(\mathbf{X})\log\left(\frac{p_{\boldsymbol{Y}|\boldsymbol{X}}(\mathbf{Y}|\mathbf{X})p_{\boldsymbol{X}}(\mathbf{X})}{q_{\boldsymbol{X}}(\mathbf{X})}\right) d\mathbf{X}\\
& = \int\limits q_{\boldsymbol{X}}(\mathbf{X})\log\left(\frac{p_{\boldsymbol{X}|\boldsymbol{Y}}(\mathbf{X}|\mathbf{Y})p_{\boldsymbol{Y}}(\mathbf{Y})}{q_{\boldsymbol{X}}(\mathbf{X})}\right) d\mathbf{X}\\
& = \int\limits q_{\boldsymbol{X}}(\mathbf{X})\log(p(\mathbf{Y})) d\mathbf{X} - \int\limits q_{\boldsymbol{X}}(\mathbf{X})\log\left(\frac{q_{\boldsymbol{X}}(\mathbf{X})}{p_{\boldsymbol{X}|\boldsymbol{Y}}(\mathbf{X}|\mathbf{Y})}\right) d\mathbf{X}\\
& = \log(p(\mathbf{Y})) - \mathbf{KL}(q_{\boldsymbol{X}}||p_{\boldsymbol{X}|\boldsymbol{Y}}).
\end{align*}
Assume we have $q_{\boldsymbol{X}} = p_{\boldsymbol{X}|\boldsymbol{Y}}$. Then we see that the bound from above is sharp.

\subsection{Computation of the statistics $\Psi$}
\label{sec:Computationofthestatistics}

In the following we use the abbreviations
\begin{itemize}
	\item B.1,~\citep[see Section~4.1]{gal2015improving},
	\item B.2,~\citep[see A.7.]{rasmussen2006gaussian},
	\item B.3,~\citep[see A.3.]{gal2015improving} with $\sin(x) = \cos(\frac{\pi}{2} + x)$,
\end{itemize}
AT1 means Addition Theorem 1, which is
\begin{align*}
&\sin(x+a)\sin(y+c) -\sin(x+a)\sin(y+d)\\
&-\sin(x+b)\sin(y+c)+\sin(x+b)\sin(y+d)\\
&\stackrel{\mathrm{AT1}}=\frac{1}{2}(\cos(a-c+x-y)-\cos(a+c+x+y)\\
&-\cos(a-d+x-y)+\cos(a+d+x+y)\\
&-\cos(b-c+x-y)+\cos(b+c+x+y)\\
&+\cos(b-d+x-y)-\cos(b+d+x+y)),
\end{align*}
AT2 means Addition Theorem 2, which is
\begin{align*}
\frac{1}{2}(\cos(a-b+x-y)+\cos(a+b+x+y))\stackrel{\mathrm{AT2}}=\cos(x+a)\cos(y+b).
\end{align*}
What follows are three different versions of the statistics $\Psi_1, \Psi_2$, where we begin with the most general version. After this we show the version for DRGP-VSS and last the version for DRGP-SS. Therefore we also use the notation $\mathcal{N}_{\mathbf{h}_n}$ for a Gaussian density belonging to the random vector ${\boldsymbol{h}_n}$ and filled in the data $\mathbf{h}_n$ and $\mathcal{N}_{\boldsymbol{h}_n}$ just the density belonging to the random vector ${\boldsymbol{h}_n}$.

\subsubsection{General case of the statistics $\Psi$}
\label{Generalcase}

\noindent The statistics $\Psi_1 = \mathbf{E}\left[\Phi\right]_{q_{\boldsymbol\omega}q_{\mathbf{H}}}\in\mathbb R^{N\times M}$ and $\Psi_2 = \mathbf{E}\left[\Phi^T\Phi\right]_{q_{\boldsymbol\omega}q_{\mathbf{H}}}\in\mathbb R^{M\times M}$ are calculated as
\begin{align*}
(\Psi_1)_{nm}& = \mathbf{E}\left[\Phi\right]_{q_{\boldsymbol\omega_m}q_{\boldsymbol{h}_n}}\\
& \stackrel{\mathrm{B.1}}= \mathbf{E}\left[\sqrt{\frac{2\upsigma_{\text{power}}^2}{M(\updelta_m-\upgamma_m)}}e^{-\frac{1}{2}\bar{\mathbf{h}}_{nm}^T\boldsymbol\upbeta_m\bar{\mathbf{h}}_{nm}}(\sin(\hat{\boldsymbol\upalpha}_m^T(\mathbf{h}_n-\mathbf{u}_m)+\updelta_m)\right.\\
& \left.-\sin(\hat{\boldsymbol\upalpha}_m^T(\mathbf{h}_n-\mathbf{u}_m)+\upgamma_m))\right]_{q_{\boldsymbol{h}_n}}\\
& = \sqrt{\frac{2(2\pi)^Q\upsigma_{\text{power}}^2\prod\limits_{q=1}^Q\left(\frac{\mathrm{l}_q^2}{\upbeta_{m_q}}\right)}{M(\updelta_m-\upgamma_m)}}\int\limits\mathcal{N}_{\mathbf{h}_n}(\mathbf{u}_m,(2\pi)^{-2}\mathfrak{L}^2\boldsymbol\upbeta_m^{-1})\mathcal{N}_{\mathbf{h}_n}(\boldsymbol\upmu_n,\boldsymbol\uplambda_n)\\
& (\sin(\hat{\boldsymbol\upalpha}_m^T(\mathbf{h}_n-\mathbf{u}_m)+\updelta_m)-\sin(\hat{\boldsymbol\upalpha}_m^T(\mathbf{h}_n-\mathbf{u}_m)+\upgamma_m))d\mathbf{h}_n\\
& \stackrel{\mathrm{B.2}}= \sqrt{\frac{2\upsigma_{\text{power}}^2\prod\limits_{q=1}^Q\left(\frac{\mathrm{l}_q^2}{\upbeta_{m_q}}\right)}{M(\updelta_m-\upgamma_m)}}\int\limits Z_{nm}\mathcal{N}_{\mathbf{h}_n}(\mathbf{c}_{nm},\mathrm{C}_{nm})\\
& (\sin(\hat{\boldsymbol\upalpha}_m^T(\mathbf{h}_n-\mathbf{u}_m)+\updelta_m)-\sin(\hat{\boldsymbol\upalpha}_m^T(\mathbf{h}_n-\mathbf{u}_m)+\upgamma_m))d\mathbf{h}_n\\
& = \sqrt{\frac{2\upsigma_{\text{power}}^2\prod\limits_{q=1}^Q\left(\frac{\mathrm{l}_q^2}{\upbeta_{m_q}}\right)}{M(\updelta_m-\upgamma_m)}}Z_{nm}\mathbf{E}\left[\sin(\hat{\boldsymbol\upalpha}_m^T(\mathbf{h}_n-\mathbf{u}_m)+\updelta_m)\right.\\
& \left.-\sin(\hat{\boldsymbol\upalpha}_m^T(\mathbf{h}_n-\mathbf{u}_m)+\upgamma_m)\right]_{\mathcal{N}_{\boldsymbol{h}_n}(\mathbf{c}_{nm},\mathrm{C}_{nm})}\\
& \stackrel{\mathrm{B.3}}= \mathfrak{S}_m^1 Z_{nm}e^{-\frac{1}{2}\hat{\boldsymbol\upalpha}_m^T \mathrm{C}_{nm}\hat{\boldsymbol\upalpha}_m}\left(\sin(\hat{\boldsymbol\upalpha}_m^T(\mathbf{c}_{nm}-\mathbf{u}_m)+\updelta_m)\right. \\
& \left.-\sin(\hat{\boldsymbol\upalpha}_m^T(\mathbf{c}_{nm}-\mathbf{u}_m)+\upgamma_m)\right),
\end{align*}
for $m, = 1,\dots, M$, $n, = 1,\dots, N$ with 
\begin{align*}
\bar{\mathbf{h}}_{nm} &= 2\pi \mathfrak{L}^{-1}(\mathbf{h}_n-\mathbf{u}_m),\\
\hat{\boldsymbol\upalpha}_m &= 2\pi(\mathfrak{L}^{-1}\boldsymbol\upalpha_m+\mathbf{p}),\\
\mathbf{c}_{nm} &= \mathrm{C}_{nm}(\boldsymbol\upbeta_m (2\pi)^2\mathfrak{L}^{-2}\mathbf{u}_m + \boldsymbol\uplambda_n^{-1}\boldsymbol\upmu_n),\\
\mathrm{C}_{nm} &= (\boldsymbol\upbeta_m(2\pi)^2\mathfrak{L}^{-2} + \boldsymbol\uplambda_n^{-1})^{-1},\\
\mathbf{v}_{nm} &= \mathbf{u}_m-\boldsymbol\upmu_n,\\
\mathrm{V}_{nm} &= (2\pi)^{-2}\mathfrak{L}^2\boldsymbol\upbeta_m^{-1}  + \boldsymbol\uplambda_n,\\
\mathrm{Z}_{nm} &= \frac{1}{\sqrt{\left|\mathrm{V}_{nm}\right|}}e^{-\frac{1}{2}\mathbf{v}_{nm}^T \mathrm{V}_{nm}^{-1}\mathbf{v}_{nm}},\\
\mathfrak{S}_m^1&= \sqrt{\frac{2\upsigma_{\text{power}}^2\prod\limits_{q=1}^{Q}\left(\frac{\mathrm{l}_q^2}{\upbeta_{m_q}}\right)}{M(\updelta_m-\upgamma_m)}}.
\end{align*}
\noindent$\Psi_2 = \sum\limits_{n=1}^N \Psi_2^n$, where
\begin{align*}
(\Psi_2^n)_{mm'}& = \mathbf{E}\left[\Phi_{mn}^T\Phi_{nm'}\right]_{q_{\boldsymbol\omega_m}q_{\boldsymbol\omega_{m'}}q_{\boldsymbol{h}_n}}\\
& \stackrel{\mathrm{B.1}}= \mathbf{E}\left[\frac{2\upsigma_{\text{power}}^2}{M(\updelta_m-\upgamma_m)^2(\updelta_{m'}-\upgamma_{m'})^2}e^{-\frac{1}{2}(\bar{\mathbf{h}}_{nm}^T\boldsymbol\upbeta_m\bar{\mathbf{h}}_{nm}+\bar{\mathbf{h}}_{nm'}^T\boldsymbol\upbeta_{m'}\bar{\mathbf{h}}_{nm'})}\right.\\
& \left.(\sin(\hat{\boldsymbol\upalpha}_m^T(\mathbf{h}_n-\mathbf{u}_m)+\updelta_m)-\sin(\hat{\boldsymbol\upalpha}_m^T(\mathbf{h}_n-\mathbf{u}_m)+\upgamma_m))\right.\\
& \left.(\sin(\hat{\boldsymbol\upalpha}_{m'}^T(\mathbf{h}_n-\mathbf{u}_{m'})+\updelta_{m'})-\sin(\hat{\boldsymbol\upalpha}_{m'}^T(\mathbf{h}_n-\mathbf{u}_{m'})+\upgamma_{m'}))\right]_{q_{\boldsymbol{h}_n}}\\
& \stackrel{\mathrm{AT1}}= \frac{(2\pi)^Q\upsigma_{\text{power}}^2\prod\limits_{q=1}^Q\left(\frac{\mathrm{l}_q^2}{\sqrt{\upbeta_{m_q}\upbeta_{m'_q}}}\right)}{M(\updelta_m-\upgamma_m)^2(\updelta_{m'}-\upgamma_{m'})^2}\\
& \int\limits \mathcal{N}_{\mathbf{h}_n}(\mathbf{u}_m,(2\pi)^{-2}\mathfrak{L}^2\boldsymbol\upbeta_m^{-1})\mathcal{N}_{\mathbf{h}_n}(\mathbf{u}_{m'},(2\pi)^{-2}\mathfrak{L}^2\boldsymbol\upbeta_{m'}^{-1})\mathcal{N}_{\mathbf{h}_n}(\boldsymbol\upmu_n,\boldsymbol\uplambda_n)\\
&(\cos(\bar{\boldsymbol\upalpha}_{mm'}^T\mathbf{h}_n - \bar{\uptau}_{mm'} + \bar{\updelta}_{mm'})-\cos(\overset{+}{\boldsymbol\upalpha}_{mm'}^T\mathbf{h}_n -  \overset{+}{\uptau}_{mm'} + \overset{+}{\updelta}_{mm'})\\
&  - \cos(\bar{\boldsymbol\upalpha}_{mm'}^T\mathbf{h}_n - \bar{\uptau}_{mm'} + \bar{\upiota}_{mm'})+\cos(\overset{+}{\boldsymbol\upalpha}_{mm'}^T\mathbf{h}_n - \overset{+}{\uptau}_{mm'} + \overset{+}{\upiota}_{mm'})\\
& - \cos(\bar{\boldsymbol\upalpha}_{mm'}^T\mathbf{h}_n - \bar{\uptau}_{mm'} + \bar{\upiota}_{m'm})+\cos(\overset{+}{\boldsymbol\upalpha}_{mm'}^T\mathbf{h}_n - \overset{+}{\uptau}_{mm'} + \overset{+}{\upiota}_{m'm})\\
& +\cos(\bar{\boldsymbol\upalpha}_{mm'}^T\mathbf{h}_n - \bar{\uptau}_{mm'} + \bar{\upgamma}_{mm'})-\cos(\overset{+}{\boldsymbol\upalpha}_{mm'}^T\mathbf{h}_n - \overset{+}{\uptau}_{mm'} + \overset{+}{\upgamma}_{mm'}))d\mathbf{h}_n\\
& \stackrel{\mathrm{B.2}}= \frac{\upsigma_{\text{power}}^2\prod\limits_{q=1}^Q\left(\frac{\mathrm{l}_q^2}{\sqrt{\upbeta_{m_q}\upbeta_{m'_q}}}\right)}{M(\updelta_m-\upgamma_m)^2(\updelta_{m'}-\upgamma_{m'})^2}\int\limits \mathrm{Z}_{mm'}^n \mathcal{N}_{\mathbf{h}_n}(\mathbf{d}_{mm'}^{n},\mathrm{D}_{mm'}^{n})\overline\cos(\mathbf{h}_n)d\mathbf{h}_n\\
& = \frac{\upsigma_{\text{power}}^2\prod\limits_{q=1}^Q\left(\frac{\mathrm{l}_q^2}{\sqrt{\upbeta_{m_q}\upbeta_{m'_q}}}\right)}{M(\updelta_m-\upgamma_m)^2(\updelta_{m'}-\upgamma_{m'})^2}\mathrm{Z}_{mm'}^n\mathbf{E}\left[\overline\cos(\mathbf{h}_n)\right]_{\mathcal{N}_{\boldsymbol{h}_n}(\mathbf{d}_{mm'}^n,\mathrm{D}_{mm'}^n)}\\
& \stackrel{\mathrm{B.3}}= \mathfrak{S}_m^2\mathrm{Z}_{mm'}^n \left(e^{-\frac{1}{2}{\bar{\boldsymbol\upalpha}_{mm'}}^T {\mathrm{D}_{mm'}^n}\bar{\boldsymbol\upalpha}_{mm'}}\left(\cos(\bar{\boldsymbol\upalpha}_{mm'}^T\mathbf{d}_{mm'}^n - \bar{\uptau}_{mm'} + \bar{\updelta}_{mm'})\right.\right.\\
& \left.\left.-\cos(\bar{\boldsymbol\upalpha}_{mm'}^T\mathbf{d}_{mm'}^n - \bar{\uptau}_{mm'} + \bar{\upiota}_{mm'})-\cos(\bar{\boldsymbol\upalpha}_{mm'}^T\mathbf{d}_{mm'}^n - \bar{\uptau}_{mm'} + \bar{\upiota}_{m'm})\right.\right.\\
& \left.\left.+\cos(\bar{\boldsymbol\upalpha}_{mm'}^T\mathbf{d}_{mm'}^n - \bar{\uptau}_{mm'} + \bar{\upgamma}_{mm'})\right)\right.\\
&\left.-e^{-\frac{1}{2}{\overset{+}{\boldsymbol\upalpha}_{mm'}^T} {\mathrm{D}_{mm'}^n}\overset{+}{\boldsymbol\upalpha}_{mm'}}\left(\cos(\overset{+}{\boldsymbol\upalpha}_{mm'}^T\mathbf{d}_{mm'}^n -\overset{+}{\uptau}_{mm'} + \overset{+}{\updelta}_{mm'})\right.\right.\\
&\left.\left.-\cos(\overset{+}{\boldsymbol\upalpha}_{mm'}^T\mathbf{d}_{mm'}^n - \overset{+}{\uptau}_{mm'} + \overset{+}{\upiota}_{mm'})-\cos(\overset{+}{\boldsymbol\upalpha}_{mm'}^T\mathbf{d}_{mm'}^n - \overset{+}{\uptau}_{mm'} + \overset{+}{\upiota}_{m'm})\right.\right.\\
&\left.\left.+\cos(\overset{+}{\boldsymbol\upalpha}_{mm'}^T\mathbf{d}_{mm'}^n - \overset{+}{\uptau}_{mm'} + \overset{+}{\upgamma}_{mm'})\right)\right),
\end{align*}
for $m,m' = 1,\dots, M$, $m\neq m'$, with
\begin{align*}
\overline\cos(\mathbf{h}_n)&\stackrel{\mathrm{def}}= \cos(\bar{\boldsymbol\upalpha}_{mm'}^T\mathbf{h}_n - \bar{\uptau}_{mm'} + \bar{\updelta}_{mm'})-\cos(\overset{+}{\boldsymbol\upalpha}_{mm'}^T\mathbf{h}_n +  \overset{+}{\uptau}_{mm'} + \overset{+}{\updelta}_{mm'})\\
& - \cos(\bar{\boldsymbol\upalpha}_{mm'}^T\mathbf{h}_n - \bar{\uptau}_{mm'} + \bar{\upiota}_{mm'}) +\cos(\overset{+}{\boldsymbol\upalpha}_{mm'}^T\mathbf{h}_n + \overset{+}{\uptau}_{mm'} + \overset{+}{\upiota}_{mm'})\\
& - \cos(\bar{\boldsymbol\upalpha}_{mm'}^T\mathbf{h}_n - \bar{\uptau}_{mm'} + \bar{\upiota}_{m'm})+\cos(\overset{+}{\boldsymbol\upalpha}_{mm'}^T\mathbf{h}_n + \overset{+}{\uptau}_{mm'} + \overset{+}{\upiota}_{m'm})\\
& +\cos(\bar{\boldsymbol\upalpha}_{mm'}^T\mathbf{h}_n - \bar{\uptau}_{mm'} + \bar{\upgamma}_{mm'})-\cos(\overset{+}{\boldsymbol\upalpha}_{mm'}^T\mathbf{h}_n+ \overset{+}{\uptau}_{mm'} + \overset{+}{\upgamma}_{mm'}),
\end{align*}
\noindent and
\begin{align*}
\bar{\updelta}_{mm'} &= \updelta_m - \updelta_{m'},\\
\overset{+}{\updelta}_{mm'} &= \updelta_m + \updelta_{m'},\\
\bar{\upgamma}_{mm'} &= \upgamma_m - \upgamma_{m'},\\
\overset{+}{\upgamma}_{mm'} &= \upgamma_m + \upgamma_{m'},\\
\bar{\upiota}_{mm'} &= \updelta_m - \upgamma_{m'},\\
\overset{+}{\upiota}_{mm'} &= \updelta_m + \upgamma_{m'},\\
\bar{\upiota}_{m'm} &= \upgamma_m - \updelta_{m'},\\
\overset{+}{\upiota}_{m'm} &= \upgamma_m + \updelta_m,\\
\bar{\updelta}_{mm'} &= \updelta_m - \updelta_{m'},\\
\overset{+}{\updelta}_{mm'} &= \updelta_m + \updelta_{m'},\\
\bar{\uptau}_{mm'}&=\hat{\boldsymbol\upalpha}_m^T \mathbf{u}_m- \hat{\boldsymbol\upalpha}_{m'}^T \mathbf{u}_{m'},\\
\overset{+}{\uptau}_{mm'}&=\hat{\boldsymbol\upalpha}_m^T \mathbf{u}_m+ \hat{\boldsymbol\upalpha}_{m'}^T \mathbf{u}_{m'},\\
\bar{\boldsymbol\upalpha}_{mm'} &= \hat{\boldsymbol\upalpha}_m - \hat{\boldsymbol\upalpha}_{m'},\\
\overset{+}{\boldsymbol\upalpha}_{mm'} &= \hat{\boldsymbol\upalpha}_m + \hat{\boldsymbol\upalpha}_{m'},\\
\mathbf{b}_{mm'} &= \mathrm{B}_{mm'}(2\pi)^2\mathfrak{L}^{-2}(\boldsymbol\upbeta_m\mathbf{u}_m+\boldsymbol\upbeta_{m'}\mathbf{u}_{m'}),\\
\boldsymbol\upbeta_{mm'} &= \boldsymbol\upbeta_m + \boldsymbol\upbeta_{m'},\\
\mathrm{B}_{mm'} &= (2\pi)^{-2}\mathfrak{L}^2\boldsymbol\upbeta_{mm'}^{-1},\\
\mathbf{d}_{mm'}^n &= \mathrm{D}_{mm'}^n(\mathrm{B}_{mm'}^{-1}\mathbf{b}_{mm'}+\boldsymbol\uplambda_n^{-1}\boldsymbol\upmu_n),\\
\mathrm{D}_{mm'}^n &= (\mathrm{B}_{mm'}^{-1} + \boldsymbol\uplambda_n^{-1})^{-1},\\
\mathbf{w}_{mm'}^n &= \mathbf{b}_{mm'}-\boldsymbol\upmu_n,\\
\mathrm{W}_{mm'}^n &= \mathrm{B}_{mm'}  + \boldsymbol\uplambda_n,\\
\mathbf{u}_{mm'} &= \mathbf{u}_{m}-\mathbf{u}_{m'},\\
\mathbf{U}_{mm'} &= (2\pi)^{-2}\mathfrak{L}^2(\boldsymbol\upbeta_m^{-1}  + \boldsymbol\upbeta_{m'}^{-1}),\\
\mathrm{Z}_{mm'}^n &= \frac{1}{\sqrt{\left|\mathrm{W}_{mm'}^n\right|\left|\mathbf{U}_{mm'}\right|}}e^{-\frac{1}{2}({\mathbf{w}_{mm'}^n}^T{\mathrm{W}_{mm'}^n}^{-1}\mathbf{w}_{mm'}^n+\mathbf{u}_{mm'}^T\mathbf{U}_{mm'}^{-1}\mathbf{u}_{mm'})},\\
\mathfrak{S}_m^2&=\frac{\upsigma_{\text{power}}^2\prod\limits_{q=1}^Q\left(\frac{\mathrm{l}_q^2}{\sqrt{\upbeta_{m_q}\upbeta_{m'_q}}}\right)}{M(\updelta_m-\upgamma_m)^2(\updelta_{m'}-\upgamma_{m'})^2},
\end{align*}
\newpage
\noindent and 
\begin{align*}
(\Psi_2^n)_{mm}& = \mathbf{E}\left[\Phi_{mn}^T\Phi_{nm}\right]_{q_{\boldsymbol\omega_m}q_{\boldsymbol{h}_n}}\\
& \stackrel{\mathrm{B.1}}= \mathbf{E}\left[\frac{2\upsigma_{\text{power}}^2}{M}(\frac{1}{2}+\frac{1}{4}e^{-2\bar{\mathbf{h}}_{nm}^T\boldsymbol\upbeta_m\bar{\mathbf{h}}_{nm}}\right.\\
& \left.(\sin(2(\hat{\boldsymbol\upalpha}_m^T(\mathbf{h}_n-\mathbf{u}_m)+\updelta_m))-\sin(2(\hat{\boldsymbol\upalpha}_m^T(\mathbf{h}_n-\mathbf{u}_m)+\upgamma_m)))\right]_{q_{\boldsymbol{h}_n}}\\
& \stackrel{\mathrm{B.1}}= \mathbf{E}\left[\frac{\upsigma_{\text{power}}^2}{M}(1-\frac{1}{2}e^{-2\bar{\mathbf{h}}_{nm}^T\boldsymbol\upbeta_m\bar{\mathbf{h}}_{nm}}\right.\\
& \left.(\cos(2(\hat{\boldsymbol\upalpha}_m^T(\mathbf{h}_n-\mathbf{u}_m)+\updelta_m)+\frac{\pi}{2})-\cos(2(\hat{\boldsymbol\upalpha}_m^T(\mathbf{h}_n-\mathbf{u}_m)+\upgamma_m)+\frac{\pi}{2}))\right]_{q_{\boldsymbol{h}_n}}\\
&= \frac{\upsigma_{\text{power}}^2}{M} - \frac{\upsigma_{\text{power}}^2\sqrt{(2\pi)^Q\prod\limits_{q=1}^Q\left(\frac{\mathrm{l}_q^2}{\upbeta_{m_q}}\right)}}{2^{Q+1}M}\\
&\int\limits\mathcal{N}_{\mathbf{h}_n}(\mathbf{u}_m,2^{-2}(2\pi)^{-2}\mathfrak{L}^2\boldsymbol\upbeta_m^{-1})\mathcal{N}_{\mathbf{h}_n}(\boldsymbol\upmu_n,\boldsymbol\uplambda_n)\\
& \left.(\cos(2(\hat{\boldsymbol\upalpha}_m^T(\mathbf{h}_n-\mathbf{u}_m)+\updelta_m)+\frac{\pi}{2})-\cos(2(\hat{\boldsymbol\upalpha}_m^T(\mathbf{h}_n-\mathbf{u}_m)+\upgamma_m)+\frac{\pi}{2}))\right]_{q_{\boldsymbol{h}_n}}\\
& \stackrel{\mathrm{B.2}}= \frac{\upsigma_{\text{power}}^2}{M} \left(1-\frac{\sqrt{\prod\limits_{q=1}^Q\left(\frac{\mathrm{l}_q^2}{\upbeta_{m_q}}\right)}}{2^{Q+1}}\int\limits \tilde{\mathrm{Z}}_{nm}\mathcal{N}_{\mathbf{h}_n}(\tilde{\mathbf{c}}_{nm},\tilde{\mathrm{C}}_{nm})\right.\\
& \left.(\cos(2(\hat{\boldsymbol\upalpha}_m^T(\mathbf{h}_n-\mathbf{u}_m)+\updelta_m)+\frac{\pi}{2})-\cos(2(\hat{\boldsymbol\upalpha}_m^T(\mathbf{h}_n-\mathbf{u}_m)+\upgamma_m)+\frac{\pi}{2}))\right]_{q_{\boldsymbol{h}_n}}\\
& = \frac{\upsigma_{\text{power}}^2}{M} \left(1-\frac{\sqrt{\prod\limits_{q=1}^Q\left(\frac{\mathrm{l}_q^2}{\upbeta_{m_q}}\right)}}{2^{Q+1}}\tilde{\mathrm{Z}}_{nm}\right.\\
&\left.\mathbf{E}\left[\cos(2(\hat{\boldsymbol\upalpha}_m^T(\mathbf{h}_n-\mathbf{u}_m)+\updelta_m)+\frac{\pi}{2})\right.\right.\\
&\left.\left.-\cos(2(\hat{\boldsymbol\upalpha}_m^T(\mathbf{h}_n-\mathbf{u}_m)+\upgamma_m)+\frac{\pi}{2})\right]_{\mathcal{N}_{\boldsymbol{h}_n}(\tilde{\mathbf{c}}_{nm},\tilde{\mathrm{C}}_{nm})}\right)\\
& \stackrel{\mathrm{B.3}}= \frac{\upsigma_{\text{power}}^2}{M} \left(1-\tilde{\mathfrak{S}}_m^2 \tilde{\mathrm{Z}}_{nm}(e^{-2\hat{\boldsymbol\upalpha}_m^T\tilde{\mathrm{C}}_{nm}\hat{\boldsymbol\upalpha}_m}(\cos(2(\hat{\boldsymbol\upalpha}_m^T(\mathbf{h}_n-\mathbf{u}_m)+\updelta_m)+\frac{\pi}{2})\right.\\
&\left.-\cos(2(\hat{\boldsymbol\upalpha}_m^T(\mathbf{h}_n-\mathbf{u}_m)+\upgamma_m)+\frac{\pi}{2})))\right),
\end{align*}
for $m,m' = 1,\dots, M$, $m=m'$, with
\begin{align*}
\tilde{\mathbf{c}}_{nm} &= \tilde{\mathrm{C}}_{nm}(\boldsymbol\upbeta_m 2^2(2\pi)^2\mathfrak{L}^{-2}\mathbf{u}_m + \boldsymbol\uplambda_n^{-1}\boldsymbol\upmu_n),\\
\tilde{\mathrm{C}}_{nm} &= (\boldsymbol\upbeta_m 2^2(2\pi)^2\mathfrak{L}^{-2} + \boldsymbol\uplambda_n^{-1})^{-1},\\
\tilde{\mathrm{V}}_{nm} &= 2^{-2}(2\pi)^{-2}\mathfrak{L}^2\boldsymbol\upbeta_m^{-1}  + \boldsymbol\uplambda_n,\\
\tilde{\mathrm{Z}}_{nm} &= \frac{1}{\sqrt{\left|\tilde{\mathrm{V}}_{nm}\right|}}e^{-\frac{1}{2}\mathbf{v}_{nm}^T \tilde{\mathrm{V}}_{nm}^{-1}\mathbf{v}_{nm}},\\
\tilde{\mathfrak{S}}_m^2 &=\frac{\sqrt{\prod\limits_{q=1}^Q\left(\frac{\mathrm{l}_q^2}{\upbeta_{m_q}}\right)}}{2^{Q+1}}.
\end{align*}

\subsubsection{DRGP-VSS in case of the statistics $\Psi$}

We denote a simplified version DRGP-VSS, where we randomized the phases $\mathrm{b}_m$. This means no VI for $\mathrm{b}_m$ and optimization of the parameters $\mathrm{b}_m$. We have $p_{\boldsymbol\omega_m} = p_{\boldsymbol{z}_m}$, $q_{\boldsymbol\omega_m} = q_{\boldsymbol{z}_m}$.
\begin{align*}
(\Psi_1)_{nm}& = \mathbf{E}\left[\Phi_{nm}\right]_{q_{\boldsymbol{z}_m}q_{\boldsymbol{h}_n}}\\
& \stackrel{\mathrm{B.1}}= \mathbf{E}\left[\sqrt{\frac{2\upsigma_{\text{power}}^2}{M}}e^{-\frac{1}{2}\bar{\mathbf{h}}_{nm}^T\boldsymbol\upbeta_m\bar{\mathbf{h}}_{nm}}\cos(\hat{\boldsymbol\upalpha}_m^T(\mathbf{h}_n-\mathbf{u}_m)+\mathrm{b}_m)\right]_{q_{\boldsymbol{h}_n}}\\
& = \sqrt{\frac{2(2\pi)^Q\upsigma_{\text{power}}^2\prod\limits_{q=1}^Q\left(\frac{\mathrm{l}_q^2}{\upbeta_{m_q}}\right)}{M}}\int\limits\mathcal{N}_{\mathbf{h}_n}(\mathbf{u}_m,(2\pi)^{-2}\mathfrak{L}^2\boldsymbol\upbeta_m^{-1})\mathcal{N}_{\mathbf{h}_n}(\boldsymbol\upmu_n,\boldsymbol\uplambda_n)\\
& \cos(\hat{\boldsymbol\upalpha}_m^T(\mathbf{h}_n-\mathbf{u}_m)+\mathrm{b}_m)d\mathbf{h}_n\\
& \stackrel{\mathrm{B.2}}= \sqrt{\frac{2\upsigma_{\text{power}}^2\prod\limits_{q=1}^Q\left(\frac{\mathrm{l}_q^2}{\upbeta_{m_q}}\right)}{M}}\int\limits \mathrm{Z}_{nm}\mathcal{N}_{\mathbf{h}_n}(\mathbf{c}_{nm},\mathrm{C}_{nm})\cos(\hat{\boldsymbol\upalpha}_m^T(\mathbf{h}_n-\mathbf{u}_m)+\mathrm{b}_m)d\mathbf{h}_n\\
& = \sqrt{\frac{2\upsigma_{\text{power}}^2\prod\limits_{q=1}^Q\left(\frac{\mathrm{l}_q^2}{\upbeta_{m_q}}\right)}{M}}\mathrm{Z}_{nm}\mathbf{E}\left[\cos(\hat{\boldsymbol\upalpha}_m^T(\mathbf{h}_n-\mathbf{u}_m)+\mathrm{b}_m)\right]_{\mathcal{N}_{\boldsymbol{h}_n}(\mathbf{c}_{nm},\mathrm{C}_{nm})}\\
& \stackrel{\mathrm{B.3}}= \Sigma_{m}^1 \mathrm{Z}_{nm}e^{-\frac{1}{2}\hat{\boldsymbol\upalpha}_m^T \mathrm{C}_{nm}\hat{\boldsymbol\upalpha}_m}\cos(\hat{\boldsymbol\upalpha}_m^T(\mathbf{c}_{nm}-\mathbf{u}_m)+\mathrm{b}_m),
\end{align*}
for $m, = 1,\dots, M$, $n, = 1,\dots, N$ with 
\begin{align*}
\Sigma_{m}^1 & = \sqrt{\frac{2\upsigma_{\text{power}}^2\prod\limits_{q=1}^Q\left(\frac{\mathrm{l}_q^2}{\upbeta_{m_q}}\right)}{M}},
\end{align*}
for the other variables, see the defined variables in the general case \ref{Generalcase}.\\
\\
\noindent$\Psi_2 = \sum\limits_{n=1}^N \Psi_2^n$ where
\begin{align*}
(\Psi_2^n)_{mm'}& = \mathbf{E}\left[\Phi_{mn}^T\Phi_{nm'}\right]_{q_{\boldsymbol\omega_m}q_{\boldsymbol\omega_{m'}}q_{\boldsymbol{h}_n}}\\
& \stackrel{\mathrm{B.1}}= \mathbf{E}\left[\frac{2\upsigma_{\text{power}}^2}{M}e^{-\frac{1}{2}(\bar{\mathbf{h}}_{nm}^T\boldsymbol\upbeta_m\bar{\mathbf{h}}_{nm}+\bar{\mathbf{h}}_{nm'}^T\boldsymbol\upbeta_{m'}\bar{\mathbf{h}}_{nm'})}\right.\\
&\left.\cos(\hat{\boldsymbol\upalpha}_m^T(\mathbf{h}_n-\mathbf{u}_m)+\mathrm{b}_m)\cos(\hat{\boldsymbol\upalpha}_{m'}^T(\mathbf{h}_n-\mathbf{u}_{m'})+\mathrm{b}_{m'})\right]_{q_{\boldsymbol{h}_n}}\\
& \stackrel{\mathrm{AT2}}= \frac{(2\pi)^Q\upsigma_{\text{power}}^2\prod\limits_{q=1}^Q\left(\frac{\mathrm{l}_q^2}{\sqrt{\upbeta_{m_q}\upbeta_{m'_q}}}\right)}{M}\\
&\int\limits \mathcal{N}_{\mathbf{h}_n}(\mathbf{u}_m,(2\pi)^{-2}\mathfrak{L}^2\boldsymbol\upbeta_m^{-1})\mathcal{N}_{\mathbf{h}_n}(\mathbf{u}_{m'},(2\pi)^{-2}\mathfrak{L}^2\boldsymbol\upbeta_{m'}^{-1})\mathcal{N}_{\mathbf{h}_n}(\boldsymbol\upmu_n,\boldsymbol\uplambda_n)\\
&(\cos(\bar{\boldsymbol\upalpha}_{mm'}^T\mathbf{h}_n - \bar{\uptau}_{mm'} + \bar{\mathrm{b}}_{mm'})+\cos(\overset{+}{\boldsymbol\upalpha}_{mm'}^T\mathbf{h}_n - \overset{+}{\uptau}_{mm'} +  \overset{+}{\mathrm{b}}_{mm'}))d\mathbf{h}_n\\
& \stackrel{\mathrm{B.2}}= \frac{\upsigma_{\text{power}}^2\prod\limits_{q=1}^Q\left(\frac{\mathrm{l}_q^2}{\sqrt{\upbeta_{m_q}\upbeta_{m'_q}}}\right)}{M}\int\limits \mathrm{Z}_{mm'}^n \mathcal{N}_{\mathbf{h}_n}(\mathbf{d}_{mm'}^{n},\mathrm{D}_{mm'}^{n})\\
& (\cos(\bar{\boldsymbol\upalpha}_{mm'}^T\mathbf{h}_n - \bar{\uptau}_{mm'} + \bar{\mathrm{b}}_{mm'})+\cos(\overset{+}{\boldsymbol\upalpha}_{mm'}^T\mathbf{h}_n - \overset{+}{\uptau}_{mm'} +  \overset{+}{\mathrm{b}}_{mm'}))d\mathbf{h}_n
\end{align*}
\begin{align*}
& = \frac{\upsigma_{\text{power}}^2\prod\limits_{q=1}^Q\left(\frac{\mathrm{l}_q^2}{\sqrt{\upbeta_{m_q}\upbeta_{m'_q}}}\right)}{M}\mathrm{Z}_{mm'}^n\mathbf{E}\left[\cos(\bar{\boldsymbol\upalpha}_{mm'}^T\mathbf{h}_n - \bar{\uptau}_{mm'} + \bar{\mathrm{b}}_{mm'})\right.\\
&\left.+\cos(\overset{+}{\boldsymbol\upalpha}_{mm'}^T\mathbf{h}_n - \overset{+}{\uptau}_{mm'} +  \overset{+}{\mathrm{b}}_{mm'})\right]_{\mathcal{N}_{\boldsymbol{h}_n}(\mathbf{d}_{mm'}^n,\mathrm{D}_{mm'}^n)}\\
& \stackrel{\mathrm{B.3}}= \Sigma_{mm'}^2\mathrm{Z}_{mm'}^n \left(e^{-\frac{1}{2}\bar{\boldsymbol\upalpha}_{mm'}^T {\mathrm{D}_{mm'}^n}\bar{\boldsymbol\upalpha}_{mm'}}\cos(\bar{\boldsymbol\upalpha}_{mm'}^T\mathbf{d}_{mm'}^n - \bar{\uptau}_{mm'} + \bar{\mathrm{b}}_{mm'})\right.\\
& \left. + e^{-\frac{1}{2}\overset{+}{\boldsymbol\upalpha}_{mm'}^T {\mathrm{D}_{mm'}^n}\overset{+}{\boldsymbol\upalpha}_{mm'}}\cos(\overset{+}{\boldsymbol\upalpha}_{mm'}^T\mathbf{d}_{mm'}^n - \overset{+}{\uptau}_{mm'} +  \overset{+}{\mathrm{b}}_{mm'})\right).
\end{align*}
for $m,m' = 1,\dots, M$, $m\neq m'$, with 
\begin{align*}
\Sigma_{mm'}^2 &= \frac{\upsigma_{\text{power}}^2\prod\limits_{q=1}^Q\left(\frac{\mathrm{l}_q^2}{\sqrt{\upbeta_{m_q}\upbeta_{{m'}_q}}}\right)}{M},\\
\overline{\mathrm{b}}_{mm'}&=\mathrm{b}_{m}-\mathrm{b}_{m'},\\
\overset{+}{\mathrm{b}}_{mm'}&=\mathrm{b}_{m}+\mathrm{b}_{m'},
\end{align*}
again, for the other variables, see the defined variables in the general case \ref{Generalcase}, and
\begin{align*}
(\Psi_2^n)_{mm}& = \mathbf{E}\left[\Phi_{mn}^T\Phi_{nm}\right]_{q_{\boldsymbol\omega_{m}}q_{\boldsymbol{h}_n}}\\
& \stackrel{\mathrm{B.1}}= \mathbf{E}\left[\frac{2\upsigma_{\text{power}}^2}{M}(\frac{1}{2}+\frac{1}{2}e^{-2\bar{\mathbf{h}}_{nm}^T\boldsymbol\upbeta_m\bar{\mathbf{h}}_{nm}}\cos(2(\hat{\boldsymbol\upalpha}_m^T(\mathbf{h}_n-\mathbf{u}_m)+\mathrm{b}_m))\right]_{q_{\boldsymbol{h}_n}}\\
& = \frac{\upsigma_{\text{power}}^2}{M} \left(1 + \frac{\sqrt{(2\pi)^Q\prod\limits_{q=1}^Q\left(\frac{\mathrm{l}_q^2}{\upbeta_{m_q}}\right)}}{2^Q}\int\limits\mathcal{N}_{\mathbf{h}_n}(\mathbf{u}_m,2^{-2}(2\pi)^{-2}\mathfrak{L}^2\boldsymbol\upbeta_m^{-1})\mathcal{N}_{\mathbf{h}_n}(\boldsymbol\upmu_n,\boldsymbol\uplambda_n)\right.\\
&\left. \cos(2(\hat{\boldsymbol\upalpha}_m^T(\mathbf{h}_n-\mathbf{u}_m)+\mathrm{b}_m))d\mathbf{h}_n\right)\\
& \stackrel{\mathrm{B.2}}= \frac{\upsigma_{\text{power}}^2}{M} \left(1+\frac{\sqrt{\prod\limits_{q=1}^Q\left(\frac{\mathrm{l}_q^2}{\upbeta_{m_q}}\right)}}{2^Q}\int\limits \tilde{\mathrm{Z}}_{nm}\mathcal{N}_{\mathbf{h}_n}(\tilde{\mathbf{c}}_{nm},\tilde{\mathrm{C}}_{nm})\right.\\
&\left.\cos(2(\hat{\boldsymbol\upalpha}_m^T(\mathbf{h}_n-\mathbf{u}_m)+\mathrm{b}_m))d\mathbf{h}_n\right)\\
& = \frac{\upsigma_{\text{power}}^2}{M} \left(1+\frac{\sqrt{\prod\limits_{q=1}^Q\left(\frac{\mathrm{l}_q^2}{\upbeta_{m_q}}\right)}}{2^Q}\tilde{\mathrm{Z}}_{nm}\mathbf{E}\left[\cos(2(\hat{\boldsymbol\upalpha}_m^T(\mathbf{h}_n-\mathbf{u}_m)+\mathrm{b}_m))\right]_{\mathcal{N}_{\boldsymbol{h}_n}(\tilde{\mathbf{c}}_{nm},\tilde{\mathrm{C}}_{nm})}\right)\\
& \stackrel{\mathrm{B.3}}= \frac{\upsigma_{\text{power}}^2}{M} \left(1+\tilde{\Sigma}_m^2 \tilde{\mathrm{Z}}_{nm}e^{-2\hat{\boldsymbol\upalpha}_m^T\tilde{\mathrm{C}}_{nm}\hat{\boldsymbol\upalpha}_m}\cos(2(\hat{\boldsymbol\upalpha}_m^T(\tilde{\mathbf{c}}_{nm}-\mathbf{u}_m)+\mathrm{b}_m))\right),
\end{align*}
for $m,m' = 1,\dots, M$, $m=m'$, with
\begin{align*}
\tilde{\Sigma}_m^2&= \frac{\sqrt{\prod\limits_{q=1}^Q\left(\frac{\mathrm{l}_q^2}{\upbeta_{m_q}}\right)}}{2^Q},
\end{align*}
again, for the other variables, see the defined variables in the general case \ref{Generalcase}.

\subsubsection{DRGP-SS in case of the statistics $\Psi$}
We denote a simplified version DRGP-SS, where we also randomize the spectral points and optimize over $\mathbf{z}_m$. In the Equations \eqref{identity000}, \eqref{identity0} the integration over $\text{\textbf{\textomega}}$, as well as $p_{\boldsymbol\omega}(\text{\textbf{\textomega}})$, vanishes and so $\mathbf{KL}(q_{\boldsymbol\omega}||p_{\boldsymbol\omega})$ vanishes in the lower bound \eqref{identity1}, \eqref{identity2} too.
\begin{align*}
(\Psi_1)_{nm}& = \mathbf{E}\left[\Phi_{nm}\right]_{q_{\boldsymbol{h}_n}}\\
& =  \mathbf{E}\left[\sqrt{\frac{2\upsigma_{\text{power}}^2}{M}}\cos(\hat{\mathbf{z}}_m^T(\mathbf{h}_n-\mathbf{u}_m)+\mathrm{b}_m)\right]_{q_{\boldsymbol{h}_n}}\\
& \stackrel{\scriptstyle\mathrm{B.2}}= \sqrt{\frac{2\upsigma_{\text{power}}^2}{M}} e^{-\frac{1}{2}\hat{\mathbf{z}}_m^T\boldsymbol\uplambda_n\hat{\mathbf{z}}_m}\cos(\hat{\mathbf{z}}_m^T(\boldsymbol\upmu_n-\mathbf{u}_m)+\mathrm{b}_m),
\end{align*}
for $m = 1,\dots, M$, $n, = 1,\dots, N$ with 
\begin{align*}
\hat{\mathbf{z}}_m = 2\pi(\mathfrak{L}^{-1}\mathbf{z}_m+\mathbf{p}),
\end{align*}
again, for the other variables, see the defined variables in the general case \ref{Generalcase}.\\
\\
\noindent$\Psi_2 = \sum\limits_{n=1}^N \Psi_2^n$ where
\begin{align*}
(\Psi_2^n)_{mm'}& = \mathbf{E}\left[\Phi_{mn}^T\Phi_{nm'}\right]_{q_{\boldsymbol{h}_n}}\\
&= \mathbf{E}\left[\frac{2\upsigma_{\text{power}}^2}{M}\cos(\hat{\mathbf{z}}_m^T(\mathbf{h}_n-\mathbf{u}_m)+\mathrm{b}_m)\cos(\hat{\mathbf{z}}_{m'}^T(\mathbf{h}_n-\mathbf{u}_{m'})+\mathrm{b}_{m'})\right]_{q_{\boldsymbol{h}_n}}\\
& \stackrel{\mathrm{AT2}}= \mathbf{E}\left[\frac{\upsigma_{\text{power}}^2}{M}\left(\cos(\bar{\mathbf{z}}_{mm'}^T\mathbf{h}_n - \bar{\uprho}_{mm'} + \bar{\mathrm{b}}_{mm'})\right.\right.\\
&\left.\left.+\cos(\overset{+}{\mathbf{z}}_{mm'}^T\mathbf{h}_n -\overset{+}{\uprho}_{mm'}+  \overset{+}{\mathrm{\mathrm{b}}}_{mm'})\right)\right]_{q_{\boldsymbol{h}_n}}\\
& = \frac{\upsigma_{\text{power}}^2}{M}\left(e^{-\frac{1}{2}\bar{\mathbf{z}}_{mm'}^T\boldsymbol\uplambda_n\bar{\mathbf{z}}_{mm'}}\cos(\bar{\mathbf{z}}_{mm'}^T\boldsymbol\upmu_n - \bar{\uprho}_{mm'} + \bar{\mathrm{b}}_{mm'})\right.\\
&\left.+e^{-\frac{1}{2}\overset{+}{\mathbf{z}}_{mm'}^T\boldsymbol\uplambda_n\overset{+}{\mathbf{z}}_{mm'}}\cos(\overset{+}{\mathbf{z}}_{mm'}^T\boldsymbol\upmu_n - \overset{+}{\uprho}_{mm'} + \overset{+}{\mathrm{b}}_{mm'}))\right),
\end{align*}
for $m,m' = 1,\dots, M$ with 
\begin{align*}
\bar{\uprho}_{mm'}&=\hat{\mathbf{z}}_m^T \mathbf{u}_m- \hat{\mathbf{z}}_{m'}^T \mathbf{u}_{m'},\\
\overset{+}{\uprho}_{mm'}&=\hat{\mathbf{z}}_m^T \mathbf{u}_m+ \hat{\mathbf{z}}_{m'}^T \mathbf{u}_{m'},\\
\bar{\mathbf{z}}_{mm'} &= \hat{\mathbf{z}}_m - \hat{\mathbf{z}}_{m'},\\
\overset{+}{\mathbf{z}}_{mm'} &= \hat{\mathbf{z}}_m + \hat{\mathbf{z}}_{m'},
\end{align*}
again, for the other variables, see the defined variables in the general case \ref{Generalcase}.
\newpage
\subsection{Variational bound}
\label{sec:Variationalbound}

{\bf Proposition 1.} {\it Let 
\begin{align*}
p_{\boldsymbol{Y}}(\mathbf{Y})&= \int\limits p_{\boldsymbol{Y},\boldsymbol{A},\boldsymbol\omega,\boldsymbol{H}}(\mathbf{Y},\mathbf{A},\text{\textbf{\textomega}},\mathbf{H})d\mathbf{A}d\text{\textbf{\textomega}}d\mathbf{H}\\
&= \int\limits p_{\boldsymbol{Y}|\boldsymbol{A},\boldsymbol\omega,\boldsymbol{H}}(\mathbf{Y}|\mathbf{A},\text{\textbf{\textomega}},\mathbf{H})p_{\boldsymbol{A}}(\mathbf{A})p_{\boldsymbol\omega}(\text{\textbf{\textomega}})p_{\boldsymbol{H}}(\mathbf{H})d\mathbf{A}d\text{\textbf{\textomega}}d\mathbf{H} 
\end{align*}
be the ML of our model with sparse approximation SM as in Section \ref{sec:SparseSpectrumGaussianProcess}, priors \eqref{prior1}, \eqref{prior2}, \eqref{prior3} variational distributions \eqref{var1}, \eqref{var2}, \eqref{identity11} and $P_{\text{\tiny{DGP}}}\stackrel{\mathrm{def}}= p_{\boldsymbol{A}}p_{\boldsymbol\omega}p_{\boldsymbol{H}}$, $Q_{\text{\tiny{DGP}}} \stackrel{\mathrm{def}}=q_{\boldsymbol{A}}q_{\boldsymbol\omega}q_{\boldsymbol{H}}$. Then with the VI procedure, $\mathcal{G}_{\text{\tiny{DGP}}} \stackrel{\mathrm{def}}= \log(p_{\boldsymbol{Y}|\boldsymbol{A},\boldsymbol\omega,\boldsymbol{H}}(\mathbf{Y}|\mathbf{A},\text{\textbf{\textomega}},\mathbf{H}))$, $\mathbf{M} = [\mathbf{m}_1, \dots  ,\mathbf{m}_D]\in\mathbb R^{M\times D}$ and $\mathbf{S} = \sum\limits_{i=1}^D \mathbf{s}_d$ we get the bound
\begin{align*} 
\hspace{2.4cm}\log(p_{\boldsymbol{Y}}(\mathbf{Y}))&\geq \mathbf{E}[\mathcal{G}_{\text{\tiny{DGP}}}]_{Q_{\text{\tiny{DGP}}}} - \mathbf{KL}(Q_{\text{\tiny{DGP}}}||P_{\text{\tiny{DGP}}}),\hspace{3cm}\eqref{identity1}\\
& = -\frac{ND}{2}\log(2\pi \upsigma_{\text{noise}}^2) - \frac{\text{\textup{tr}}\left(\mathbf{Y}^T\mathbf{Y}\right)}{2\upsigma_{\text{noise}}^2},\hspace{3cm}\eqref{identity2}\\
& + \frac{\text{\textup{tr}}\left(\mathbf{Y}^T\Psi_1 \mathbf{M}\right)}{\upsigma_{\text{noise}}^2}- \frac{\text{\textup{tr}}\left(\Psi_2(\mathbf{S} + \mathbf{M}\mathbf{M}^T)\right)}{2\upsigma_{\text{noise}}^2}\\
&- \mathbf{KL}(q_{\boldsymbol{A}}||p_{\boldsymbol{A}}) - \mathbf{KL}(q_{\boldsymbol\omega}||p_{\boldsymbol\omega})- \mathbf{KL}(q_{\boldsymbol{H}}||p_{\boldsymbol{H}})\\
&\stackrel{\mathrm{def}}=\mathbfcal{L}.
\end{align*}
}\\
\begin{proof}
We use the notation $\mathcal{N}_{\mathbf{y}^d}$ for a Gaussian density belonging to the random vector ${\boldsymbol{y}^d}$ and filled in with the data $\mathbf{y}^d$.
\begin{align*} 
\log(p_{\boldsymbol{Y}}(\mathbf{Y})) &= \log\left(\int\limits p_{\boldsymbol{Y}|\boldsymbol{A},\boldsymbol\omega,\boldsymbol{H}}(\mathbf{Y}|\mathbf{A},\boldsymbol\omega,\mathbf{H})p_{\boldsymbol{A}}(\mathbf{A})p_{\boldsymbol\omega}(\boldsymbol\omega)p(\mathbf{H})d\mathbf{A}d\text{\textbf{\textomega}} d\mathbf{H}\right)\\
&= \log\left(\int\limits p(\mathbf{Y}|\mathbf{A},\boldsymbol\omega,\mathbf{H})P_{\text{\tiny{DGP}}}(\mathbf{A},\text{\textbf{\textomega}},\mathbf{H}) d\text{\textbf{\textomega}} d\mathbf{H}\right)\\
& \stackrel{\mathrm{JI}}\geq \int Q_{\text{\tiny{DGP}}}(\mathbf{A},\text{\textbf{\textomega}},\mathbf{H})\log\left(p(\mathbf{Y}|\mathbf{A},\boldsymbol\omega,\mathbf{H})\right)d\mathbf{A} d\text{\textbf{\textomega}} d\mathbf{H}- \mathbf{KL}(Q_{\text{\tiny{DGP}}}||P_{\text{\tiny{DGP}}})\\
& = \mathbf{E}[\mathcal{G}{\text{\tiny{DGP}}}]_{Q_{\text{\tiny{DGP}}}} - \mathbf{KL}(Q_{\text{\tiny{DGP}}}||P_{\text{\tiny{DGP}}})\\
& = \int q_{\boldsymbol\omega}(\text{\textbf{\textomega}})q_{\boldsymbol{A}}(\mathbf{A})q_{\boldsymbol{H}}(\mathbf{H})\sum_{d=1}^D\log\left(p_{\boldsymbol{y}^d|\boldsymbol{a}_d,\boldsymbol\omega,\boldsymbol{H}}(\mathbf{y}^d|\mathbf{a}_d,\text{\textbf{\textomega}},\mathbf{H}))\right)d\mathbf{A}  d\text{\textbf{\textomega}} d\mathbf{H}\\
&- \mathbf{KL}(Q_{\text{\tiny{DGP}}}||P_{\text{\tiny{DGP}}})\\
& = \sum_{d=1}^D\left(\int q_{\boldsymbol{a}_d}(\mathbf{a}_d)q_{\boldsymbol\omega}(\text{\textbf{\textomega}})q_{\boldsymbol{H}}(\mathbf{H})\log\left(p_{\boldsymbol{y}^d|\boldsymbol{a}_d,\boldsymbol\omega,\boldsymbol{H}}(\mathbf{y}^d|\mathbf{a}_d,\text{\textbf{\textomega}},\mathbf{H})\right)d\mathbf{a}_d d\text{\textbf{\textomega}} d\mathbf{H}\right)\\
&- \mathbf{KL}(Q_{\text{\tiny{DGP}}}||P_{\text{\tiny{DGP}}})\\
& = \sum_{d=1}^D\left(\int q_{\boldsymbol{a}_d}(\mathbf{a}_d)q_{\boldsymbol\omega}(\text{\textbf{\textomega}})q_{\boldsymbol{H}}(\mathbf{H})\log(\mathcal{N}_{\mathbf{y}^d}(\Phi \mathbf{a}_d,\sigma_{\text{noise}}^2 I_N))d\mathbf{a}_d d\text{\textbf{\textomega}} d\mathbf{H}\right)\\
&-\mathbf{KL}(Q_{\text{\tiny{DGP}}}||P_{\text{\tiny{DGP}}})\\
& = \sum_{d=1}^D\left(\int q_{\boldsymbol\omega}(\text{\textbf{\textomega}})q_{\boldsymbol{a}_d}(\mathbf{a}_d)q_{\boldsymbol{H}}(\mathbf{H})\left(-\frac{N}{2}\log(2\pi) - \frac{\log(|\upsigma_{\text{noise}}^2 I_N|)}{2}\right.\right.\\
&\left.\left.- \frac{(\mathbf{y}^d)^T\mathbf{y}^d - 2(\mathbf{y}^d)^T\Phi \mathbf{a}_d + \mathbf{a}_d^T\Phi^T\Phi\mathbf{a}_d}{2\upsigma_{\text{noise}}^2}\right)d\mathbf{a}_d d\text{\textbf{\textomega}} d\mathbf{H}\right)- \mathbf{KL}(Q_{\text{\tiny{DGP}}}||P_{\text{\tiny{DGP}}})
\end{align*}
\begin{align*}
& = \sum_{d=1}^D\left(-\frac{N}{2}\log(2\pi \upsigma_{\text{noise}}^2) - \frac{(\mathbf{y}^d)^T\mathbf{y}^d}{2\upsigma_{\text{noise}}^2} + \frac{(\mathbf{y}^d)^T\mathbf{E}\left[\Phi \mathbf{a}_d\right]_{q_{\boldsymbol{a}_d}q_{\boldsymbol\omega}q_{\boldsymbol{H}}}}{\upsigma_{\text{noise}}^2} \right.\\
&\left.- \frac{\mathbf{E}\left[\mathbf{a}_d^T\Phi^T\Phi\mathbf{a}_d\right]_{q_{\boldsymbol{a}_d}q_{\boldsymbol\omega}q_{\boldsymbol{H}}}}{2\upsigma_{\text{noise}}^2}\right)- \mathbf{KL}(Q_{\text{\tiny{DGP}}}||P_{\text{\tiny{DGP}}})\\
& = \sum_{d=1}^D\left(-\frac{N}{2}\log(2\pi \upsigma_{\text{noise}}^2) - \frac{(\mathbf{y}^d)^T\mathbf{y}^d}{2\upsigma_{\text{noise}}^2} - \frac{(\mathbf{y}^d)^T\mathbf{E}\left[\Phi\right]_{q_{\boldsymbol\omega}q_{\boldsymbol{H}}}\mathbf{m}_d}{\upsigma_{\text{noise}}^2}\right.\\
& \left.+ \frac{\text{tr}\left(\mathbf{E}\left[\Phi^T\Phi\right]_{q_{\boldsymbol\omega}q_{\boldsymbol{H}}}(\mathbf{s}_d+\mathbf{m}_d \mathbf{m}_d^T)\right)}{2\upsigma_{\text{noise}}^2}\right)- \mathbf{KL}(Q_{\text{\tiny{DGP}}}||P_{\text{\tiny{DGP}}})\\
& = -\frac{ND}{2}\log(2\pi \upsigma_{\text{noise}}^2) - \frac{\text{tr}\left(\mathbf{Y}^T\mathbf{Y}\right)}{2\upsigma_{\text{noise}}^2} + \frac{\text{tr}\left(\mathbf{Y}^T\Psi_1 \mathbf{M}\right)}{\upsigma_{\text{noise}}^2} - \frac{\text{tr}\left(\Psi_2(\mathbf{S} + \mathbf{M}\mathbf{M}^T)\right)}{2\upsigma_{\text{noise}}^2}\\
& - \mathbf{KL}(q_{\boldsymbol{A}}||p_{\boldsymbol{A}}) - \mathbf{KL}(q_{\boldsymbol\omega}||p_{\boldsymbol\omega})- \mathbf{KL}(q_{\boldsymbol{H}}||p_{\boldsymbol{H}})
\end{align*}
\end{proof}
\qed
\subsection{Optimal variational distribution}
\label{sec:Optimalvariationaldistributionproof}
We show here the proof for the optimal variational distribution for DRGP-VSS. It also holds for DRGP-SS, because both bounds have the same structure.\\
\\
{\bf Proposition 2.} {\it
Let
\begin{align*}
\mathbfcal{L}(\boldsymbol\Theta) & = -\frac{ND}{2}\log(2\pi \upsigma_{\text{noise}}^2) - \frac{\text{\textup{tr}}\left(\mathbf{Y}^T\mathbf{Y}\right)}{2\upsigma_{\text{noise}}^2} + \frac{\text{\textup{tr}}\left(\mathbf{Y}^T\Psi_1 \mathbf{M}\right)}{\upsigma_{\text{noise}}^2} - \frac{\text{\textup{tr}}\left(\Psi_2(\mathbf{S} + \mathbf{M}\mathbf{M}^T)\right)}{2\upsigma_{\text{noise}}^2}\\
& - \mathbf{KL}(q_{\boldsymbol{A}}||p_{\boldsymbol{A}}) - \mathbf{KL}(q_{\boldsymbol\omega}||p_{\boldsymbol\omega})- \mathbf{KL}(q_{\boldsymbol{H}}||p_{\boldsymbol{H}}),
\end{align*}
be the cost function for our approximate model, then for $\mathbfcal{L}$ and for $\boldsymbol{A}$ an optimal distribution in the sense of a local functional optima $\mathbfcal{L}(q^{\text{opt}}_{\boldsymbol{A}})$, with priors \eqref{prior1}, \eqref{prior2}, \eqref{prior3}, variational distributions \eqref{var1}, \eqref{var2}, \eqref{identity11} is given by
\begin{align*}
q^{\text{opt}}_{\boldsymbol{A}}\stackrel{\mathrm{def}}= \prod_{d=1}^{D}q_{\boldsymbol{a}_d},\text{ where }\boldsymbol{a}_d \sim\mathcal{N}(A^{-1}\Psi_1^T\mathbf{y}_d,\upsigma_{\text{noise}}^2 A^{-1}),
\end{align*}
with $A = \Psi_2 + \upsigma_{\text{noise}}^2 I_M$, for $d=1,\dots,D$.
}\\
\begin{proof}
We show the proof for a specific $\boldsymbol{a}_d$, just writing $\boldsymbol{a}$, as $\mathbfcal{L}$ decomposes in a sum of $D$ independent terms. The general case for $\boldsymbol{A}$ follows immediately. Let
\begin{align*}
&\mathcal{L}(q_{\boldsymbol{a}})= \int\limits q_{\boldsymbol{a}}(\mathbf{a})\int\limits q_{\boldsymbol\omega}(\text{\textbf{\textomega}})\int\limits q_{\boldsymbol{H}}(\mathbf{H}) \log(p_{\boldsymbol{y}|\boldsymbol{a},\boldsymbol\omega,\boldsymbol{H}}(\mathbf{y}|\mathbf{a},\mathbf{H},\text{\textbf{\textomega}}))d\mathbf{a}d\text{\textbf{\textomega}} d\mathbf{H} \\
&\;- \int\limits q_{\boldsymbol{a}}(\mathbf{a})\log\left(\frac{q_{\boldsymbol{a}}(\mathbf{a})}{p_{\boldsymbol{a}}(\mathbf{a})}\right)d\mathbf{a} - \int\limits q_{\boldsymbol\omega}(\text{\textbf{\textomega}})\log\left(\frac{q_{\boldsymbol\omega}(\text{\textbf{\textomega}})}{p_{\boldsymbol\omega}(\text{\textbf{\textomega}})}\right)d\text{\textbf{\textomega}} - \int\limits q_{\boldsymbol{H}}(\mathbf{H})\log\left(\frac{q_{\boldsymbol{H}}(\mathbf{H})}{p_{\boldsymbol{H}}(\mathbf{H})}\right)d\mathbf{H}.
\end{align*}
The aim is to solve the following Lagrange multiplier condition
\begin{align*}
\frac{d(\mathcal{L}(q_{\boldsymbol{a}}) + \boldsymbol\lambda\left(\int\limits q_{\boldsymbol{a}}(\mathbf{a})d\mathbf{a} - 1\right))}{d q(\mathbf{a})} = 0
\end{align*}
for some $\boldsymbol\lambda\in\mathbb R$. With the standard functional derivative rules
\begin{align*}
F(\mathrm{f})=\int_{\mathbb{R}^Q}\mathrm{f}(\mathbf{x})\mathrm{g}(\mathbf{x})d\mathbf{x}\;\;\;\text{and}\;\;\;\frac{d F(\mathrm{f})}{d\mathrm{f}(\mathbf{x})}=\mathrm{g}(\mathbf{x})
\end{align*}
we get
\begin{align*}
\int\limits q_{\boldsymbol\omega}(\text{\textbf{\textomega}})\int\limits q_{\boldsymbol{H}}(\mathbf{H}) \log(p_{\boldsymbol{y}|\boldsymbol{a},\boldsymbol\omega,\boldsymbol{H}}(\mathbf{y}|\mathbf{a},\mathbf{H},\text{\textbf{\textomega}})d\text{\textbf{\textomega}} d\mathbf{H}- \log\left(\frac{q_{\boldsymbol{a}}(\mathbf{a})}{p_{\boldsymbol{a}}(\mathbf{a})}\right) - 1 + \boldsymbol\lambda = 0.
\end{align*}
So with $\boldsymbol\lambda = 1 - \log\left(\sqrt{|A^{-1}|}\right) + \frac{\mathbf{y}^T\left(I_M + \Psi_1 A^{-1}\Psi_1^T)\right)\mathbf{y}}{2\upsigma_{\text{\tiny noise}}^2}$ we get
\begin{align*}
q_{\boldsymbol{a}}(\mathbf{a})& = e^{\boldsymbol\lambda - 1}e^{\int\limits q_{\boldsymbol\omega}(\text{\textbf{\textomega}})\int\limits q_{\boldsymbol{H}}(\mathbf{H}) \log(p_{\boldsymbol{y}|\boldsymbol{a},\boldsymbol\omega,\boldsymbol{H}}(\mathbf{y}|\mathbf{a},\mathbf{H},\text{\textbf{\textomega}})d\text{\textbf{\textomega}} d\mathbf{H}} p_{\boldsymbol{a}}(\mathbf{a})\\
& = \frac{1}{\sqrt{|A^{-1}|}}e^{-\frac{1}{2}\mathbf{a}^T\frac{A}{\upsigma_{\text{{\tiny noise}}}^2}\mathbf{a} + \frac{\mathbf{y}^T\Psi_1 \mathbf{a}}{\upsigma_{\text{\tiny noise}}^2} - \frac{\mathbf{y}^T\Psi_1 A^{-1}\Psi_1^T\mathbf{y}}{2\upsigma_{\text{\tiny noise}}^2} - \frac{N}{2}\log(2\pi \upsigma_{\text{\tiny noise}}^2)}\\
& = \frac{1}{\sqrt{\left(2\pi \right)^N\left|\upsigma_{\text{noise}}^2 A^{-1}\right|}}e^{-\frac{1}{2}(\mathbf{a}-A^{-1}\Psi_1^T \mathbf{y})^T\frac{A}{\upsigma_{\text{\tiny noise}}^2}(\mathbf{a}-A^{-1}\Psi_1^T \mathbf{y})},
\end{align*}
and as $q_{\boldsymbol{a}}$ is constrained to be Gaussian, it must be the same as
\begin{align*}
\mathcal{N}(A^{-1}\Psi_1^T\mathbf{y},\upsigma_{\text{noise}}^2 A^{-1}),
\end{align*}
with $A = \Psi_2 + \upsigma_{\text{noise}}^2 I_M$.
\end{proof}
\qed
\subsection{Optimal lower bound}
\label{sec:Optimallowerbound}
We show here the exact derivation of the optimal lower bound for DRGP-VSS. It also holds for DRGP-SS, because both bounds follow the same structural property.\\
\\
{\bf Proposition 3.} {\it The cost function
\begin{align*}
\mathbfcal{L}(\boldsymbol\Theta) & = -\frac{ND}{2}\log(2\pi \upsigma_{\text{noise}}^2) - \frac{\text{\textup{tr}}\left(\mathbf{Y}^T\mathbf{Y}\right)}{2\upsigma_{\text{noise}}^2} + \frac{\text{\textup{tr}}\left(\mathbf{Y}^T\Psi_1 \mathbf{M}\right)}{\upsigma_{\text{noise}}^2} - \frac{\text{\textup{tr}}\left(\Psi_2(\mathbf{S} + \mathbf{M}\mathbf{M}^T)\right)}{2\upsigma_{\text{noise}}^2}\\
& - \mathbf{KL}(q_{\boldsymbol{A}}||p_{\boldsymbol{A}}) - \mathbf{KL}(q_{\boldsymbol\omega}||p_{\boldsymbol\omega})- \mathbf{KL}(q_{\boldsymbol{H}}||p_{\boldsymbol{H}}),
\end{align*}
reduces to the following optimal cost function, when using Proposition 2.,
\begin{align*}
\mathbfcal{L}^{\text{opt}}(\boldsymbol\Theta^{\text{opt}}) & =-\frac{(N-M)D}{2}\log(\upsigma_{\text{noise}}^2)- \frac{ND}{2}\log(2\pi) \\
&- \frac{\text{\textup{tr}}\left(\mathbf{Y}^T\mathbf{Y}\right)}{2\upsigma_{\text{noise}}^2} + \frac{\text{\textup{tr}}\left(\mathbf{Y}^T\Psi_1 A^{-1}\Psi_1^T\mathbf{Y}\right)}{2\upsigma_{\text{noise}}^2} -\frac{D\log(\left|A^{-1}\right|)}{2}\\
&- \mathbf{KL}(q_{\boldsymbol\omega}||p_{\boldsymbol\omega})- \mathbf{KL}(q_{\boldsymbol{H}}||p_{\boldsymbol{H}}),
\end{align*}
where $\boldsymbol\Theta^{\text{opt}}$ is the reduced set of parameters to optimize in the training.
}
\newpage
\begin{proof}
\begin{align*}
&\log(p(\mathbf{Y})) + \mathbf{KL}(q_{\boldsymbol\omega}||p_{\boldsymbol\omega})+\mathbf{KL}(q_{\boldsymbol{H}}||p_{\boldsymbol{H}}) \\
& \geq -\frac{ND}{2}\log(2\pi \upsigma_{\text{noise}}^2) - \frac{\text{tr}\left(\mathbf{Y}^T\mathbf{Y}\right)}{2\upsigma_{\text{noise}}^2} + \frac{\text{tr}\left(\mathbf{Y}^T\Psi_1 \mathbf{M}\right)}{\upsigma_{\text{noise}}^2} - \frac{\text{tr}\left(\Psi_2(\mathbf{S} + \mathbf{M}\mathbf{M}^T)\right)}{2\upsigma_{\text{noise}}^2}\\
&  - \mathbf{KL}(q_{\boldsymbol{A}}||p_{\boldsymbol{A}}) \\
& =-\frac{ND}{2}\log(2\pi \upsigma_{\text{noise}}^2) - \frac{\text{tr}\left(\mathbf{Y}^T\mathbf{Y}\right)}{2\upsigma_{\text{noise}}^2} + \frac{\text{tr}\left(\mathbf{Y}^T\Psi_1 A^{-1}\Psi_1^T\mathbf{Y}\right)}{\upsigma_{\text{noise}}^2}\\
& - \frac{\text{tr}\left(\Psi_2(D \upsigma_{\text{noise}}^2 A^{-1} + A^{-1}\Psi_1^T\mathbf{Y}\mathbf{Y}^T\Psi_1 A^{-1})\right)}{2\upsigma_{\text{noise}}^2} - \mathbf{KL}(q_{\boldsymbol{A}}||p_{\boldsymbol{A}}) \\
& \stackrel{\mathrm{\ref{KLA}}}= -\frac{ND}{2}\log(2\pi \upsigma_{\text{noise}}^2) - \frac{\text{tr}\left(\mathbf{Y}^T\mathbf{Y}\right)}{2\upsigma_{\text{noise}}^2} + \frac{\text{tr}\left(\mathbf{Y}^T\Psi_1 A^{-1}\Psi_1^T\mathbf{Y}\right)}{\upsigma_{\text{noise}}^2} - \frac{D\text{tr}\left(\upsigma_{\text{noise}}^2\Psi_2 A^{-1}\right)}{2\upsigma_{\text{noise}}^2}\\
&  - \frac{\text{tr}\left(\Psi_2A^{-1}\Psi_1^T\mathbf{Y}\mathbf{Y}^T\Psi_1 A^{-1}\right)}{2\upsigma_{\text{noise}}^2} - \frac{D\text{tr}\left(\upsigma_{\text{noise}}^4 A^{-1} \right)}{2\upsigma_{\text{noise}}^2}-\frac{\text{tr}\left(\mathbf{Y}^T\Psi_1A^{-1}A^{-1}\Psi_1^T \mathbf{Y}\right)}{2} \\
&+\frac{D\log(\left|\upsigma_{\text{noise}}^2A^{-1}\right|)}{2}+\frac{MD}{2}\\
& = -\frac{ND}{2}\log(2\pi \upsigma_{\text{noise}}^2) - \frac{\text{tr}\left(\mathbf{Y}^T\mathbf{Y}\right)}{2\upsigma_{\text{noise}}^2} + \frac{\text{tr}\left(\mathbf{Y}^T\Psi_1 A^{-1}\Psi_1^T\mathbf{Y}\right)}{\upsigma_{\text{noise}}^2}- \frac{D\text{tr}\left(\upsigma_{\text{noise}}^2 A^{-1} -\Psi_2 A^{-1}\right)}{2\upsigma_{\text{noise}}^2}\\
&  -\frac{D\text{tr}\left(\Psi_2A^{-1}\Psi_1^T\mathbf{Y}\mathbf{Y}^T\Psi_1 A^{-1}+\upsigma_{\text{noise}}^2A^{-1}\Psi_1^T\mathbf{Y}\mathbf{Y}^T\Psi_1 A^{-1}\right)}{2\upsigma_{\text{noise}}^2}+\frac{D\log(\left|\upsigma_{\text{noise}}^2A^{-1}\right|)}{2}+\frac{MD}{2}\\
& = -\frac{(N-M)D}{2}\log(\upsigma_{\text{noise}}^2) - \frac{ND}{2}\log(2\pi) - \frac{\text{tr}\left(\mathbf{Y}^T\mathbf{Y}\right)}{2\upsigma_{\text{noise}}^2} + \frac{\text{tr}\left(\mathbf{Y}^T\Psi_1 A^{-1}\Psi_1^T\mathbf{Y}\right)}{\upsigma_{\text{noise}}^2} - \frac{MD}{2}\\
& -\frac{\text{tr}\left((\Psi_2 + \upsigma_{\text{noise}}^2)(A^{-1}\Psi_1^T\mathbf{Y}\mathbf{Y}^T\Psi_1 A^{-1})\right)}{2\upsigma_{\text{noise}}^2}-\frac{\log(\left|A^{-1}\right|)}{2}+\frac{MD}{2}\\
& = -\frac{(N-M)D}{2}\log(\upsigma_{\text{noise}}^2) - \frac{ND}{2}\log(2\pi) - \frac{\text{tr}\left(\mathbf{Y}^T\mathbf{Y}\right)}{2\upsigma_{\text{noise}}^2} + \frac{\text{tr}\left(\mathbf{Y}^T\Psi_1 A^{-1}\Psi_1^T\mathbf{Y}\right)}{\upsigma_{\text{noise}}^2} \\
& -\frac{\text{tr}\left(A(A^{-1}\Psi_1^T\mathbf{Y}\mathbf{Y}^T\Psi_1 A^{-1}\right)}{2\upsigma_{\text{noise}}^2}-\frac{D\log(\left|A^{-1}\right|)}{2}\\
& = -\frac{(N-M)D}{2}\log(\upsigma_{\text{noise}}^2) - \frac{ND}{2}\log(2\pi) - \frac{\text{tr}\left(\mathbf{Y}^T\mathbf{Y}\right)}{2\upsigma_{\text{noise}}^2} + \frac{\text{tr}\left(\mathbf{Y}^T\Psi_1 A^{-1}\Psi_1^T\mathbf{Y}\right)}{\upsigma_{\text{noise}}^2}\\
& -\frac{\text{tr}\left(A(A^{-1}\Psi_1^T\mathbf{Y}\mathbf{Y}^T\Psi_1 A^{-1}\right)}{2\upsigma_{\text{noise}}^2}-\frac{D\log(\left|A^{-1}\right|)}{2} \\
& = -\frac{(N-M)D}{2}\log(\upsigma_{\text{noise}}^2) - \frac{ND}{2}\log(2\pi) - \frac{\text{tr}\left(\mathbf{Y}^T\mathbf{Y}\right)}{2\upsigma_{\text{noise}}^2} + \frac{\text{tr}\left(\mathbf{Y}^T\Psi_1 A^{-1}\Psi_1^T\mathbf{Y}\right)}{2\upsigma_{\text{noise}}^2}\\
& -\frac{D\log(\left|A^{-1}\right|)}{2}
\end{align*}
\end{proof}
\qed
\newpage
\subsection{REVARB bound for (V)SSGP}
\label{sec:REVARBboundforVSSGP}
We show here the exact derivation of the REVARB-VSS. DRGP-SS follows immediately because of structural correspondence. Therefore, we condition on $\boldsymbol{X}$ from the beginning on.\\
\\
\noindent{\bf Proposition 4.} {\it The REVARB-(V)SS bound $\mathbfcal{L}_{\text{\scalebox{.8}{\tiny{REVARB}}}}$ to optimize, with priors and variational distributions defined in \eqref{revarb}, and
\begin{align*}
\mathcal{G}_{\text{\scalebox{.8}{\tiny{REVARB}}}} &\stackrel{\mathrm{def}}= \log(p_{\boldsymbol{y}_{H_{\mathbf{x}}+1:}|\boldsymbol{a}^{(L+1)},\boldsymbol\omega^{(L+1)},\hat{\boldsymbol{H}}^{(L+1)}}(\mathbf{y}_{H_{\mathbf{x}}+1:}|\mathbf{a}^{(L+1)},\text{\textbf{\textomega}}^{(L+1)},\hat{\mathbf{H}}^{(L+1)}))\\
&+\sum_{l=1}^{L}\log(p_{\boldsymbol{h}^{(l)}_{H_{\mathbf{x}}+1:}|\boldsymbol{a}^{(l)},\boldsymbol\omega^{(l)},\hat{\boldsymbol{H}}^{(l)}}(\mathbf{h}^{(l)}_{H_{\mathbf{x}}+1:}|\mathbf{a}^{(l)},\text{\textbf{\textomega}}^{(l)},\hat{\mathbf{H}}^{(l)}))
\end{align*}
is
\begin{align*}
\log(p_{\boldsymbol{y}_{H_{\mathbf{x}}+1:}|\boldsymbol{X}}(\mathbf{y}_{H_{\mathbf{x}}+1:}|\mathbf{X})) \geq \mathbfcal{L}_{\text{\scalebox{.8}{\tiny{REVARB}}}}=\mathbf{E}[\mathcal{G}_{\text{\scalebox{.8}{\tiny{REVARB}}}}]_{Q_{\text{\scalebox{.8}{\tiny{REVARB}}}}} - \mathbf{KL}(Q_{\text{\scalebox{.8}{\tiny{REVARB}}}}||P_{\text{\scalebox{.8}{\tiny{REVARB}}}}).
\end{align*}
Additionally, the optimal bound $\mathbfcal{L}^{\text{opt}}_{\text{\scalebox{.8}{\tiny{REVARB}}}}$ follows immediately analogous to Proposition 3. and the fact, that the bound decomposes into a sum of independent terms for $\boldsymbol{A}_L$.
}\\
\begin{proof}
\begin{align*}
& \log(p_{\boldsymbol{y}_{H_{\mathbf{x}}+1:}|\boldsymbol{X}}(\mathbf{y}_{H_{\mathbf{x}}+1:}|\mathbf{X}))\\
& = \int p_{\boldsymbol{y}_{H_{\mathbf{x}}+1:},\boldsymbol{a}^{(L+1)},\boldsymbol\omega^{(L+1)},\left[\boldsymbol{a}^{(l)},\boldsymbol\omega^{(l)},\boldsymbol{h}^{(l)}\right]_{l=1}^L|\boldsymbol{X}}\\
&\left(\mathbf{y}_{H_{\mathbf{x}}+1:},\mathbf{a}^{(L+1)},\text{\textbf{\textomega}}^{(L+1)}\left[\mathbf{a}^{(l)},\text{\textbf{\textomega}}^{(l)},\mathbf{h}^{(l)}\right]_{l=1}^L|\mathbf{X}\right)d\mathbf{A}_L d\text{\textbf{\textomega}}_L d\mathbf{H}_L\\
& = \int \frac{Q_{\text{\scalebox{.8}{\tiny{REVARB}}}}(\mathbf{A}_L,\text{\textbf{\textomega}}_L,\mathbf{H}_L)}{Q_{\text{\scalebox{.8}{\tiny{REVARB}}}}(\mathbf{A}_L,\text{\textbf{\textomega}}_L,\mathbf{H}_L)}p_{\boldsymbol{y}_{H_{\mathbf{x}}+1:},\boldsymbol{a}^{(L+1)},\boldsymbol\omega^{(L+1)},\left[\boldsymbol{a}^{(l)},\boldsymbol\omega^{(l)},\boldsymbol{h}^{(l)}\right]_{l=1}^L|\boldsymbol{X}}\\
&\left(\mathbf{y}_{H_{\mathbf{x}}+1:},\mathbf{a}^{(L+1)},\text{\textbf{\textomega}}^{(L+1)}\left[\mathbf{a}^{(l)},\text{\textbf{\textomega}}^{(l)},\mathbf{h}^{(l)}\right]_{l=1}^L|\mathbf{X}\right)d\mathbf{A}_L d\text{\textbf{\textomega}}_L d\mathbf{H}_L\\
& \stackrel{\mathrm{JI}}\geq \int Q_{\text{\scalebox{.8}{\tiny{REVARB}}}}(\mathbf{A}_L,\text{\textbf{\textomega}}_L,\mathbf{H}_L)\log\left(\frac{1}{Q_{\text{\scalebox{.8}{\tiny{REVARB}}}}(\mathbf{A}_L,\text{\textbf{\textomega}}_L,\mathbf{H}_L)}\right.\\
&\left.p_{\boldsymbol{y}_{H_{\mathbf{x}}+1:},\boldsymbol{a}^{(L+1)},\boldsymbol\omega^{(L+1)},\left[\boldsymbol{a}^{(l)},\boldsymbol\omega^{(l)},\boldsymbol{h}^{(l)}\right]_{l=1}^L|\boldsymbol{X}}\right.\\
&\left.\left(\mathbf{y}_{H_{\mathbf{x}}+1:},\mathbf{a}^{(L+1)},\text{\textbf{\textomega}}^{(L+1)},\left[\mathbf{a}^{(l)},\text{\textbf{\textomega}}^{(l)},\mathbf{h}^{(l)}\right]_{l=1}^L|\mathbf{X}\right)\right)d\mathbf{A}_L d\text{\textbf{\textomega}}_L d\mathbf{H}_L\\
& = \int Q_{\text{\scalebox{.8}{\tiny{REVARB}}}}(\mathbf{A}_L,\text{\textbf{\textomega}}_L,\mathbf{H}_L)\log\left(\frac{1}{Q_{\text{\scalebox{.8}{\tiny{REVARB}}}}(\mathbf{A}_L,\text{\textbf{\textomega}}_L,\mathbf{H}_L)}p_{\boldsymbol{y}_{H_{\mathbf{x}}+1:}|\boldsymbol{a}^{(L+1)},\boldsymbol\omega^{(L+1)},\hat{\boldsymbol{H}}^{(L+1)}}\right.\\
&\left(\mathbf{y}_{H_{\mathbf{x}}+1:}|\mathbf{a}^{(L+1)},\text{\textbf{\textomega}}^{(L+1)},\hat{\mathbf{H}}^{(L+1)})\prod\limits_{l=1}^L p_{\boldsymbol{h}^{(l)}_{H_{\mathbf{x}}+1:}|\boldsymbol{a}^{(l)},\boldsymbol\omega^{(l)},\hat{\boldsymbol{H}}^{(l)}}\right.\\
&\left.(\mathbf{h}^{(l)}_{H_{\mathbf{x}}+1:}|\mathbf{a}^{(l)},\text{\textbf{\textomega}}^{(l)},\hat{\mathbf{H}}^{(l)})P_{\text{\scalebox{.8}{\tiny{REVARB}}}}(\mathbf{A}_L,\text{\textbf{\textomega}}_L,\mathbf{H}_L)\right)d\mathbf{A}_L d\text{\textbf{\textomega}}_L d\mathbf{H}_L\\
& = \mathbf{E}\left[\log\left(p_{\boldsymbol{y}_{H_{\mathbf{x}}+1:}|\boldsymbol{a}^{(L+1)},\boldsymbol\omega^{(L+1)},\hat{\boldsymbol{H}}^{(L+1)}}(\mathbf{y}_{H_{\mathbf{x}}+1:}|\mathbf{a}^{(L+1)},\text{\textbf{\textomega}}^{(L+1)},\hat{\mathbf{H}}^{(L+1)})\right.\right.\\
&\left.\left.\prod_{l=1}^{L}p_{\boldsymbol{h}^{(l)}_{H_{\mathbf{x}}+1:}|\boldsymbol{a}^{(l)},\boldsymbol\omega^{(l)},\hat{\boldsymbol{H}}^{(l)}}(\mathbf{h}^{(l)}_{H_{\mathbf{x}}+1:}|\mathbf{a}^{(l)},\text{\textbf{\textomega}}^{(l)},\hat{\mathbf{H}}^{(l)})\right)\right]_{Q_{\text{\scalebox{.8}{\tiny{REVARB}}}}}- \mathbf{KL}(Q_{\text{\scalebox{.8}{\tiny{REVARB}}}}||P_{\text{\scalebox{.8}{\tiny{REVARB}}}})\\
& = \mathbf{E}\left[\log(p_{\boldsymbol{y}_{H_{\mathbf{x}}+1:}|\boldsymbol{a}^{(L+1)},\boldsymbol\omega^{(L+1)},\hat{\boldsymbol{H}}^{(L+1)}}(\mathbf{y}_{H_{\mathbf{x}}+1:}|\mathbf{a}^{(L+1)},\text{\textbf{\textomega}}^{(L+1)},\hat{\mathbf{H}}^{(L+1)}))\right.\\
&\left.+\sum_{l=1}^{L}\log(p_{\boldsymbol{h}^{(l)}_{H_{\mathbf{x}}+1:}|\boldsymbol{a}^{(l)},\boldsymbol\omega^{(l)},\hat{\boldsymbol{H}}^{(l)}}(\mathbf{h}^{(l)}_{H_{\mathbf{x}}+1:}|\mathbf{a}^{(l)},\text{\textbf{\textomega}}^{(l)},\hat{\mathbf{H}}^{(l)}))\right]_{Q_{\text{\scalebox{.8}{\tiny{REVARB}}}}}- \mathbf{KL}(Q_{\text{\scalebox{.8}{\tiny{REVARB}}}}||P_{\text{\scalebox{.8}{\tiny{REVARB}}}})\\
&=\mathbf{E}[\mathcal{G}_{\text{\scalebox{.8}{\tiny{REVARB}}}}]_{Q_{\text{\scalebox{.8}{\tiny{REVARB}}}}} - \mathbf{KL}(Q_{\text{\scalebox{.8}{\tiny{REVARB}}}}||P_{\text{\scalebox{.8}{\tiny{REVARB}}}})
\end{align*}
\begin{align*}
& =\int q_{\boldsymbol{a}^{(L+1)}}(\mathbf{a}^{(L+1)})q_{\boldsymbol\omega^{(L+1)}}(\text{\textbf{\textomega}}^{(L+1)})q_{\boldsymbol{h}^{(L+1)}}(\mathbf{h}^{(L+1)})\\ 
&p_{\boldsymbol{y}_{H_{\mathbf{x}}+1:}|\boldsymbol{a}^{(L+1)},\boldsymbol\omega^{(L+1)},\hat{\boldsymbol{H}}^{(L+1)}}(\mathbf{y}_{H_{\mathbf{x}}+1:}|\mathbf{a}^{(L+1)},\text{\textbf{\textomega}}^{(L+1)},\hat{\mathbf{H}}^{(L+1)})\\
&d\mathbf{a}^{(L+1)}d\text{\textbf{\textomega}}^{(L+1)}d\mathbf{h}^{(L+1)}\\
& +\sum_{l=1}^{L}\int q_{\boldsymbol{a}^{(l)}}(\mathbf{a}^{(l)})q_{\boldsymbol\omega^{(l)}}(\text{\textbf{\textomega}}^{(l)})q_{\boldsymbol{h}^{(l)}}(\mathbf{h}^{(l)})\\
&p_{\boldsymbol{h}^{(l)}_{H_{\mathbf{x}}+1:}|\boldsymbol{a}^{(l)},\boldsymbol\omega^{(l)},\hat{\boldsymbol{H}}^{(l)}}(\mathbf{h}^{(l)}_{H_{\mathbf{x}}+1:}|\mathbf{a}^{(l)},\text{\textbf{\textomega}}^{(l)},\hat{\mathbf{H}}^{(l)})d\mathbf{a}^{(l)}d\text{\textbf{\textomega}}^{(l)}d\mathbf{h}^{(l)}\\
&-\sum_{l=1}^{L}\left(\sum_{i=1+H_{\mathbf{x}}-H_{\mathrm{h}}}^N\int_{\mathrm{h}_i^{(l)}}q_{h_i^{(l)}}\left(\mathrm{h}_i^{(l)}\right)\log\left(q_{h_i^{(l)}}\left(\mathrm{h}_i^{(l)}\right)\right)d \mathrm{h}_i^{(l)}\right.\\
&\left.\sum_{i=1+H_{\mathbf{x}}-H_{\mathrm{h}}}^{H_{\mathrm{h}}}\int_{\mathrm{h}_i^{(l)}}q_{h_i^{(l)}}\left(\mathrm{h}_i^{(l)}\right)\log\left(p_{h_i^{(l)}}\left(\mathrm{h}_i^{(l)}\right)\right)d \mathrm{h}_i^{(l)}\right)\\
&- \mathbf{KL}(q_{\boldsymbol{A}_L}||p_{\boldsymbol{A}_L}) - \mathbf{KL}(q_{\boldsymbol\omega_L}||p_{\boldsymbol\omega_L}),\numberthis\label{proof1}
\end{align*}
Using the derived lower bound expression in \eqref{identity2} for the 1-dimensional output case and\linebreak $\hat{N}=N-H_{\mathbf{x}}$ we come to
\begin{align*}
\eqref{proof1} & = -\frac{\hat{N}}{2}\sum_{l=1}^{L+1}\log(2\pi(\upsigma_{\text{noise}}^{(l)})^2)-\frac{\mathbf{y}_{H_{\mathbf{x}}+1:}^T\mathbf{y}_{H_{\mathbf{x}}+1:}}{2\left(\upsigma_{\text{noise}}^{\text{(L+1)}}\right)^2}\\
& + \frac{\mathbf{y}^T_{H_{\mathbf{x}}+1:}\Psi_1^{\text{(L+1)}} \mathbf{m}^{(L+1)}}{\left(\upsigma_{\text{noise}}^{\text{(L+1)}}\right)^2} - \frac{\text{tr}\left(\Psi_2^{\text{(L+1)}}(\mathbf{S} + \mathbf{m}^{(L+1)}(\mathbf{m}^{(L+1)})^T)\right)}{2\left(\upsigma_{\text{noise}}^{\text{(L+1)}}\right)^2}\\
& +\sum_{l=1}^{L}\left(-\frac{1}{2\left(\upsigma_{\text{noise}}^{(l)}\right)^2}\left(\left(\sum_{i=L+1}^N\uplambda_i^{(l)}\right)+ (\boldsymbol\upmu^{(l)})^T_{H_{\mathbf{x}}+1:}\boldsymbol\upmu^{(l)}_{H_{\mathbf{x}}+1:}\right)\right.\\
& \left. + \frac{\left((\boldsymbol\upmu^{(l)}_{H_{\mathbf{x}}+1:})^T\Psi_1^{(l)} \mathbf{m}^{(l)}\right)}{\left(\upsigma_{\text{noise}}^{(l)}\right)^2} - \frac{\text{tr}\left(\Psi_2^{(l)}(\mathbf{S} + \mathbf{m}^{(l)}(\mathbf{m}^{(l)})^T)\right)}{2\left(\upsigma_{\text{noise}}^{(l)}\right)^2}\right.\\
&\left.-\sum_{i=1+H_{\mathbf{x}}-H_{\mathrm{h}}}^N \int_{\mathrm{h}_i^{(l)}}q_{h_i^{(l)}}\left(\mathrm{h}_i^{(l)}\right)\log\left(q_{h_i^{(l)}}\left(\mathrm{h}_i^{(l)}\right)\right)d \mathrm{h}_i^{(l)}\right.\\
& \left.+\sum_{i=1+H_{\mathbf{x}}-H_{\mathrm{h}}}^{H_{\mathrm{h}}}\int_{\mathrm{h}_i^{(l)}}q_{h_i^{(l)}}\left(\mathrm{h}_i^{(l)}\right)\log\left(p_{h_i^{(l)}}\left(\mathrm{h}_i^{(l)}\right)\right)d \mathrm{h}_i^{(l)}\right)\\
& - \mathbf{KL}(q_{\boldsymbol{A}_L}||p_{\boldsymbol{A}_L}) - \mathbf{KL}(q_{\boldsymbol\omega_L}||p_{\boldsymbol\omega_L}). \numberthis\label{proof2} 
\end{align*}
For DRGP-SS we get the same as \eqref{proof2} but without $\mathbf{KL}(q_{\boldsymbol\omega_L}||p_{\boldsymbol\omega_L})$.\\
\newpage
Using \ref{sec:Optimallowerbound}, the optimal distribution for $\boldsymbol{A}_L$, we obtain
\begin{align*}
\eqref{proof2} & = -\frac{\hat{N}-M}{2}\sum_{l=1}^{L+1}\log\left(\left(\upsigma_{\text{noise}}^{(l)}\right)^2\right)-\frac{\mathbf{y}_{H_{\mathbf{x}}+1:}^T\mathbf{y}_{H_{\mathbf{x}}+1:}}{2\left(\upsigma_{\text{noise}}^{\text{(L+1)}}\right)^2}\\
& + \frac{\mathbf{y}_{H_{\mathbf{x}}+1:}^T\Psi_{1}^{\text{(L+1)}} (A^{(L+1)})^{-1}(\Psi_{1}^{\text{(L+1)}})^T\mathbf{y}_{H_{\mathbf{x}}+1:}}{2\left(\upsigma_{\text{noise}}^{\text{(L+1)}}\right)^2} -\frac{\log(|(A^{(L+1)})^{-1}|)}{2}\\
&+\sum_{l=1}^{L}\left(-\frac{1}{2\left(\upsigma_{\text{noise}}^{(l)}\right)^2}\left(\left(\sum_{i=L+1}^N\uplambda_i^{(l)}\right) + (\boldsymbol\upmu^{(l)}_{H_{\mathbf{x}}+1:})^T\boldsymbol\upmu_{H_{\mathbf{x}}+1:}^{(l)}\right)\right.\\
&\left.+ \frac{(\boldsymbol\upmu_{H_{\mathbf{x}}+1:}^{(l)})^T\Psi_{1}^{(l)} (A^{(l)})^{-1}(\Psi_{1}^{(l)})^T\boldsymbol\upmu_{H_{\mathbf{x}}+1:}^{(l)}}{2\left(\upsigma_{\text{noise}}^{(l)}\right)^2} -\frac{\log(|(A^{(l)})^{-1}|)}{2}\right.\\
&\left.-\sum_{i=1+H_{\mathbf{x}}-H_{\mathrm{h}}}^N \int_{\mathrm{h}_i^{(l)}}q_{h_i^{(l)}}\left(\mathrm{h}_i^{(l)}\right)\log\left(q_{h_i^{(l)}}\left(\mathrm{h}_i^{(l)}\right)\right)d \mathrm{h}_i^{(l)}\right.\\
& \left.+\sum_{i=1+H_{\mathbf{x}}-H_{\mathrm{h}}}^{H_{\mathrm{h}}}\int_{\mathrm{h}_i^{(l)}}q_{h_i^{(l)}}\left(\mathrm{h}_i^{(l)}\right)\log\left(p_{h_i^{(l)}}\left(\mathrm{h}_i^{(l)}\right)\right)d \mathrm{h}_i^{(l)}\right)\\
& -\frac{(L+1)\hat{N}}{2}\log(2\pi)- \mathbf{KL}(q_{\boldsymbol\omega_L}||p_{\boldsymbol\omega_L}).\numberthis\label{proof3} 
\end{align*}
For DRGP-SS we get the same as \eqref{proof3} but without $\mathbf{KL}(q_{\boldsymbol\omega_L}||p_{\boldsymbol\omega_L})$.\\
Furthermore, we have
\begin{align*}
\int\limits_{\mathrm{h}_i^{(l)}} q_{h_i^{(l)}}\left(\mathrm{h}_i^{(l)}\right)\log\left(q_{h_i^{(l)}}\left(\mathrm{h}_i^{(l)}\right)\right)d \mathrm{h}_i^{(l)} & = - \frac{1}{2}\log(2\pi\uplambda_i^{(l)}) - \frac{1}{2},\\
\int\limits_{\mathrm{h}_i^{(l)}} q_{h_i^{(l)}}\left(\mathrm{h}_i^{(l)}\right)\log\left(p_{h_i^{(l)}}\left(\mathrm{h}_i^{(l)}\right)\right)d \mathrm{h}_i^{(l)} & = - \frac{1}{2}\log(2\pi) - \frac{1}{2} \left(\uplambda_i^{(l)} + \left(\upmu_i^{(l)}\right)^2 \right).
\end{align*}
The statistics are $\Psi_1^{(1)} = \mathbf{E}\left[\Phi^{(1)}\right]_{q_{\boldsymbol\omega^{(1)}}q_{\boldsymbol{h}^{(1)}}}$ and $\Psi_2^{(1)} = \mathbf{E}\left[(\Phi^{(1)})^T\Phi^{(1)}\right]_{q_{\boldsymbol\omega^{(1)}}q_{\boldsymbol{h}^{(1)}}}$,\linebreak for $l = 1$, and $\Psi_1^{(l)} = \mathbf{E}\left[\Phi^{(l)}\right]_{q_{\boldsymbol\omega^{(l)}}q_{\boldsymbol{h}^{(l)}}q_{\boldsymbol{h}^{(l-1)}}}$ and $\Psi_2^{(l)} = \mathbf{E}\left[(\Phi^{(l)})^T\Phi^{(l)}\right]_{q_{\boldsymbol\omega^{(l)}}q_{\boldsymbol{h}^{(l)}}q_{\boldsymbol{h}^{(l-1)}}}$, for $l = 2,\dots,L$, and $\Psi_1^{(L+1)} = \mathbf{E}\left[\Phi^{(l)}\right]_{q_{\boldsymbol\omega^{(L+1)}}q_{\boldsymbol{h}^{(L)}}}$ and $\Psi_2^{(L+1)} = \mathbf{E}\left[(\Phi^{(l)})^T\Phi^{(l)}\right]_{q_{\boldsymbol\omega^{(L+1)}}q_{\boldsymbol{h}^{(L)}}}$, for $l = L+1$, where we have
\begin{align*}
\Phi^{(l)} \stackrel{\mathrm{def}}= \left[\phi(\hat{\mathbf{h}}_1^{(l)},\text{\textbf{\textomega}}^{(l)})\dots\phi(\hat{\mathbf{h}}_{\hat{N}}^{(l)},\text{\textbf{\textomega}}^{(l)})\right]^T\in\mathbb R^{\hat{N}\times M},
\end{align*}
and 
\begin{align*}
\phi(\hat{\mathbf{h}}^{(l)},\text{\textbf{\textomega}}^{(l)}) &\stackrel{\mathrm{def}}= \sqrt{\frac{2(\upsigma_{\text{power}}^{(l)})^2}{M}}\left[\cos(2\pi((\mathfrak{L}^{(l)})^{-1}(\mathbf{z}_1^{(l)})+(\mathbf{p}^{(l)}))^T(\hat{\mathbf{h}}^{(l)}-\mathbf{u}_1^{(l)}) + \mathrm{b}_1^{(l)}),\dots,\right.\\
&\left. \cos(2\pi((\mathfrak{L}^{(l)})^{-1}(\mathbf{z}_M^{(l)})+(\mathbf{p}^{(l)}))^T(\hat{\mathbf{h}}^{(l)}-\mathbf{u}_M^{(l)}) + \mathrm{b}_M^{(l)})\right]\in\mathbb R^{1\times M}
\end{align*}
where
\begin{align*}
\text{\textbf{\textomega}}^{(l)} \stackrel{\mathrm{def}}= [(\text{\textbf{\textomega}}_1^{(l)}),\dots,(\text{\textbf{\textomega}}_M^{(l)})]^T\stackrel{\mathrm{def}}= \begin{bmatrix}\begin{bmatrix}\mathbf{z}_1^{(l)}\\\mathrm{b}_1^{(l)}\end{bmatrix},\dots,\begin{bmatrix}\mathbf{z}_M^{(l)}\\\mathrm{b}_M^{(l)}\end{bmatrix}\end{bmatrix}^T\in\mathbb R^{M\times Q+1},
\end{align*}
with $\mathbf{u}_m^{(l)}\in\mathbb R^{Q}$, $m = 1,\dots,M$, pseudo-inputs and $\upsigma_{\text{power}}^{(l)}\in\mathbb R$ the amplitudes, scaling matrices $\mathfrak{L}^{(l)} = \text{diag}([2\pi \mathrm{l}_q^{(l)}]_{q=1}^{Q_l})$ and scaling vectors $\mathbf{p}^{(l)} = [(\mathrm{p}_1^{(l)})^{-1},\dots,(\mathrm{p}_Q^{(l)})^{-1}]^T$,\linebreak $A^{(l)}=\Psi_2^{(l)} + \left(\upsigma_{\text{noise}}^{(l)}\right)^2 I_M$, for $l = 1,\dots,L+1$.\qed
\end{proof}

\subsection{Mean and variance of the predictive distribution}
\label{sec:Meanandvarianceofthepredictivedistribution}
We show here the form of the mean of the predictive distribution of $f^{(l)}(\hat{\mathbf{h}}_{\ast}^{(l)})$.\\
Denoting $\mathbf{m}^{(l)}\in\mathbb R^M$ the mean and $\mathbf{s}^{(l)}\in\mathbb R^{M\times M}$ the variance following
\begin{align*}
\mathcal{N}(\mathbf{m}^{(l)},\mathbf{s}^{(l)})=\mathcal{N}((A^{(l)})^{-1}(\Psi_1^{(l)})^T\boldsymbol\upmu^{(l)}_{H_{\mathbf{x}}+1:},\upsigma_{\text{noise}}^2 (A^{(l)})^{-1}),
\end{align*}
for $1,\dots,L$, and for $l=L+1$ replacing $\boldsymbol\upmu^{(l)}_{H_{\mathbf{x}}+1:}$ with $\mathbf{y}_{H_{\mathbf{x}}+1:}$. With the approximate predictive distributions in \eqref{pred1}, \eqref{pred2}, \eqref{pred3} and the abbreviation $q_{f^{(l)}(\hat{\mathbf{h}}_{\ast}^{(l)})}$ for all $l=1,\dots,L+1$ in the sequel, we can derive for the first prediction the following.\\
\\
\noindent{\bf Proposition 5.} {\it Predictions for each layer $l$ and new $\hat{\mathbf{h}}_{*}^{(l)}$ are performed with
\begin{align*}
\mathbf{E}\left[f^{(l)}_{\hat{\mathbf{h}}_{\ast}^{(l)}}\right]_{q_{f^{(l)}_{\hat{\mathbf{h}}_{\ast}^{(l)}}}}  &= \Psi_{1\ast}^{(l)}\mathbf{m}^{(l)},\\
\mathbf{V}\left[f^{(l)}_{\hat{\mathbf{h}}_{\ast}^{(l)}}\right]_{q_{f^{(l)}_{\hat{\mathbf{h}}_{\ast}^{(l)}}}} & = \left(\textbf{m}^{(l)}\right)^T\left(\Psi_{2\ast}^{(l)}-\left(\Psi_{1\ast}^{(l)}\right)^T\Psi_{1\ast}^{(l)}\right)\textbf{m}^{(l)}+ \text{\textup{tr}}\left(\mathbf{s}^{(l)}\Psi_{2\ast}^{(l)}\right),
\end{align*}
where
\begin{align*}
\mathbf{m}^{(l)} \stackrel{\mathrm{opt}}= \left(A^{(l)}\right)^{-1}\left(\Psi_1^{(l)}\right)^T\boldsymbol\upmu^{(l)}_{H_{\mathbf{x}}+1:},\quad \mathbf{s}^{(l)}\stackrel{\mathrm{opt}}= \upsigma_{\text{noise}}^{(l)}\left(A^{(l)}\right)^{-1},
\end{align*}
for $1,\dots,L$, and fully analog for $l=L+1$ by replacing $\boldsymbol\upmu^{(l)}_{H_{\mathbf{x}}+1:}$ with $\mathbf{y}_{H_{\mathbf{x}}+1:}$.}\\
\begin{proof}
With 
\begin{align*}
\mathbf{E}\left[f^{(l)}_{\hat{\mathbf{h}}_{\ast}^{(l)}}\right]_{p_{f^{(l)}_{\hat{\mathbf{h}}_{\ast}^{(l)}}|\boldsymbol{a}^{(l)}, \boldsymbol\omega^{(l)}, \hat{\boldsymbol{h}}_{\ast}^{(l)}}}=\mathbf{E}\left[h^{(l)}_{\ast}\right]_{p_{h^{(l)}_{\ast}|\boldsymbol{a}^{(l)}, \boldsymbol\omega^{(l)}, \hat{\boldsymbol{h}}_{\ast}^{(l)}}}=\Phi_{\ast}^{(l)} \mathbf{a}^{(l)},
\end{align*}
and
\begin{align*}
\mathbf{V}\left[f^{(l)}_{\hat{\mathbf{h}}_{\ast}^{(l)}}\right]_{p_{f^{(l)}_{\hat{\mathbf{h}}_{\ast}^{(l)}}|\boldsymbol{a}^{(l)}, \boldsymbol\omega^{(l)}, \hat{\boldsymbol{h}}_{\ast}^{(l)}}}=\mathbf{V}\left[h^{(l)}_{\ast}\right]_{p_{h^{(l)}_{\ast}|\boldsymbol{a}^{(l)}, \boldsymbol\omega^{(l)}, \hat{\boldsymbol{h}}_{\ast}^{(l)}}}- \upsigma_{\text{noise}}^{(l)}=0,
\end{align*}
for $l=2\dots,L$ and for $l = L+1$ replacing $h^{(l)}_{\ast}$ with $y_{\ast}$, we derive the following.\\
We choose the abbreviation $Q_{*}=q_{\boldsymbol{h}^{(l)}}q_{\boldsymbol{h}^{(l-1)}}q_{\boldsymbol{h}_{*}^{(l)}}q_{\boldsymbol{h}_{*}^{(l-1)}}$.
\begin{align*}
\mathbf{E}\left[f^{(l)}_{\hat{\mathbf{h}}_{\ast}^{(l)}}\right]_{q_{f^{(l)}_{\hat{\mathbf{h}}_{\ast}^{(l)}}}}&=\mathbf{E}\left[\mathbf{E}\left[h^{(l)}_{\ast}\right]_{p_{h^{(l)}_{\ast}|\boldsymbol{a}^{(l)}, \boldsymbol\omega^{(l)}, \hat{\boldsymbol{h}}_{\ast}^{(l)}}}\right]_{q_{\boldsymbol{a}^{(l)}}q_{\boldsymbol\omega^{(l)}}Q_{*}}\\
& = \int \left(\Phi_{\ast}^{(l)} \mathbf{a}^{(l)}\right)q_{\boldsymbol{a}^{(l)}}(\mathbf{a}^{(l)})q_{\boldsymbol\omega^{(l)}}\text{\textbf{\textomega}}^{(l)})\\
&Q_{*}(\mathbf{h}^{(l)},\mathbf{h}^{(l-1)},\mathbf{h}_{*}^{(l)},\mathbf{h}_{*}^{(l-1)})d\mathbf{a}^{(l)}d\text{\textbf{\textomega}}^{(l)}d\mathbf{h}^{(l)}d\mathbf{h}^{(l-1)}d\mathbf{h}_{*}^{(l)}d\mathbf{h}_{*}^{(l-1)}\\
& = \int \Phi_{\ast}^{(l)}q_{\boldsymbol\omega^{(l)}}(\text{\textbf{\textomega}}^{(l)})Q_{*}(\mathbf{h}^{(l)},\mathbf{h}^{(l-1)},\mathbf{h}_{*}^{(l)},\mathbf{h}_{*}^{(l-1)})\\
&d\text{\textbf{\textomega}}^{(l)}d\mathbf{h}^{(l)}d\mathbf{h}^{(l-1)}\mathbf{h}_{*}^{(l)}d\mathbf{h}_{*}^{(l-1)}\int \mathbf{a}^{(l)}q_{\boldsymbol{a}^{(l)}}(\mathbf{a}^{(l)})d\mathbf{a}^{(l)}\\
& = \mathbf{E}\left[\Phi_{\ast}^{(l)}\right]_{q_{\boldsymbol\omega^{(l)}}Q_{*}}\mathbf{m}^{(l)}\\
& = \Psi_{1\ast}^{(l)}(A^{(l)})^{-1}\left(\Psi_1^{(l)}\right)^T\boldsymbol\upmu^{(l)}_{H_{\mathbf{x}}+1:},
\end{align*}
for $l=2\dots,L$. For $l=1$ and $l = L+1$ we cancel $q_{\boldsymbol{h}^{(l-1)}}(\mathbf{h}^{(l-1)})$ from the beginning and for $l = L+1$ we additionally replacing $\boldsymbol\upmu^{(l)}_{H_{\mathbf{x}}+1:}$ with $\mathbf{y}_{H_{\mathbf{x}}+1:}$.
\newpage

\begin{align*}
\mathbf{V}\left[f^{(l)}_{\hat{\mathbf{h}}_{\ast}^{(l)}}\right]_{q_{f^{(l)}_{\hat{\mathbf{h}}_{\ast}^{(l)}}}}& = \mathbf{E}\left[f^{(l)}_{\hat{\mathbf{h}}_{\ast}^{(l)}}f^{(l)}_{\hat{\mathbf{h}}_{\ast}^{(l)}}\right]_{q_{f^{(l)}_{\hat{\mathbf{h}}_{\ast}^{(l)}}}} - \mathbf{E}\left[f^{(l)}_{\hat{\mathbf{h}}_{\ast}^{(l)}}\right]_{q_{f^{(l)}_{\hat{\mathbf{h}}_{\ast}^{(l)}}}}^T\mathbf{E}\left[f^{(l)}_{\hat{\mathbf{h}}_{\ast}^{(l)}}\right]_{q_{f^{(l)}_{\hat{\mathbf{h}}_{\ast}^{(l)}}}}\\
& = \int \left(\mathbf{V}\left[f^{(l)}_{\hat{\mathbf{h}}_{\ast}^{(l)}}\right]_{p_{f^{(l)}_{\hat{\mathbf{h}}_{\ast}^{(l)}}|\boldsymbol{a}^{(l)}, \boldsymbol\omega^{(l)}, \hat{\boldsymbol{h}}_{\ast}^{(l)}}}\right.\\
&\left. +  \mathbf{E}\left[f^{(l)}_{\hat{\mathbf{h}}_{\ast}^{(l)}}\right]_{p_{f^{(l)}_{\hat{\mathbf{h}}_{\ast}^{(l)}}|\boldsymbol{a}^{(l)}, \boldsymbol\omega^{(l)}, \hat{\boldsymbol{h}}_{\ast}^{(l)}}}^T\mathbf{E}\left[f^{(l)}_{\hat{\mathbf{h}}_{\ast}^{(l)}}\right]_{p_{f^{(l)}_{\hat{\mathbf{h}}_{\ast}^{(l)}}|\boldsymbol{a}^{(l)}, \boldsymbol\omega^{(l)}, \hat{\boldsymbol{h}}_{\ast}^{(l)}}}\right)\\
& q_{\boldsymbol{a}^{(l)}}(\mathbf{a}^{(l)})q_{\boldsymbol\omega^{(l)}}\text{\textbf{\textomega}}^{(l)})Q_{*}(\mathbf{h}^{(l)},\mathbf{h}^{(l-1)},\mathbf{h}_{*}^{(l)},\mathbf{h}_{*}^{(l-1)})d\mathbf{a}^{(l)}d\text{\textbf{\textomega}}^{(l)}\\
&d\mathbf{h}^{(l)}d\mathbf{h}^{(l-1)}d\mathbf{h}_{*}^{(l)}d\mathbf{h}_{*}^{(l-1)} - \left(\mathbf{m}^{(l)}\right)^T\mathbf{E}\left[\Phi_{\ast}^{(l)}\right]_{q_{\boldsymbol\omega^{(l)}}Q_{*}}^T\mathbf{E}\left[\Phi_{\ast}^{(l)}\right]_{q_{\boldsymbol\omega^{(l)}}Q_{*}}\mathbf{m}^{(l)}\\
& = \int \left(\mathbf{V}\left[h^{(l)}_{\ast}\right]_{p_{h^{(l)}_{\ast}|\boldsymbol{a}^{(l)}, \boldsymbol\omega^{(l)}, \hat{\boldsymbol{h}}_{\ast}^{(l)}}}- \upsigma_{\text{noise}}^{(l)}  \right.\\
&\left. +  \mathbf{E}\left[h^{(l)}_{\ast}\right]_{p_{h^{(l)}_{\ast}|\boldsymbol{a}^{(l)}, \boldsymbol\omega^{(l)}, \hat{\boldsymbol{h}}_{\ast}^{(l)}}}^T\mathbf{E}\left[h^{(l)}_{\ast}\right]_{p_{h^{(l)}_{\ast}|\boldsymbol{a}^{(l)}, \boldsymbol\omega^{(l)}, \hat{\boldsymbol{h}}_{\ast}^{(l)}}}\right)\\
& q_{\boldsymbol{a}^{(l)}}(\mathbf{a}^{(l)})q_{\boldsymbol\omega^{(l)}}\text{\textbf{\textomega}}^{(l)})Q_{*}(\mathbf{h}^{(l)},\mathbf{h}^{(l-1)},\mathbf{h}_{*}^{(l)},\mathbf{h}_{*}^{(l-1)})d\mathbf{a}^{(l)}d\text{\textbf{\textomega}}^{(l)}\\
& d\mathbf{h}^{(l)}d\mathbf{h}^{(l-1)}d\mathbf{h}_{*}^{(l)}d\mathbf{h}_{*}^{(l-1)} -  \left(\mathbf{m}^{(l)}\right)^T\mathbf{E}\left[\Phi_{\ast}^{(l)}\right]_{q_{\boldsymbol\omega^{(l)}}Q_{*}}^T\mathbf{E}\left[\Phi_{\ast}^{(l)}\right]_{q_{\boldsymbol\omega^{(l)}}Q_{*}}\mathbf{m}^{(l)}\\
& = \mathbf{E}\left[\left(\mathbf{a}^{(l)}\right)^T\left(\Phi_{\ast}^{(l)}\right)^T\Phi_{\ast}^{(l)} \mathbf{a}^{(l)}\right]_{q_{\mathbf{a}^{(l)}}q_{\boldsymbol\omega^{(l)}}Q_{*}}\\
& - \left(\mathbf{m}^{(l)}\right)^T\mathbf{E}\left[\Phi_{\ast}^{(l)}\right]_{q_{\boldsymbol\omega^{(l)}}Q_{*}}^T\mathbf{E}\left[\Phi_{\ast}^{(l)}\right]_{q_{\boldsymbol\omega^{(l)}}Q_{*}}\mathbf{m}^{(l)}\\
& = \left(\mathbf{m}^{(l)}\right)^T\mathbf{E}\left[\left(\Phi_{\ast}^{(l)}\right)^T\Phi_{\ast}^{(l)}\right]_{q_{\boldsymbol\omega^{(l)}}Q_{*}}\mathbf{m}^{(l)}\\
& + \text{tr}\left(\mathbf{s}^{(l)}\mathbf{E}\left[\left(\Phi_{\ast}^{(l)}\right)^T\Phi_{\ast}^{(l)}\right]_{q_{\boldsymbol\omega^{(l)}}Q_{*}}\right)\\
&- (\mathbf{m}^{(l)})^T\mathbf{E}\left[\Phi_{\ast}^{(l)}\right]_{q_{\boldsymbol\omega^{(l)}}Q_{*}}^T\mathbf{E}\left[\Phi_{\ast}^{(l)}\right]_{q_{\boldsymbol\omega^{(l)}}Q_{*}}\mathbf{m}^{(l)}\\
& = \left(\mathbf{m}^{(l)}\right)^T\left(\Psi_{2\ast}^{(l)}-\left(\Psi_{1\ast}^{(l)}\right)^T\Psi_{1\ast}^{(l)}\right)\mathbf{m}^{(l)}+\text{tr}\left(\left(\mathbf{s}^{(l)}\right)\Psi_{2\ast}^{(l)}\right)\\
& = \left(\boldsymbol\mu^{(l)}_{H_{\mathbf{x}}+1:}\right)^T\Psi_1^{(l)}(A^{(l)})^{-1} \left(\Psi_{2\ast}^{(l)}-\left(\Psi_{1\ast}^{(l)}\right)^T\Psi_{1\ast}^{(l)}\right)(A^{(l)})^{-1}(\Psi_1^{(l)})^T\boldsymbol\mu^{(l)}_{H_{\mathbf{x}}+1:}\\
&+\left(\upsigma_{\text{noise}}^{(l)}\right)^2\text{tr}\left(\left((A^{(l)})^{-1}\right)\Psi_{2\ast}^{(l)}\right), 
\end{align*}
for $l=2\dots,L$. Again for $l=1$ and $l = L+1$ we cancel $q_{\boldsymbol{h}^{(l-1)}}(\mathbf{h}^{(l-1)})$ from the beginning and for $l = L+1$ we additionally replacing $\boldsymbol\upmu^{(l)}_{H_{\mathbf{x}}+1:}$ with $\mathbf{y}_{H_{\mathbf{x}}+1:}$.
\end{proof}\qed

\subsection{NARX structure for use with standard sparse GPs and full GP}
\label{sec:NARXstructureforusewith}
Assume we have a set of input-data $\mathbf{X}=[\mathbf{x}_1,\dots,\mathbf{x}_N]^T\in\mathbb R^{N\times Q}$ and a set of output observations $\mathbf{y}=[\mathrm{y}_1,\dots,\mathrm{y}_N]^T\in\mathbb{R}^{N}$.
A non-linear auto-regressive with exogenous inputs model can be stated algebraically and gives insights of the structure as
\begin{align*}
\textstyle\mathrm{y}_i = \mathrm{f}(\underbrace{\mathrm{y}_{i-1},\dots,\mathrm{y}_{i-H_{\mathrm{y}}},\mathbf{x}_{i-1},\dots,\mathbf{x}_{i-H_{\mathbf{x}}}}_{\textstyle{\underset{\mathrm{def}}{=}}\mathbf{u}_{i-1}\in\mathbb R^{H_{\mathrm{y}}+QH_{\mathbf{x}}}})+ \epsilon_i^{\mathrm{y}},
\end{align*}
where we model the function $\mathrm{f}:\mathbb R^{H_{\mathrm{y}}+QH_{\mathbf{x}}}\to \mathbb R$ with a GP, following a GPR model as in Section \ref{sec:GaussianProcessesandGaussianProcessRegression}.
This structure implies that the current output value $\mathrm{y}_i$ is predicted from past output values, $H_{\mathrm{y}}$ many, and past exogenous input values, $H_{\mathbf{x}}$ many, being corrupted by noise $\epsilon_{i}^{\mathrm{y}}\sim\mathcal{N}(0,\upsigma_{\text{noise}}^2)$. 
\subsection{Additional Results}
\label{sec:AdditionalResults}
\begin{figure}[H]
\centering
{\includegraphics[width=0.1\textwidth]{Fig3legend1.eps}}
{\includegraphics[scale=0.185,trim=270 0 0 0, clip=true]{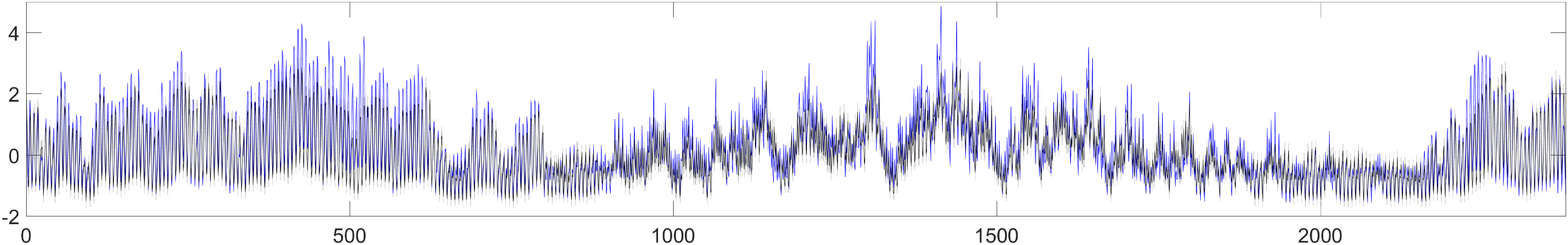}}\\
\caption{Simulation results visualized for the data-set Power for DRGP-VSS. Blue: real data, black: simulation, grey: $\pm 2$ times standard deviation}
\label{fig:Power}
\end{figure}
\clearpage


\bibliographystyle{spbasic_copie}      
\bibliography{literatur}   

\begin{thebibliography}{28}
\providecommand{\natexlab}[1]{#1}
\providecommand{\url}[1]{{#1}}
\providecommand{\urlprefix}{URL }
\expandafter\ifx\csname urlstyle\endcsname\relax
  \providecommand{\doi}[1]{DOI~\discretionary{}{}{}#1}\else
  \providecommand{\doi}{DOI~\discretionary{}{}{}\begingroup
  \urlstyle{rm}\Url}\fi
\providecommand{\eprint}[2][]{\url{#2}}

\bibitem[{Al-Shedivat et~al(2017)Al-Shedivat, Wilson, Saatchi, Hu, and
  Xing}]{al2016learning}
Al-Shedivat M, Wilson AG, Saatchi Y, Hu Z, Xing EP (2017) Learning scalable
  deep kernels with recurrent structure. \emph{Journal of Machine Learning
  Research} 18(82):1--37

\bibitem[{Bishop(2006)}]{bishop2006pattern}
Bishop CM (2006) \emph{Pattern Recognition and Machine Learning}. Springer

\bibitem[{Bui et~al(2016)Bui, Yan, and Turner}]{bui2016unifying}
Bui TD, Yan J, Turner RE (2016) {A Unifying Framework for Sparse Gaussian
  Process Approximation using Power Expectation Propagation}. \emph{arXiv
  preprint arXiv:160507066}

\bibitem[{Cutajar et~al(2016)Cutajar, Bonilla, Michiardi, and
  Filippone}]{cutajar2016practical}
Cutajar K, Bonilla EV, Michiardi P, Filippone M (2016) {Practical learning of
  deep Gaussian processes via random Fourier features}. \emph{arXiv preprint
  arXiv:161004386}

\bibitem[{Damianou(2015)}]{damianou2015deep}
Damianou A (2015) Deep gaussian processes and variational propagation of
  uncertainty. \emph{PhD Thesis, University of Sheffield}

\bibitem[{Damianou and Lawrence(2013)}]{damianou2013deep}
Damianou A, Lawrence N (2013) Deep {G}aussian processes. In: Carvalho C,
  Ravikumar P (eds) Proceedings of the Sixteenth International Workshop on
  Artificial Intelligence and Statistics (AISTATS), JMLR W\&CP 31, AISTATS '13,
  pp 207--215

\bibitem[{Frigola et~al(2014)Frigola, Chen, and
  Rasmussen}]{frigola2014variational}
Frigola R, Chen Y, Rasmussen CE (2014) Variational {G}aussian process
  state-space models. In: Ghahramani Z, Welling M, Cortes C, Lawrence N,
  Weinberger K (eds) Advances in Neural Information Processing Systems 27
  (NIPS)

\bibitem[{Gal and Turner(2015)}]{gal2015improving}
Gal Y, Turner R (2015) Improving the gaussian process sparse spectrum
  approximation by representing uncertainty in frequency inputs. In: Bach F,
  Blei D (eds) Proceedings of the 32nd International Conference on Machine
  Learning, PMLR, Lille, France, Proceedings of Machine Learning Research,
  vol~37, pp 655--664

\bibitem[{Gal et~al(2014)Gal, van~der Wilk, and Rasmussen}]{gal2014distributed}
Gal Y, van~der Wilk M, Rasmussen CE (2014) Distributed variational inference in
  sparse gaussian process regression and latent variable models. In: Ghahramani
  Z, Welling M, Cortes C, Lawrence ND, Weinberger KQ (eds) Advances in Neural
  Information Processing Systems 27, Curran Associates, Inc., pp 3257--3265

\bibitem[{Hoang et~al(2017)Hoang, Hoang, and Low}]{hoang2017generalized}
Hoang QM, Hoang TN, Low KH (2017) {A Generalized Stochastic Variational
  Bayesian Hyperparameter Learning Framework for Sparse Spectrum Gaussian
  Process Regression.} In: AAAI, pp 2007--2014

\bibitem[{Hochreiter and Schmidhuber(1997)}]{hochreiter1997long}
Hochreiter S, Schmidhuber J (1997) Long short-term memory. \emph{Neural Comput}
  9(8):1735--1780, \doi{10.1162/neco.1997.9.8.1735}

\bibitem[{Kallenberg(2006)}]{kallenberg2006foundations}
Kallenberg O (2006) \emph{{Foundations of modern probability}}. Springer
  Science \& Business Media

\bibitem[{L{\'{a}}zaro-Gredilla et~al(2010)L{\'{a}}zaro-Gredilla,
  Qui{\~{n}}onero-Candela, Rasmussen, and Figueiras-Vidal}]{quia2010sparse}
L{\'{a}}zaro-Gredilla M, Qui{\~{n}}onero-Candela J, Rasmussen CE,
  Figueiras-Vidal AR (2010) {Sparse spectrum Gaussian process regression}.
  \emph{Journal of Machine Learning Research} 11(Jun):1865--1881

\bibitem[{Mattos et~al(2015)Mattos, Dai, Damianou, Forth, Barreto, and
  Lawrence}]{mattos2015recurrent}
Mattos CLC, Dai Z, Damianou A, Forth J, Barreto GA, Lawrence ND (2015)
  {Recurrent Gaussian processes}. \emph{arXiv preprint arXiv:151106644}

\bibitem[{Nelles(2013)}]{nelles2013nonlinear}
Nelles O (2013) \emph{{Nonlinear system identification: from classical
  approaches to neural networks and fuzzy models}}. Springer Science \&
  Business Media

\bibitem[{Pascanu et~al(2013)Pascanu, Gulcehre, Cho, and
  Bengio}]{pascanu2013construct}
Pascanu R, Gulcehre C, Cho K, Bengio Y (2013) How to construct deep recurrent
  neural networks. \emph{arXiv preprint arXiv:13126026}

\bibitem[{Rasmussen and Williams(2006)}]{rasmussen2006gaussian}
Rasmussen C, Williams C (2006) \emph{Gaussian Processes for Machine Learning}.
  Adaptive Computation and Machine Learning, MIT Press, Cambridge, MA, USA

\bibitem[{Sj{\"o}berg et~al(1995)Sj{\"o}berg, Zhang, Ljung, Benveniste, Delyon,
  Glorennec, Hjalmarsson, and Juditsky}]{sjoberg1995nonlinear}
Sj{\"o}berg J, Zhang Q, Ljung L, Benveniste A, Delyon B, Glorennec PY,
  Hjalmarsson H, Juditsky A (1995) {Nonlinear black-box modeling in system
  identification: a unified overview}. \emph{Automatica} 31(12):1691--1724

\bibitem[{Snelson and Ghahramani(2006)}]{snelson2006sparse}
Snelson E, Ghahramani Z (2006) {Sparse Gaussian processes using pseudo-inputs}.
  \emph{Advances in Neural Information Processing Systems} 18:1257

\bibitem[{Stein(2012)}]{stein2012interpolation}
Stein ML (2012) \emph{{Interpolation of spatial data: some theory for
  kriging}}. Springer Science \& Business Media

\bibitem[{Svensson et~al(2015)Svensson, Solin, S{\"a}rkk{\"a}, and
  Sch{\"o}n}]{svensson2015computationally}
Svensson A, Solin A, S{\"a}rkk{\"a} S, Sch{\"o}n TB (2015) {Computationally
  efficient Bayesian learning of Gaussian process state space models}.
  \emph{arXiv preprint arXiv: 1506: 02267}

\bibitem[{Titsias(2009)}]{titsias2009variational}
Titsias M (2009) Variational learning of inducing variables in sparse gaussian
  processes. In: van Dyk D, Welling M (eds) Proceedings of the Twelth
  International Conference on Artificial Intelligence and Statistics, PMLR,
  Hilton Clearwater Beach Resort, Clearwater Beach, Florida USA, Proceedings of
  Machine Learning Research, vol~5, pp 567--574

\bibitem[{Titsias and Lawrence(2010)}]{titsias2010bayesian}
Titsias M, Lawrence ND (2010) Bayesian gaussian process latent variable model.
  In: Teh YW, Titterington M (eds) Proceedings of the Thirteenth International
  Conference on Artificial Intelligence and Statistics, PMLR, Chia Laguna
  Resort, Sardinia, Italy, Proceedings of Machine Learning Research, vol~9, pp
  844--851

\bibitem[{Turner et~al(2010)Turner, Deisenroth, and
  Rasmussen}]{turner2010state}
Turner R, Deisenroth M, Rasmussen C (2010) State-space inference and learning
  with gaussian processes. In: Teh YW, Titterington M (eds) Proceedings of the
  Thirteenth International Conference on Artificial Intelligence and
  Statistics, PMLR, Chia Laguna Resort, Sardinia, Italy, Proceedings of Machine
  Learning Research, vol~9, pp 868--875

\bibitem[{Wang et~al(2009)Wang, Sano, Chen, and Huang}]{wang2009identification}
Wang J, Sano A, Chen T, Huang B (2009) {Identification of Hammerstein systems
  without explicit parameterisation of non-linearity}. \emph{International
  Journal of Control} 82(5):937--952

\bibitem[{Wang et~al(2005)Wang, Fleet, and Hertzmann}]{wang2005gaussian}
Wang JM, Fleet DJ, Hertzmann A (2005) {Gaussian process dynamical models}. In:
  NIPS, vol~18, p~3

\bibitem[{Wigren(2010)}]{wigren2010input}
Wigren T (2010) {Input-output data sets for development and benchmarking in
  nonlinear identification}. \emph{Technical Reports from the department of
  Information Technology} 20:2010--020

\bibitem[{Williams and Seeger(2000)}]{williams2000using}
Williams CK, Seeger M (2000) {Using the Nystr{\"o}m method to speed up kernel
  machines}. In: Proceedings of the 13th International Conference on Neural
  Information Processing Systems, MIT press, pp 661--667

\end{thebibliography}

%
%

\end{document}